\documentclass[conference,compsoc]{IEEEtran}
\usepackage{booktabs}
\usepackage[T1]{fontenc} 
\usepackage{amssymb} 
\usepackage{pifont}  
\usepackage{rotating}
\usepackage{amsmath}
\usepackage{epsfig,endnotes}
\usepackage[colorinlistoftodos,prependcaption,textsize=tiny]{todonotes}
\usepackage{hyperref}

\usepackage{mathtools}
\usepackage{url}
\usepackage{makecell}
\usepackage[noend]{algpseudocode}
\usepackage{tabularx,colortbl}
\usepackage{multirow}
\usepackage[Symbolsmallscale]{upgreek}
\usepackage{algorithm}
\usepackage{algpseudocode}
\usepackage{mathtools}
\usepackage{url}
\usepackage{tablefootnote}
\usepackage{siunitx}
\usepackage{gensymb}
\usepackage{listings}
\usepackage{romannum}
\usepackage[framemethod=tikz]{mdframed}
\mdfdefinestyle{remarkstyle}{
backgroundcolor=lightgray,
roundcorner=3pt,
innerleftmargin=3pt,
innerrightmargin=3pt,
innertopmargin=3pt,
innerbottommargin=3pt,
linewidth=0.5pt,
}
\newcounter{insight}
\newenvironment{insight}{\refstepcounter{insight}
\begin{mdframed}[style=remarkstyle]
\noindent \textbf{Insight~\theinsight}: \em
}
{
\end{mdframed}
\vspace{-1mm}
}

\usepackage{color} 
\definecolor{mygreen}{RGB}{28,172,0} 
\definecolor{mylilas}{RGB}{170,55,241}

\def\EC{\textit{PaniCar}} 
\def\CC{\textit{Caracetamol}}

\title{PaniCar: Securing the Perception of Advanced Driving Assistance Systems Against Emergency Vehicle Lighting}

\author{
\IEEEauthorblockN{
Elad Feldman\IEEEauthorrefmark{1},
Jacob Shams\IEEEauthorrefmark{1},
Dudi Biton\IEEEauthorrefmark{1},
Alfred Chen\IEEEauthorrefmark{2},
Shaoyuan Xie\IEEEauthorrefmark{2}\\
Satoru Koda\IEEEauthorrefmark{3},
Yisroel Mirsky\IEEEauthorrefmark{1},
Asaf Shabtai\IEEEauthorrefmark{1},
Yuval Elovici\IEEEauthorrefmark{1},
Ben Nassi\IEEEauthorrefmark{1}
}
\IEEEauthorblockA{\IEEEauthorrefmark{1}Ben-Gurion University of the Negev, Israel\\
\{eladfeld, jacobsh, bitondud, yisroel, nassib\}@post.bgu.ac.il, \{shabtaia, elovici\}@bgu.ac.il}
\IEEEauthorblockA{\IEEEauthorrefmark{2}University of California, Irvine, USA\\
\{alfchen, shaoyux\}@uci.edu}
\IEEEauthorblockA{\IEEEauthorrefmark{3}Fujitsu Limited, Japan\\
koda.satoru@fujitsu.com}
}

\begin{document}
\maketitle
\thispagestyle{empty}

\begin{abstract}
The safety of autonomous cars has come under scrutiny in recent years, especially after 16 documented incidents involving Teslas (with autopilot engaged) crashing into parked emergency vehicles (police cars, ambulances, and firetrucks). 
While previous studies have revealed that strong light sources often introduce flare artifacts in the captured image, which degrade the image quality, the impact of flare on object detection performance remains unclear. 
In this research, we unveil \EC, a digital phenomenon that causes an object detector's confidence score to fluctuate below detection thresholds when exposed to activated emergency vehicle lighting. 
This vulnerability poses a significant safety risk, and can cause autonomous vehicles to fail to detect objects near emergency vehicles. In addition, this vulnerability could be exploited by adversaries to compromise the security of advanced driving assistance systems (ADASs).
We assess seven commercial ADASs (Telsa Model 3, "manufacturer C", HP, Pelsee, AZDOME, Imagebon, Rexing), four object detectors (YOLO, SSD, RetinaNet, Faster R-CNN), and 14 patterns of emergency vehicle lightning to understand the influence of various technical and environmental factors. We also evaluate four SOTA flare removal methods and show that their performance and latency are insufficient for real-time driving constraints.
To mitigate this risk, we propose \CC, a robust framework designed to enhance the resilience of object detectors against the effects of activated emergency vehicle lighting. Our evaluation shows that on YOLOv3 and Faster R-CNN, \CC\ improves the models’ average confidence of car detection by 0.20, the lower confidence bound by 0.33, and reduced the fluctuation range by 0.33. In addition, \CC\ is capable of processing frames at a rate of between 30-50 FPS, enabling real-time ADAS car detection.
\end{abstract}


\section{Introduction}

The safety of autonomous cars has been questioned in recent years following 16 documented incidents \cite{NHTSA-report} in which Tesla cars with engaged autopilot collided with parked emergency vehicles, such as fire engines \cite{Fire-truck-crash}, police cars \cite{Police-crash}, and ambulances \cite{Ambulances-crash}. 
These accidents resulted in significant injuries and fatalities \cite{autopilot_crashing}, causing the National Highway Transportation Safety Administration (NHTSA) to launch an in-depth investigation of the incidents to determine the factors underlying these accidents \cite{tesla_vs_emergency}. 
While driver distraction has been identified as a primary cause \cite{autopilot_crashing}, the technical aspects that led to these accidents remain a mystery \cite{autopilot_crashing}. 

\begin{figure}[]
  \centering
    
        \includegraphics[width=0.49\linewidth]{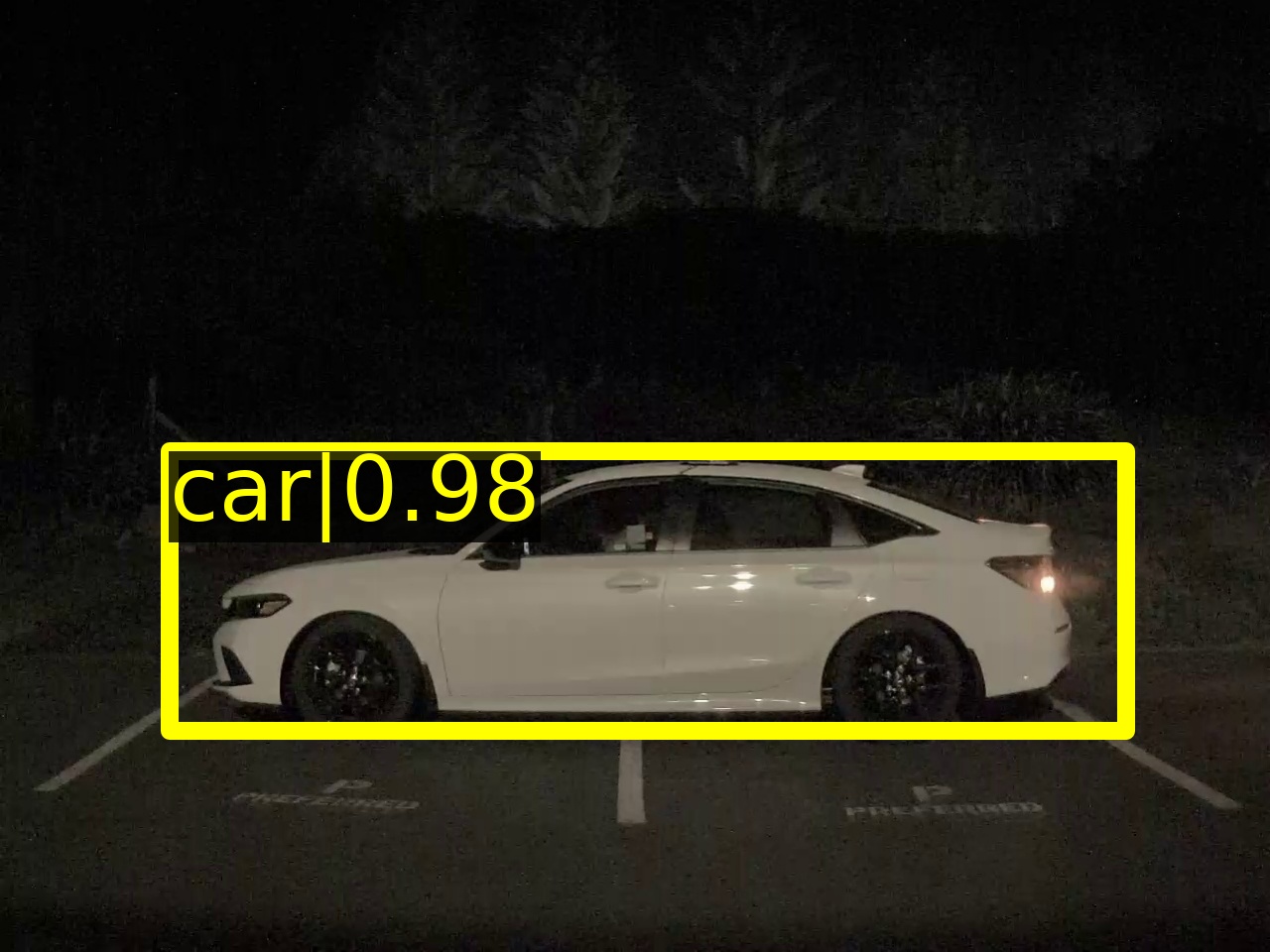}
        \includegraphics[width=0.49\linewidth]{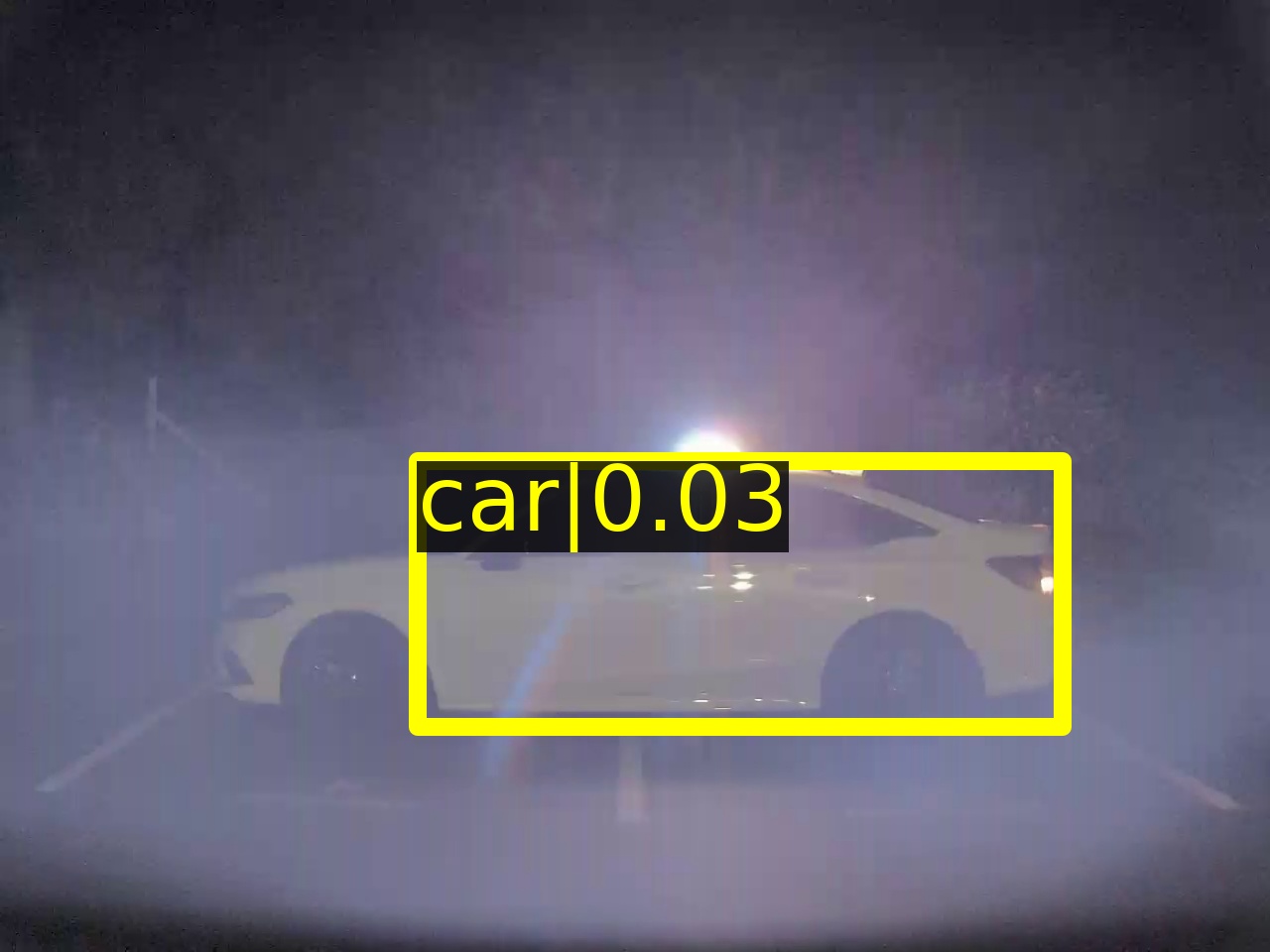}
        
           \includegraphics[width=0.49\linewidth, height=0.34\linewidth]{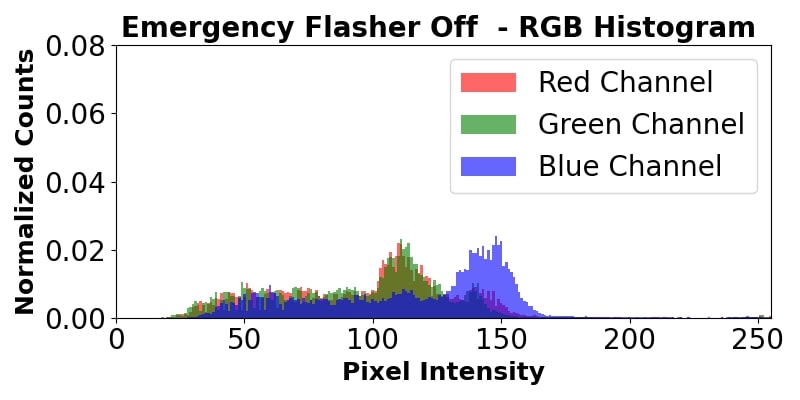}
        \includegraphics[width=0.49\linewidth, height=0.34\linewidth]{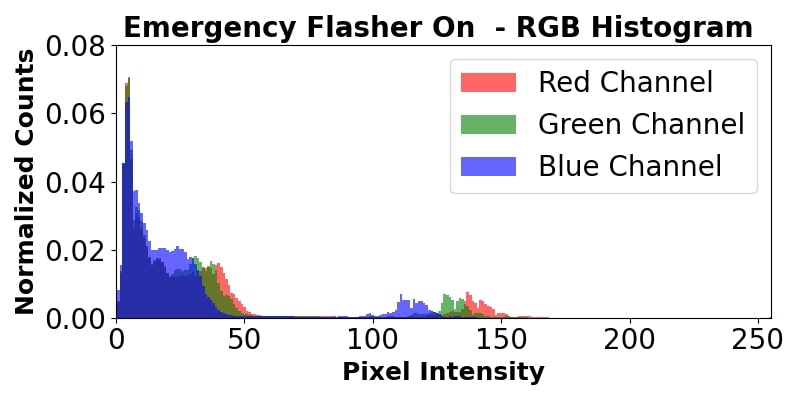}
       
    \caption{Top: A vehicle detected by SSD with a confidence of 0.98 with emergency vehicle lighting off (left) and 0.03 with emergency vehicle lighting on (right) in images obtained by a 2023 Tesla model 3 frontal camera. Bottom: The associated normalized tonal distribution of the images. 
    }
    \label{fig:vehicle_detection}
    \vspace{-0.5cm}
\end{figure}

Despite public interest and the importance of the investigation, neither the automotive industry nor NHTSA has disclosed their findings, leaving critical questions unanswered regarding the technological factors behind these incidents. 
Prior studies \cite{dai2022flare7k, zhou2023improving, wu2021train} have demonstrated that strong light sources—such as those from street lamps—can introduce flare artifacts, degrading image quality by lowering metrics like peak signal-to-noise ratio (PSNR) and structural similarity index (SSIM). 
These studies establish that flare degrades the visual quality of single images captured by standard video cameras in the presence of static light sources.
However, the incidents in question involved advanced driving assistance systems (ADASs) cameras (rather than standard video cameras), failures in perception and object detection over time (rather than quality degradation in a single frame), and emergency vehicle lighting, which emits variable-intensity light in dynamic patterns (unlike constant-intensity street lamps). 
The gap between the conditions studied in prior work \cite{dai2022flare7k,zhou2023improving,wu2021train} and the real-world conditions relevant to the documented failures leaves a critical question unaddressed: What is the impact of flare from emergency vehicle lighting on object detection performance over time in videos containing fire trucks, ambulances, and police cars?

This question must be answered, especially in light of the increased adoption of semi-autonomous cars and ADASs, for multiple reasons: (1) Such accidents can result in injuries or the death of drivers, passengers, and pedestrians  \cite{autopilot_crashing}. (2) There is an industry shift in the architectural design of new semi-autonomous cars that is increasingly leaning toward vision-based systems \cite{Tesla-Vision} without RADARs and ultrasonic sensors.
(3) Until these questions are answered, the ability to develop methods to secure the perception of semi-autonomous cars and other perception-based systems (e.g., drones, humanoids) will be curtailed, exposing the perception of autonomous vehicles to unintentional safety faults caused by emergency vehicle lightning or intentional security attacks performed by adversaries.

This study aims to explore the role of object detectors in the failure of vision-based perception when encountering a vehicle with activated emergency vehicle lighting.
This is not only important from a safety perspective, but also from a security perspective, since attackers can leverage this failure to target semi-autonomous cars and deliberately cause damage and endanger the lives of road users.
First, we analyze video footage obtained from two commercial Level 2 ADASs (Tesla Model 3 and "manufacturer C") and five commercial Level 1 ADASs (HP \cite{hp}, Pelsee \cite{pelsee}, AZDOME \cite{azdome}, Imagebon \cite{imagebon}, Rexing \cite{rexing}) and show that four object detectors are vulnerable to a phenomenon we have termed \EC, a digital fluctuation in the confidence score of object detectors of a detected object caused by the activation of emergency vehicle lighting.
This has dangerous implications, as the fluctuation in the confidence score may result in a value below the detection threshold, causing an object to go unnoticed (\textit{i.e.,} undetected) by semi-autonomous cars and ADASs.

In our investigation of the \EC\ phenomenon, we discover that the confidence score of an object detector is affected by the intensity of the light emitted from activated emergency vehicle lighting. 
The added flare from the activated emergency vehicle lighting changes the tonal distribution of the car (in the frame captured by the video camera) over time (between different frames) according to the pattern and frequency of the activated emergency vehicle lighting.
Consequently, lower confidence detection scores are produced by the object detector in different frames, affecting the object detector's behavior and performance and causing detection loss of objects that appear in the frame (see Fig. \ref{fig:vehicle_detection}).
We also investigate the factors affecting the \EC\ phenomenon and find that it appears in a variety of object detectors and commercial ADASs, creates unique detection behavior depending on the pattern of the emergency vehicle lighting, and mainly compromises object detectors' performance in dark conditions. 

To address the \EC\ phenomenon, we evaluate the effect of state-of-the-art (SOTA) flare removal methods \cite{dai2022flare7k, zhou2023improving, wu2021train} on the performance of object detectors and find a scientific gap: while SOTA flare removal methods improve the quality of the picture (the PSNR and SSIM), they do not improve the performance of the object detectors significantly and cannot be applied in real-time settings.
We introduce \CC, a framework aimed at improving the performance of a given object detector in the presence of emergency vehicles.
This is done using a multi-component architecture consisting of (1) a CycleGAN-based denoiser which removes flare from activated emergency vehicle lighting, feeding the resulting images into (2) a replication of the original ADAS object detector refined on augmented data we created due to the lack of images of emergency vehicle lighting operated at night. This is performed in conjunction with (3) feeding the original frame to the original ADAS object detector to ensure all original detections are preserved, and (4) a combiner layer to aggregate the detections of the original ADAS and the fine-tuned replication.
We tested the performance of three implementations for \CC, based on three different object detectors (YOLO, SSD, Faster R-CNN), using videos of emergency vehicles downloaded from YouTube and found that it significantly improves the resiliency of an object detector to \EC\ . 
On YOLOv3 and Faster R-CNN, \CC\ improves the models’ average confidence of car detection by 0.20, the lower confidence bound by 0.33, and reduced the fluctuation range by 0.33. In addition, \CC\ is capable of processing frames at a rate of between 30 and 50 FPS, enabling real-time ADAS car detection. \textit{Caracetamol} can improve the average detection confidence of a Faster R-CNN-based ADAS by 26.98\%, with no additional runtime overhead. Alternatively, \textit{Caracetamol} can improve the average detection in 42\% with an additional latency overhead of 23.5\%-25\% by using a YOLO-based ADAS with a \textit{Caracetamol} denoiser preprocessor.

\textbf{Contributions.} This paper makes the following contributions: (1) we empirically investigate in seven ADASs and popular open-source object detectors the reasons for the degradation in detection performance in the presence of activated emergency vehicle lighting; (2) in the absence of a relevant comprehensive dataset, we collect and augment a unique dataset that includes over 40,000 images of emergency response vehicles that we make available to the research community to enable further research; and (3) we evaluate SOTA flare removal methods and show that their performance and latency fail to satisfy real time driving constraints. We introduce a framework that significantly improves the resilience of object detectors against the \EC\ phenomenon significantly.

\textbf{Disclaimer.} 
Lacking access to the specific type of object detectors deployed in Tesla, we perform our analysis based on commonly used object detectors.
For this reason, there may be a discrepancy between our findings and the actual technological factors underlying the documented Tesla accidents.
We hope this paper will encourage the automotive industry to validate our findings on their ADASs and semi-autonomous cars with deployed object detectors. 

\textbf{Structure.} 
The paper is structured as follows: 
We present the threat model in Section \ref{sec:threat-model}.
In Section \ref{sec:understanding}, we examine why object detectors' performance degrades in the presence of activated emergency vehicle lighting. 
In Section \ref{sec:countermeasures}, we describe and evaluate \CC, a framework aimed at securing an object detector against the \EC\ phenomenon. 
In Section \ref{sec:limitaitons} we discuss the limitations of our research and in Section \ref{sec:related-works}, we review related work. We discuss our findings in Section \ref{sec:discussion}.

\textbf{Responsible Disclosure} 
We disclosed our findings to the NHTSA, Tesla, "manufacturer C", and the five manufacturers of the Level 1 ADASs. 
The NHTSA replied "\textit{We are aware of some advanced driver assistance systems that have not responded appropriately when emergency flashing lights were present in the scene of the driving path under certain circumstances}." Tesla replied "\textit{Although adversarial modifications to the environment are outside the scope of our bug bounty program, we appreciate being informed of new developments in this area.}"
We will add the responses of the rest of the manufacturers to the paper when provided.

\textbf{Ethical Considerations } In this paper, we utilized publicly available video footage of cars. 
As part of the necessary procedures for ensuring ethical research with personally identifiable information, we requested and obtained Institutional Review Board (IRB) approval regarding the usage, storage, and disposal of the data utilized in this paper. 
To protect the privacy of identifiable information in the video footage, the footage and derived experimental data are stored on an access-controlled database limited to only the researchers. 
In addition, all experiments were performed on the university premises after we had received the required approvals.
\section{Threat Model}
\label{sec:threat-model}

In this section, we review the threat model in two related yet distinct cases; unintentional and intentional threats.



\textbf{Unintentional Threat Model (resulting from a safety issue).} In this case, an ADAS encounters activated emergency vehicle lighting, causing the ADAS to unintentionally demonstrate undesired behavior. 
This is a safety issue caused by commonly occurring natural phenomena, activated emergency vehicle lighting. The involved actors are: (1) a car controlled by an ADAS, i.e., a car with an autonomy level of 2+ to 5, with autopilot functionality engaged, and (2) an emergency vehicle (e.g., police car, ambulance, fire truck) with its emergency vehicle lighting activated. 
The scenario is as follows: a distracted driver is driving their car towards an emergency vehicle in dark conditions; the autopilot functionality (controlled by the ADAS) is engaged; the emergency lights of the emergency vehicle are flashing; and the emergency vehicle is parked in the path of the car. 
A distracted driver who was not focused on monitoring the car behavior would be unable to properly respond to the failed or late detection of the emergency vehicle \cite{autopilot_crashing}.
This unfortunate scenario can result in an accident involving the car, the emergency vehicle, and nearby cars and pedestrians. 
We note that this was the case associated with 16 documented incidents in which Teslas collided with emergency vehicles.
A video of one of the incidents was recently published in \cite{wall-street-journal}. 

\textbf{Intentional Threat Model (the result of an attack).}
In this case, an attacker exploits the safety issue stemming from object detectors' undesired perception and reaction to emergency vehicle lighting, in order to attack the ADAS and deliberately cause undesired behavior.
Much like how Rowhammer \cite{kim2014flipping} is a safety issue in DRAM memory, that was exploited by attackers to compromise the security of computers \cite{razavi2016flipfengshui, veen2016drammer, frigo2018grand, gruss2018another, pessl2016drama}, \EC\ can be exploited by attackers to compromise the security and safety of ADASs.
By exploiting emergency vehicle lighting, attackers can cause undesired reactions of ADASs which could threaten the safety of road users.
We categorize two types of attacks: \textit{untargeted attack} and \textit{targeted attack}.
In an untargeted attack, the attacker utilizes an emergency vehicle lighting at random areas and random times of day to disrupt the functionality of ADASs that happen to cross the affected area. 
Such attackers have the desire to induce an undesired reaction from any ADAS and/or don't target a particular target ADAS.
In a targeted attack, the attacker targets a specific car and operates the emergency vehicle lighting when it is recognized nearby.


\textbf{Potential Outcomes.}
The failure to detect emergency vehicles can lead to:
(1) Safety Hazards: There is an increased risk of collisions with emergency vehicles, posing serious safety threats to both vehicle occupants and emergency first responders. 
Notably, this risk has already led to 16 documented accidents between Teslas and emergency vehicles, resulting in injuries. 
(2) System Reliability Issues: Detection failures, particularly those that necessitate manual intervention or overrides by the driver, can affect the overall reliability of ADASs, eroding confidence in their adoption.


\begin{table}[]
\centering
\caption{Functionalities and specifications of the ADASs used in this research.}
\label{tab:adass}
\resizebox{1.0\columnwidth}{!}{%
\begin{tabular}{|l|c|c|c|c|c|c|}
\hline
\begin{tabular}[c]{@{}l@{}}ADAS\\ Name\end{tabular} &
\begin{tabular}[c]{@{}l@{}}Lane\\ Departure\end{tabular} &
\begin{tabular}[c]{@{}l@{}}Collision\\ Warnings\end{tabular} &
\begin{tabular}[c]{@{}l@{}}Vehicle\\ Detection\end{tabular} &
FPS & Resolution & \begin{tabular}[c]{@{}l@{}}Automation\\ Level\end{tabular} \\ \hline
\href{https://www.amazon.com/dp/B081YDHHBR/}{HP} & \checkmark & \ding{55} & \ding{55} & 30 & 1080p & Level 1 \\ \hline
\href{https://www.amazon.com/dp/B0BF4XB3VP/}{Pelsee} & \checkmark & \checkmark & \checkmark & 60 & 1440p & Level 1 \\ \hline
\href{https://www.amazon.com/dp/B094YDVV7L/}{AZDOME} & \checkmark & \checkmark & \ding{55} & 24 & 720p & Level 1 \\ \hline
\href{https://www.amazon.com/dp/B0C86CV679/}{Imagebon} & \checkmark & \checkmark & \checkmark & 60 & 1080p & Level 1 \\ \hline
\href{https://www.amazon.com/dp/B08N1KMSZ7/}{Rexing} & \checkmark & \checkmark & \checkmark & 30 & 1080p & Level 1 \\ \hline
\href{https://www.tesla.com/model3}{Tesla Model 3} & \checkmark & \checkmark & \checkmark & 36 & 1280×960 & Level 2 \\ \hline
\href{https://notaURL.com}{”manufacturer C”} & \checkmark & \checkmark & \checkmark & 30 & 1920×1080 & Level 2 \\ \hline
\end{tabular}
}
\vspace{-1.0em}
\end{table}

\begin{figure}[]
  \centering
    
        \includegraphics[width=0.45\linewidth]{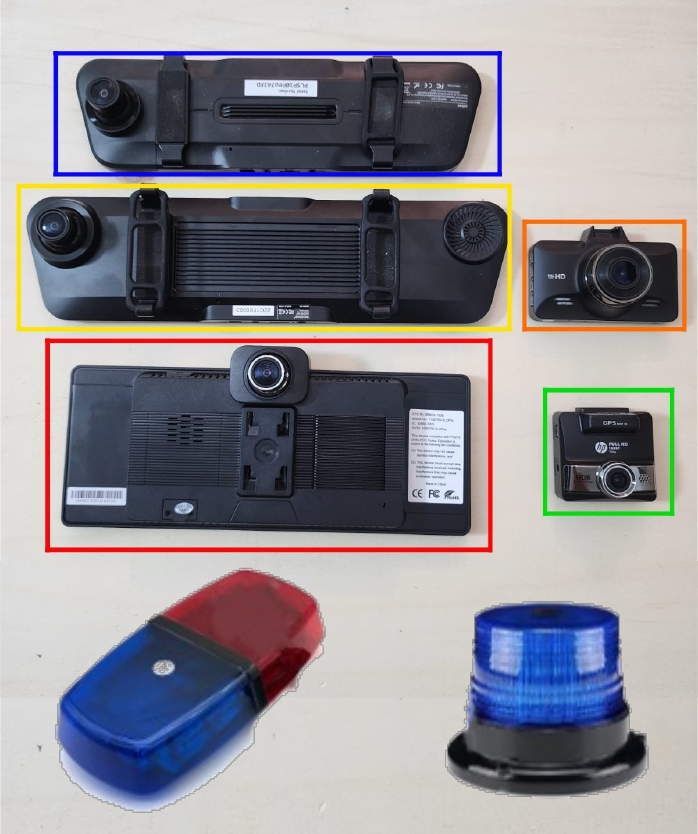}
        \includegraphics[width=0.45\linewidth]{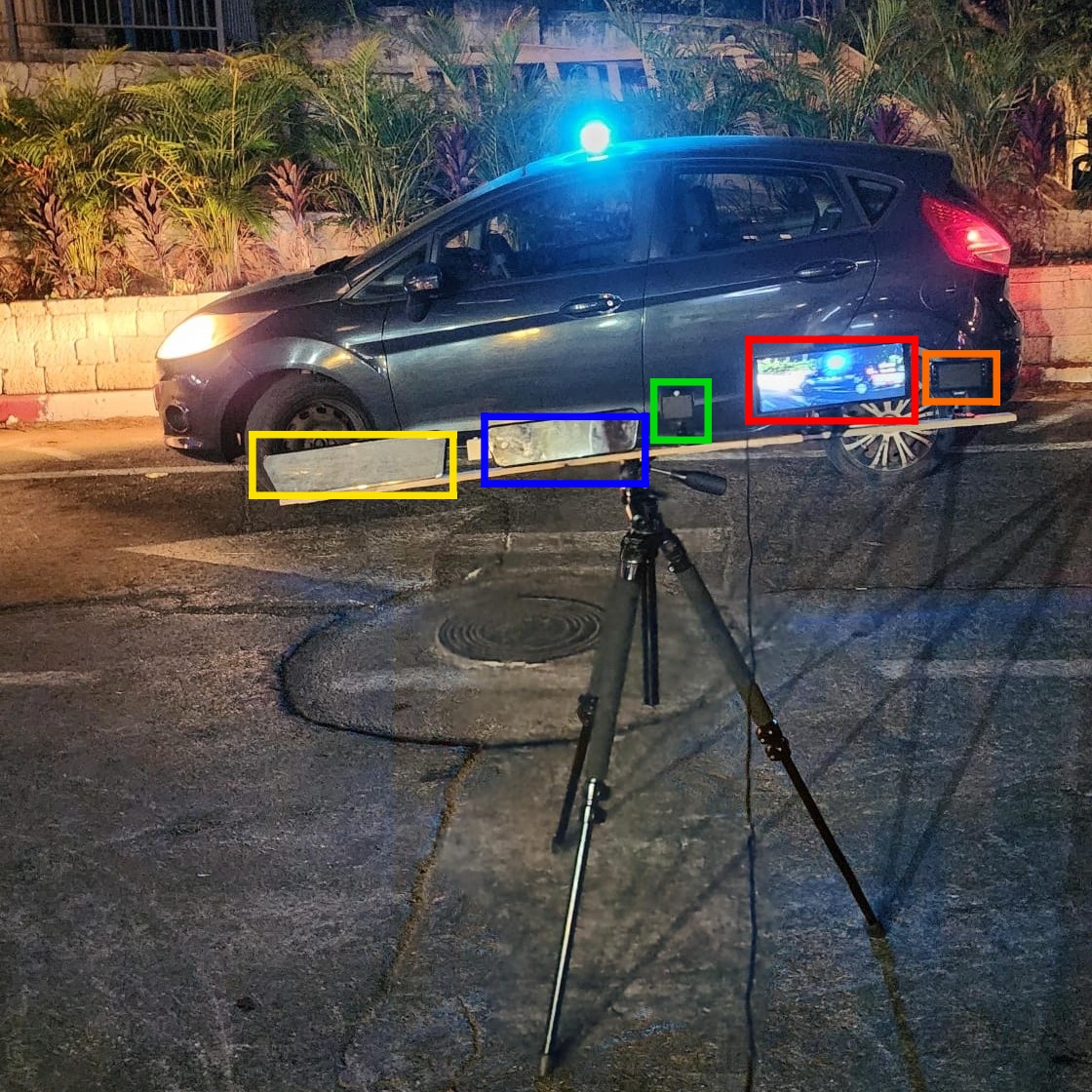}
    \caption{The emergency vehicle lightning and ADASs we used: Rexing is marked in Yellow, Pelsee is marked in Blue, HP is marked in Green, Imegebon is marked in Red, and AZDOME is marked in Orange.
    }
    \label{fig:commercial-adas}
    \vspace{-5mm}
\end{figure}

\begin{figure*}[t]
  \centering
  \includegraphics[width=0.32\textwidth]{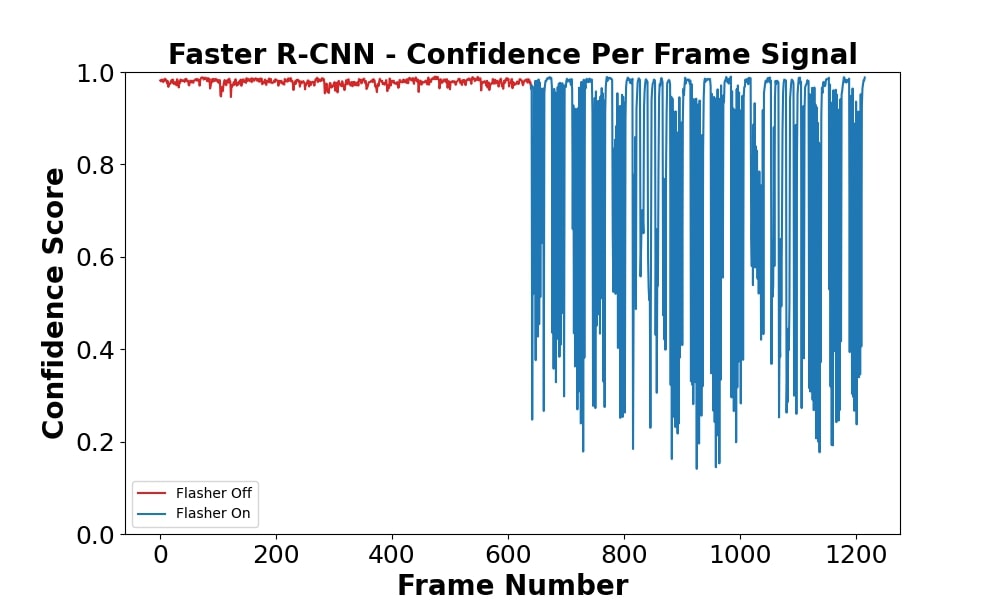}
\includegraphics[width=0.32\textwidth]{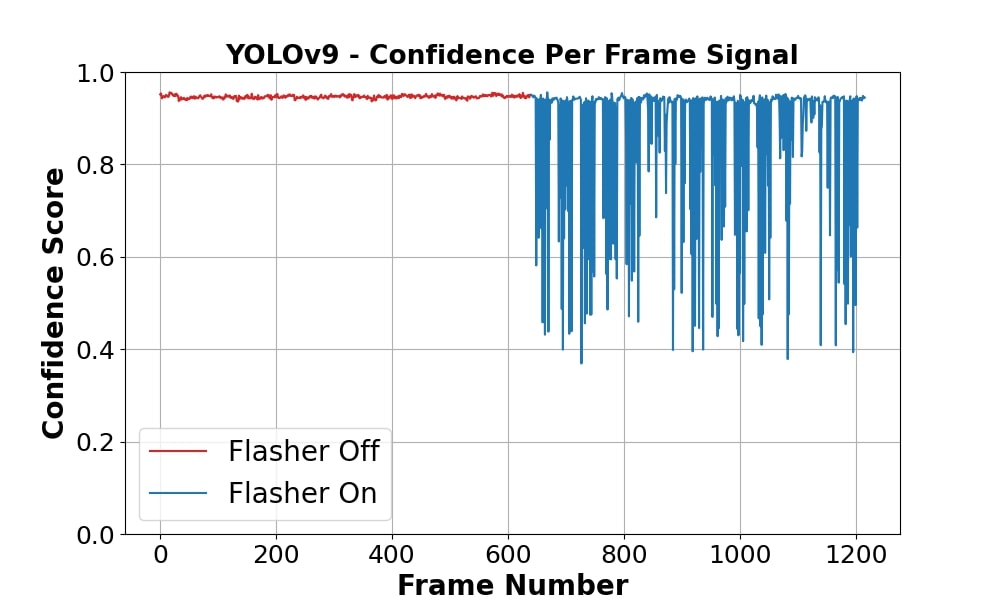}
\includegraphics[width=0.32\textwidth]{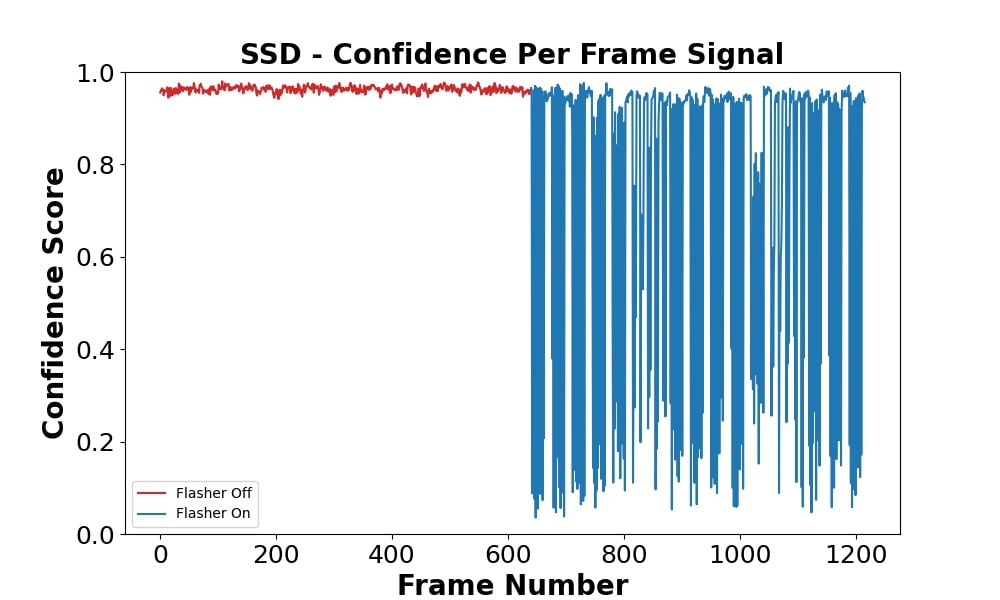}
\caption{Comparison of the confidence score signal for Faster R-CNN, YOLOv9, and SSD using video obtained by the frontal camera of a 2023 Tesla model 3.}
    \label{fig:all-ods}
    \vspace{-5mm}
\end{figure*}

\section{Understanding the PaniCar Phenomenon}
\label{sec:understanding}

In this section, we describe the analysis we performed to understand the technical factors that led advanced driver-assistance system (ADAS) controlled cars with the autopilot engaged to crash into emergency vehicles.
We note that driver distraction was already raised as a primary cause of these accidents \cite{autopilot_crashing}. 
The analysis described in this section focuses on the technical aspects of these accidents in an attempt to understand the role of object detectors in this failure.

Lens flare is a well-known optical phenomenon that occurs when light scatters within a lens system—often due to a bright light source—producing unwanted artifacts in the image.\footnote{\url{https://www.picturecorrect.com/lens-flare-from-accidental-to-artistic-in-photography/?utm_source=chatgpt.com}}
There are several types of lens flare, which can manifest as visual artifacts in various shapes and colors. This typically happens when a bright light source, such as the sun, is within or just outside the frame.
Although modern lenses incorporate specialized coatings and optical designs to reduce flare, no system is entirely immune—particularly under intense lighting conditions.
Prior studies have identified multiple causes of lens flare, including manufacturing imperfections and microscopic scratches accumulated during everyday use. \cite{hullin2011lensflare}

Recent studies have already established that flare lens degrades image quality by lowering metrics like peak signal-to-noise ratio (PSNR) and
structural similarity index (SSIM)\cite{dai2022flare7k}. However, the impact of lens flare on object detectors' performance and perception accuracy remains underexplored. Existing studies \cite{dai2022flare7k, zhou2023improving, wu2021train} have primarily focused on: (1) assessing the change in image quality due to lens flare rather than its effect on object detection performance, (2) flare caused by street lamps, which emit fixed light over time, as opposed to the dynamic and varying intensity of emergency vehicle lighting determined by their pattern, (3) individual images with flare, rather than videos (i.e., sequences of frames), and (4) regular video cameras, rather than the specialized video cameras used in Advanced Driver-Assistance Systems (ADAS). The analysis we perform in this section uses object detectors on a series of frames (videos) captured by commercial ADAS video cameras of emergency vehicle lighting.

We first explain the \EC\ phenomenon by analyzing video footage of a car with activated emergency vehicle lighting using video footage obtained by seven ADASs: the front facing camera of Tesla Model 3, "manufacturer C", HP \cite{hp}, Pelsee \cite{pelsee}, AZDOME \cite{azdome}, Imagebon \cite{imagebon}, Rexing \cite{rexing}. 
These ADASs were chosen due to them being off-the-shelf level one and two ADASs (see Fig. \ref{fig:commercial-adas}) and support functions of forward-collision warning, lane departure warning, and vehicle detection (see Table \ref{tab:adass}). 
We also obtained video using a Samsung Galaxy S22 Ultra.

The video footage was analyzed using four object detectors (YOLOv9 \cite{wang2024yolov9}, Faster R-CNN \cite{ren2015faster}, SSD \cite{liu2016ssd}, and RetinaNet \cite{lin2017focal}).
We selected these object detectors because three of them (YOLO, SSD, and RetinaNet) are considered the \textit{"most popular one-stage object detectors"}\footnote{\url{https://viso.ai/deep-learning/object-detection/}} used for real-time applications, and we selected Faster R-CNN because it is considered a time-efficient two-stage object detector \cite{dey2022robust}. 
For YOLOv9, we used its open-source pretrained implementation, and for the other three object detectors, we used the pretrained implementations provided by MMDetection \cite{chen2019mmdetection}.
All four object detectors were pretrained on the COCO dataset.
Some of the experiments in this section were conducted with YOLOv3 instead of YOLOv9 (this is noted in the relevant experiments), because YOLOv3 is also implemented in MMDetection, which made the analysis easier.
Later in this section, we demonstrate that the application of object trackers on top of object detectors does not mitigate \EC. We also evaluate the influence of ambient light.

In the appendix, we performed additional series of experiments aimed at testing the effect of an ADAS' camera settings, the effect of various characteristics of the emergency vehicle lighting (color, orientation, etc.), and the influence of distance between the camera and the emergency vehicle lighting, on the \EC\ phenomenon. Since we have not made any surprising insights from these series of experiments, we placed them in the appendix (for the completeness of the paper).

\subsection{Analysis of Object Detector Behavior}

In this section, we analyze the behavior of object detectors when detecting objects in an environment with and without activated emergency vehicle lighting.

\subsubsection{Time-based Analysis}
\label{sec:video-based-behavior}

First, we evaluate four object detectors' confidence scores for the task of identifying an emergency vehicle, comparing a scenario in which the emergency vehicle lighting is on to a scenario when it is off.

\textbf{Experimental Setup:} A 2023 Tesla model 3 front camera was used to record a 30-second video recording (the FPS and resolution of the ADAS
footage can be found in Table \ref{tab:adass}) of a white Honda Civic equipped with blue-red emergency vehicle lighting (purchased on AliExpress\footnote{\label{fn:emrgency-flasher-ali-red}\url{https://aliexpress.com/item/1005005852809515.html}}).  
In the first 15 seconds of the video, the car's emergency vehicle lighting was off, and in the last 15 seconds, it was on (see Fig. \ref{fig:vehicle_detection}).
The video was segmented into frames, and the four object detectors were applied to each frame.
The outputs were used to generate confidence score signals, a time series of the detector confidence score as a function of time/frame, for each detector regarding car detection. 

\begin{figure*}[]
  \centering
  \includegraphics[width=0.24\linewidth]{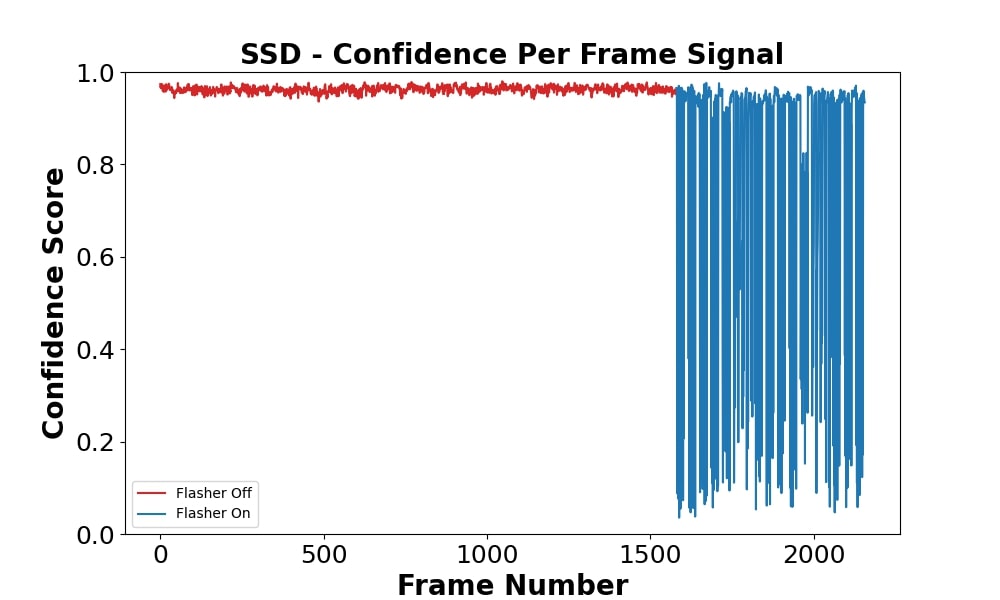}
        \includegraphics[width=0.24\linewidth]{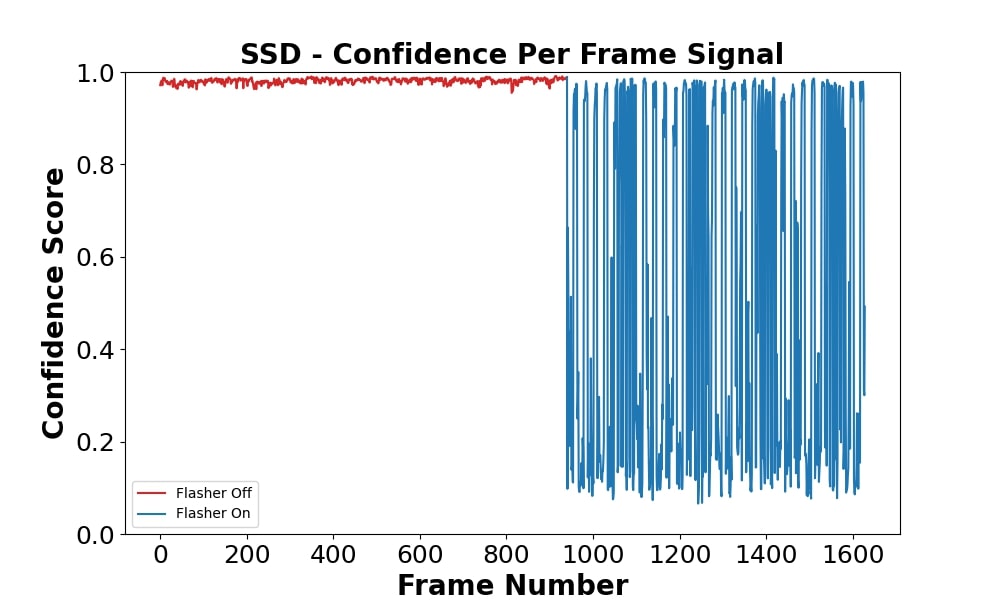}
  \includegraphics[width=0.24\textwidth]{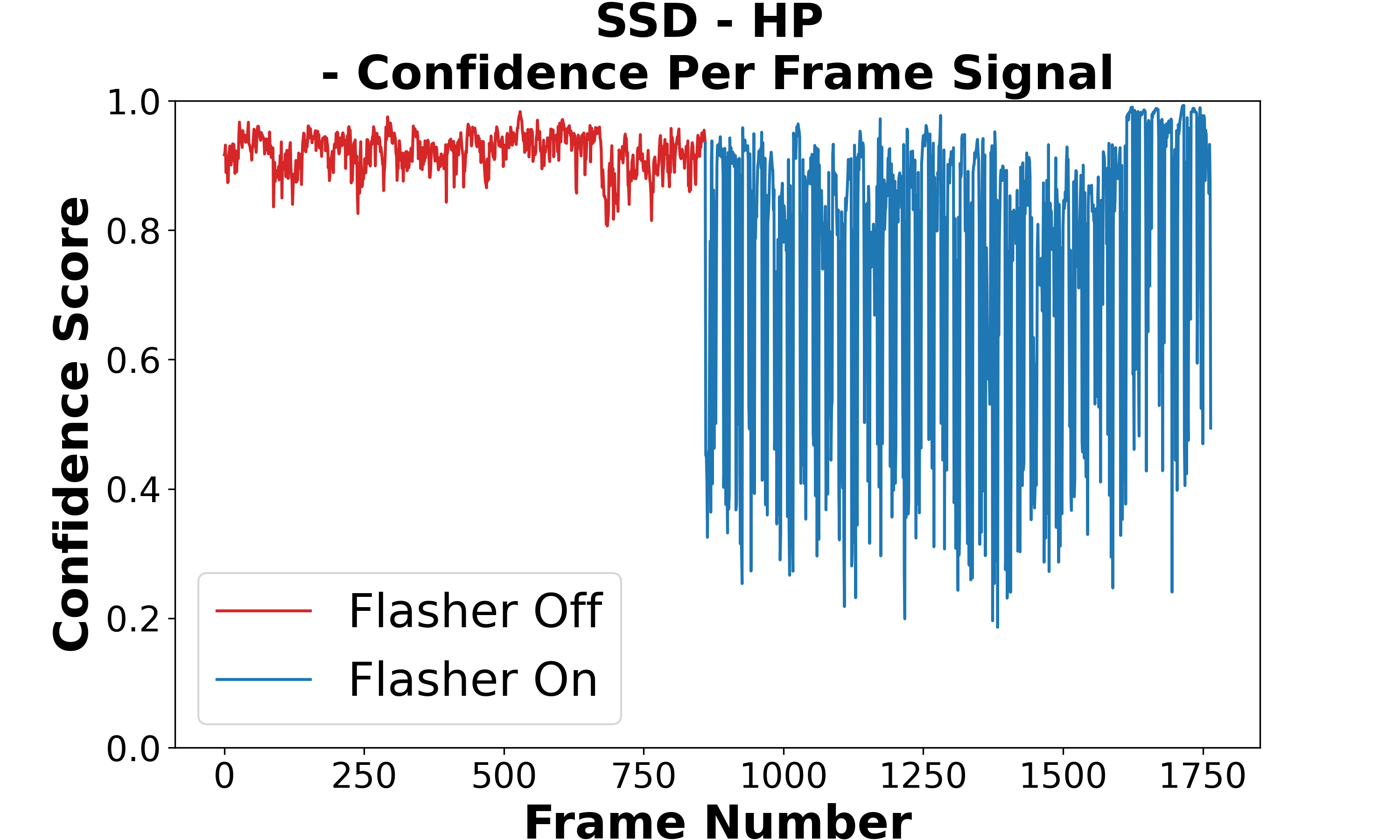}
  \includegraphics[width=0.24\textwidth]{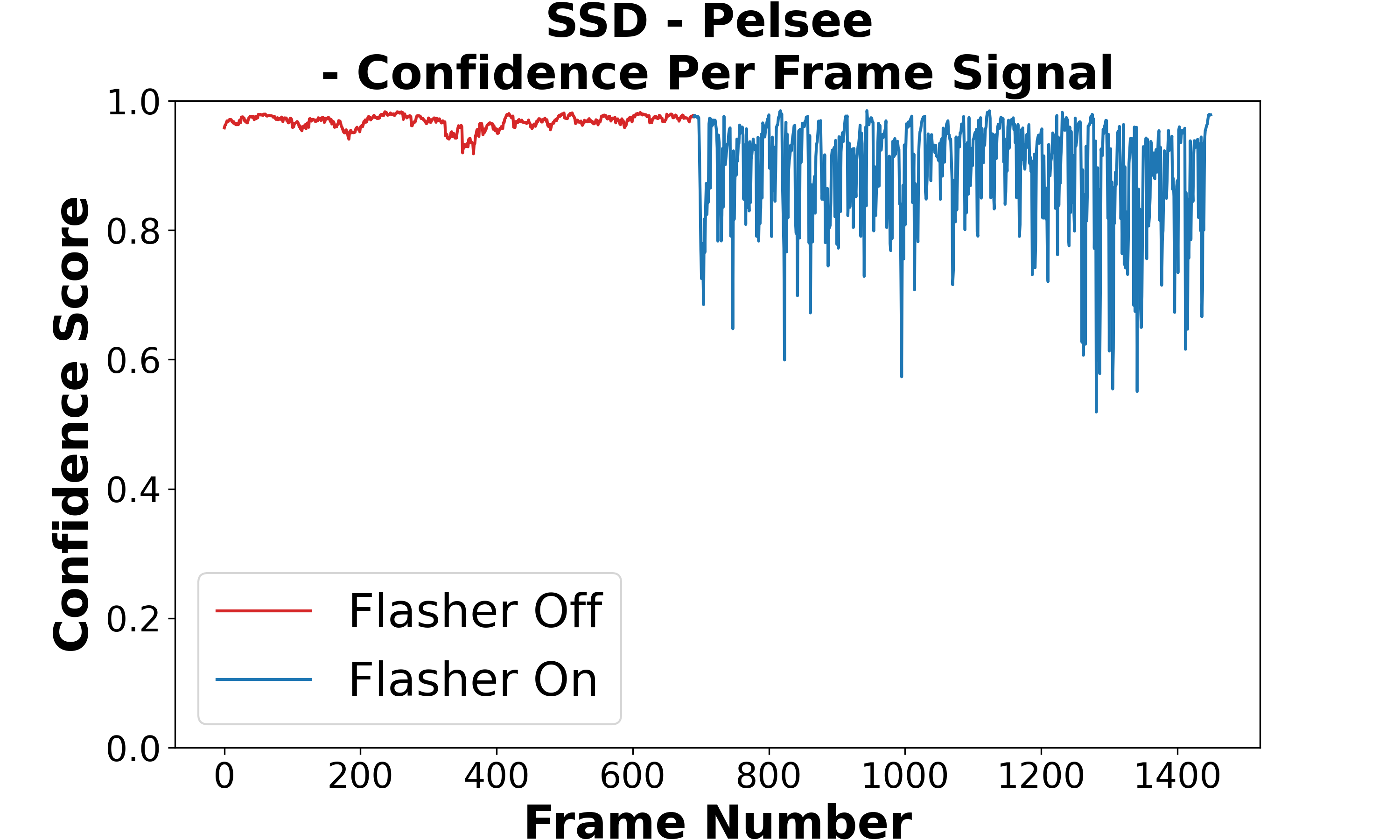} 
\includegraphics[width=0.24\textwidth]{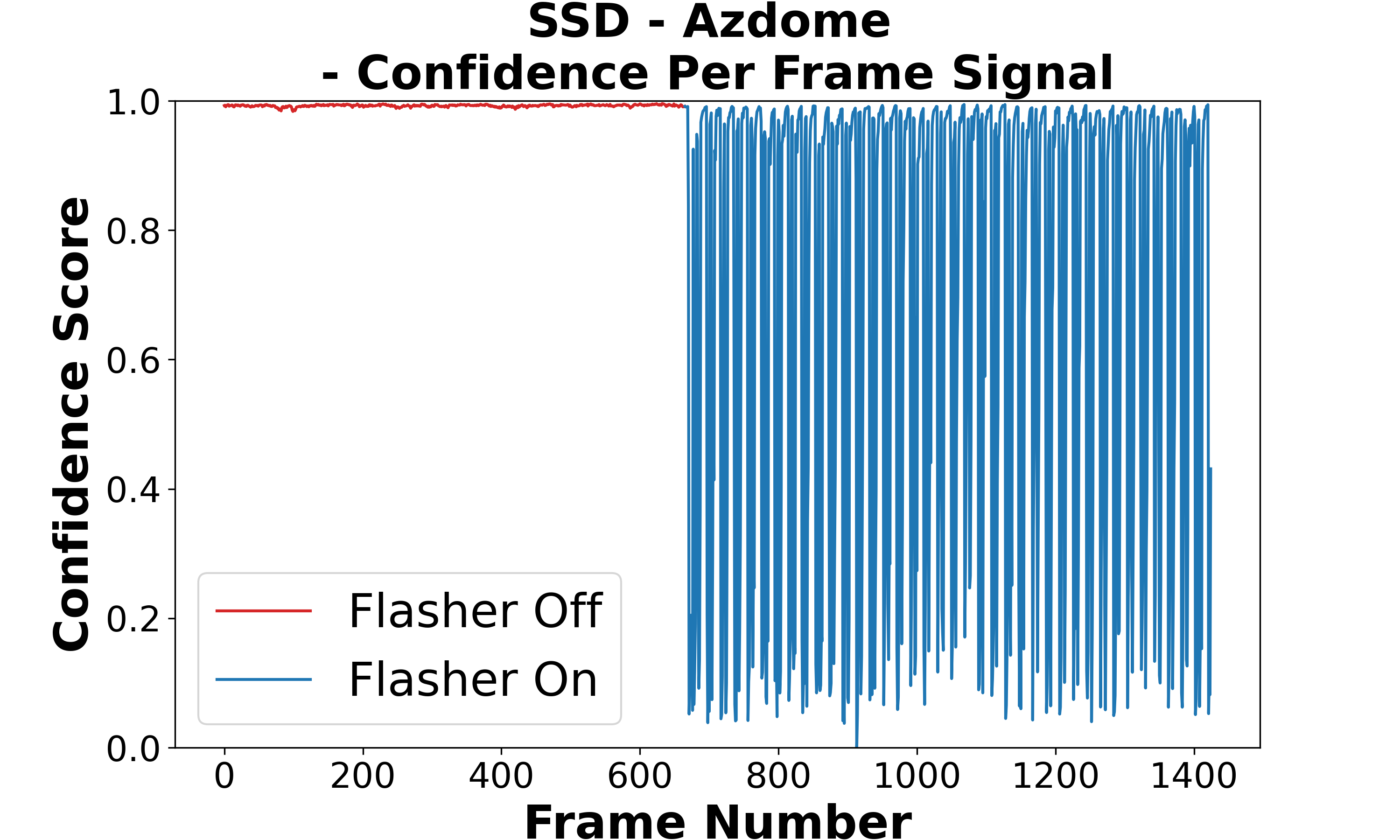}
\includegraphics[width=0.24\textwidth]{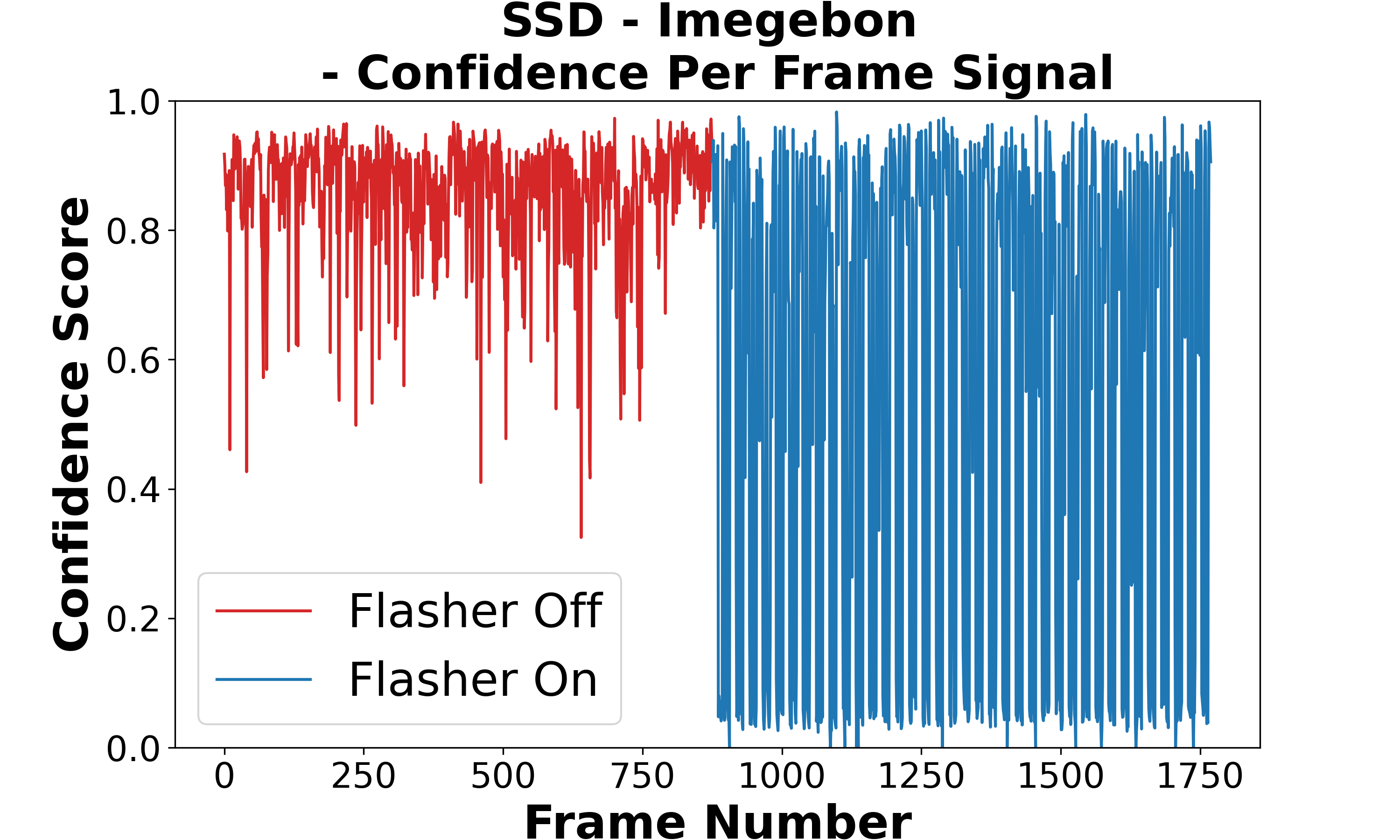} 
\includegraphics[width=0.24\textwidth]{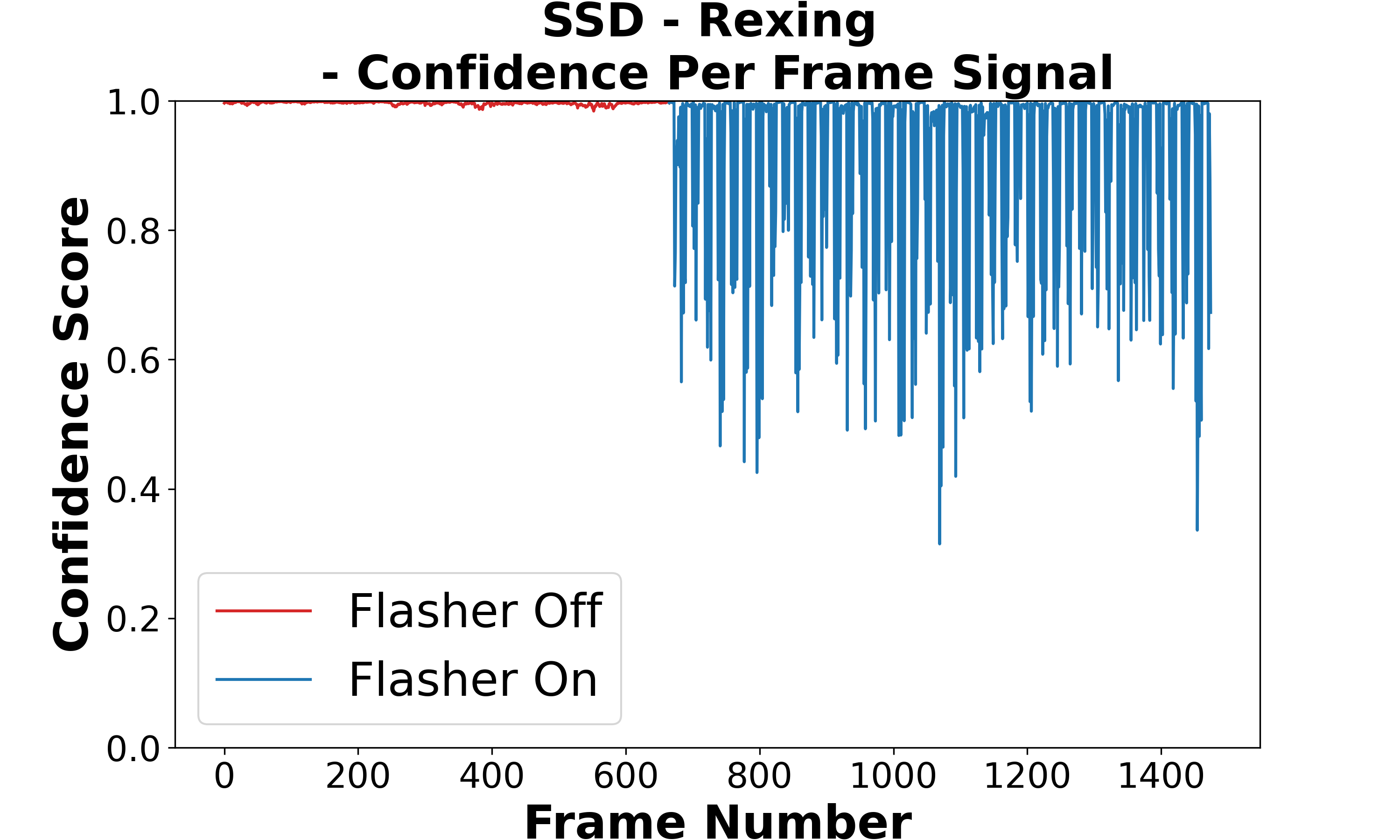}
\includegraphics[width=0.24\textwidth]{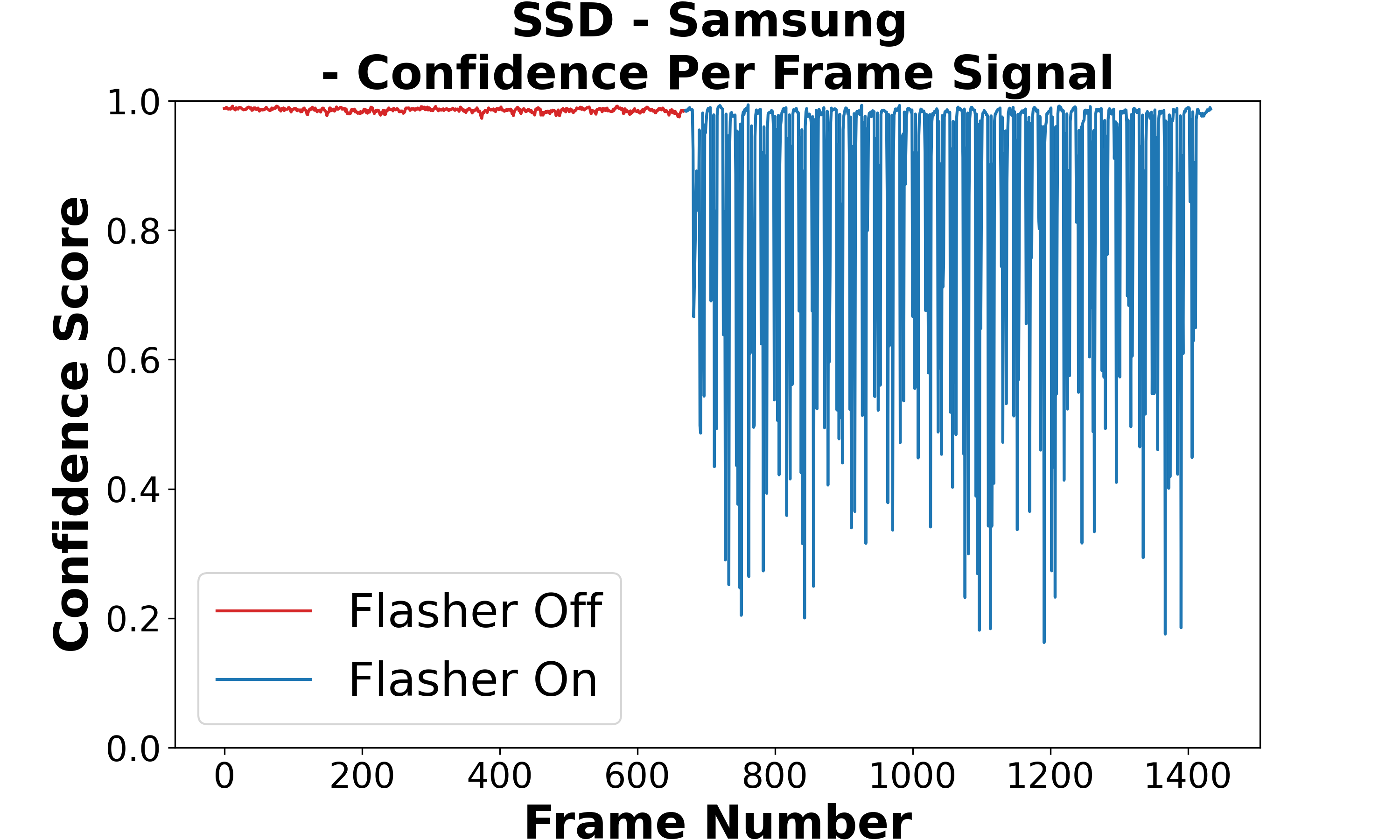}
\caption{Comparison of the confidence score of SSD per frame for videos recorded by different ADASs and a smartphone.}
    \label{fig:adas_analysis_ssd}
\end{figure*}

\textbf{Results:}
The results of YOLO, SSD and Faster R-CNN are presented in Fig. \ref{fig:all-ods}, the result of RetinaNet is presented in the appendix in Fig. \ref{Fig:retina_showcase}. 
As can be seen from the results, in the first half of the video (when the emergency vehicle lighting was off), the detectors' confidence remained stable when detecting a car with confidence higher than 0.9. 
In contrast, in the second half of the video (when the emergency vehicle lighting was on), the scores of the object detector fluctuated, down to values below 0.4 in YOLOv9, Faster R-CNN, and SSD. 

\vspace{2mm}
\begin{insight} \label{insight:1}
The activation of emergency vehicle lighting creates a phenomenon, which we term \EC\, that causes the confidence score of object detectors regarding a detected object to fluctuate, within a wide score range, with the score dipping below a reasonable detection threshold in some cases.
\end{insight}
\vspace{2mm}

As can be seen in Fig. \ref{fig:all-ods}, similar behavior was observed in all the examined object detectors, indicating a systemic issue in object detectors rather than an isolated case. 


\vspace{0.5em}
\begin{insight} \label{insight:3}
The \EC\ phenomenon is \textcolor{red}{consistent across different object detectors}, but the score fluctuation ranges differ depending on the object detector.
\end{insight}

\subsubsection{Camera Analysis}

Next, we evaluate footage from various cameras and their effect on the confidence scores of an object detector for the task of identifying an emergency vehicle, both when the emergency vehicle lighting is on and when it is off. 

\textbf{Experimental Setup:} Seven ADASs (Tesla, "manufacturer C", HP, Pelsee, AZDOME, Imagebon, Rexing) and a smartphone camera (Samsung Galaxy S22 Ultra) were used individually to record a 60-second video recording (the FPS and resolution of the ADAS footage can be found in Table \ref{tab:adass}, and the Samsung Galaxy footage was recorded at 24 FPS at FHD resolution) of a car equipped with blue emergency vehicle lighting (purchased on Amazon\footref{fn:emrgency-flasher-amazon-blue-2}). 
The experimental setup is presented in Fig. \ref{fig:commercial-adas}. 
In the first 30 seconds of each recording, the car's emergency vehicle lighting was off, and in the last 30 seconds, it was on (see Fig. \ref{fig:vehicle_detection}).
Each recorded video was segmented into frames, and the four object detectors were applied to each frame.
The outputs were used to generate confidence score signals, a time series of the detector confidence score as a function of time/frame, for each ADAS/smartphone and object detector regarding car detection.

The results for SSD are presented in Fig. \ref{fig:adas_analysis_ssd}. 
As can be seen from the results, in the first half of most of the recordings (when the emergency vehicle lighting was off), the detectors' confidence remained more stable when detecting a car than when turning on the emergency vehicle lighting afterwards. 
In contrast, in the second half of the recordings (when the emergency vehicle lighting was on), the confidence score of the object detector fluctuated, down to values below 0.4 in the case of the ADASs and the smartphone. 

\vspace{2mm}
\begin{insight} \label{insight:adas_insight}
The \EC\ phenomenon is \textcolor{red}{consistent across seven different ADASs}, with the score fluctuation ranges when the emergency vehicle lighting is off differing depending on the ADAS recording the video frames.
\end{insight}

\begin{figure*}[t]
  \centering
    \includegraphics[width=0.32\textwidth]{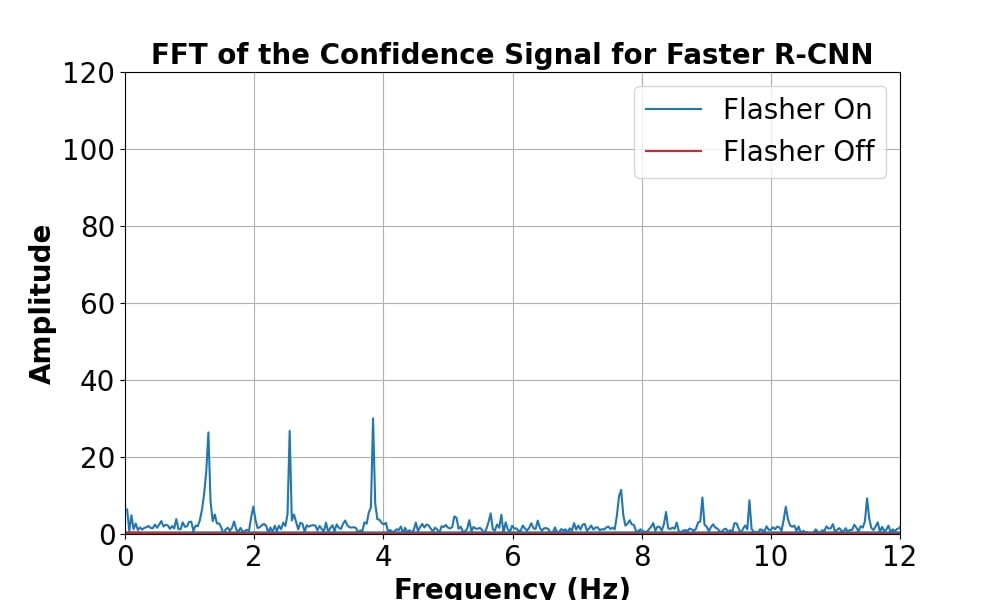}
  \includegraphics[width=0.26\textwidth]{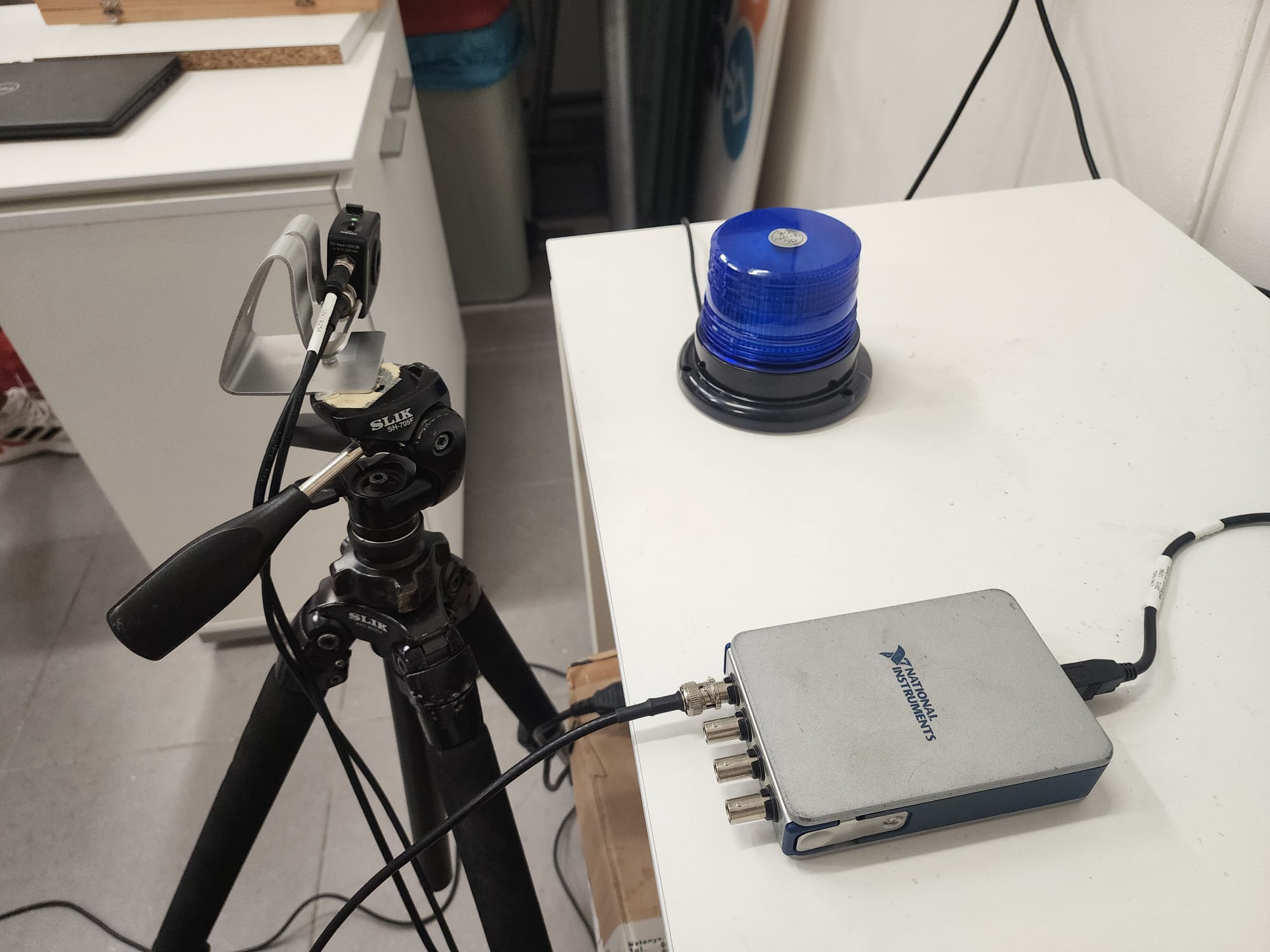}
\includegraphics[width=0.29\textwidth]{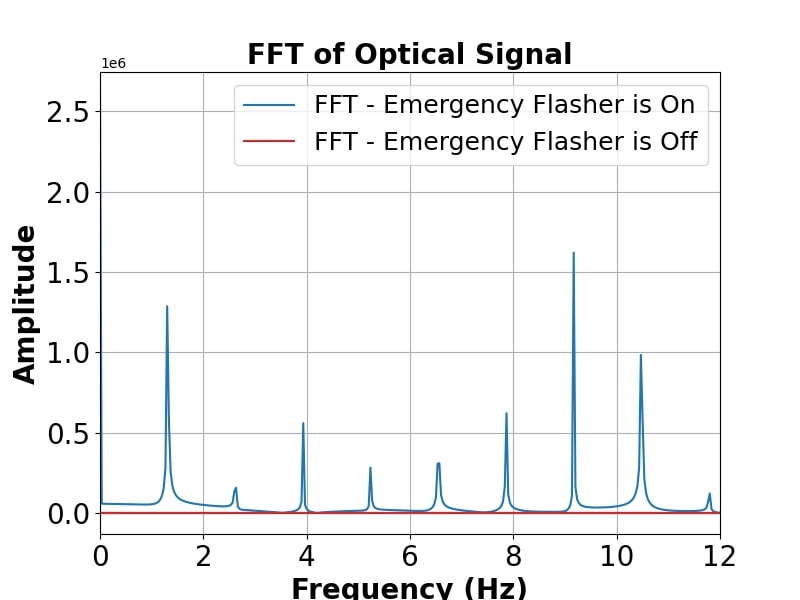}
\caption{Left: Extracted FFT graph from the confidence signal of Faster R-CNN recorded by AZDOME when the emergency vehicle lighting is on. A peak around 1.3 Hz can be observed. Center: Experimental setup; a photodiode was directed at blue emergency vehicle lighting in a lab setting. Right: An FFT graph of the recorded optical signal with a peak around 1.3 Hz.}
\vspace{-1.5em}
    \label{fig:photodiode}
\end{figure*}

\subsubsection{RGB-Based Analysis}

Next, we analyze the impact of activated emergency vehicle lighting on a captured frame.

\textbf{Experimental Setup:} We extracted two frames from the previously recorded 60-second video recording that was obtained by Tesla: (1) a random frame from the first 30 seconds of the video recording (when the emergency vehicle lighting was off) in which a car was detected by SSD with a confidence of 0.98, and (2) one frame from the last 30 seconds (when the emergency vehicle lighting was on) in which a car was detected by SSD with a confidence of 0.03. 

\textbf{Results:} We computed the tonal distribution for the two extracted frames. 
Fig. \ref{fig:vehicle_detection} presents an analysis of the normalized RGB distribution of the colors of the car when the emergency vehicle lighting is on and off. 
As can be seen, there is a clear difference in the color distributions of the two frames resulting from the blue and red lights emitted from the emergency vehicle lighting. 

\vspace{2mm}
\begin{insight} \label{insight:3.5}
The flare added by the emergency vehicle lighting changes the distribution of the colors of the car in the captured frame, which changes the confidence of the object detector regarding the car.
\end{insight}

\subsubsection{Spectral Analysis}
\label{sec:flasher-impact}

Next, we show that the presence of activated emergency vehicle lighting determines the object detector behavior.

\textbf{Experimental Setup:} We used AZDOME to obtain a 60-second video recording from blue emergency vehicle lighting 
(purchased on Amazon\footnote{\label{fn:emrgency-flasher-amazon-blue-2}\url{https://www.amazon.com.au/Xprite-Rotating-Revolving-Magnetic-Emergency/dp/B07B7YYL32}}).
In the first 30 seconds of the video, the car's emergency vehicle lighting was off, and in the last 30 seconds of the video it was on.
The video was segmented into frames, and the four object detectors were applied to each frame.
The outputs were used to generate confidence score signals (\textit{i.e.,} detector confidence as a function of time) for each detector regarding car detection.
For each of the four signals extracted from the object detector's confidence, we computed two fast Fourier transforms (FFTs), one from the first 30 seconds of the signal (when the emergency vehicle lighting is off) and one from the last 30 seconds of the signal (when the emergency vehicle lighting is on).


\begin{figure*}[]
  \centering
  \begin{minipage}[b]{0.32\textwidth}
    \includegraphics[width=\textwidth]{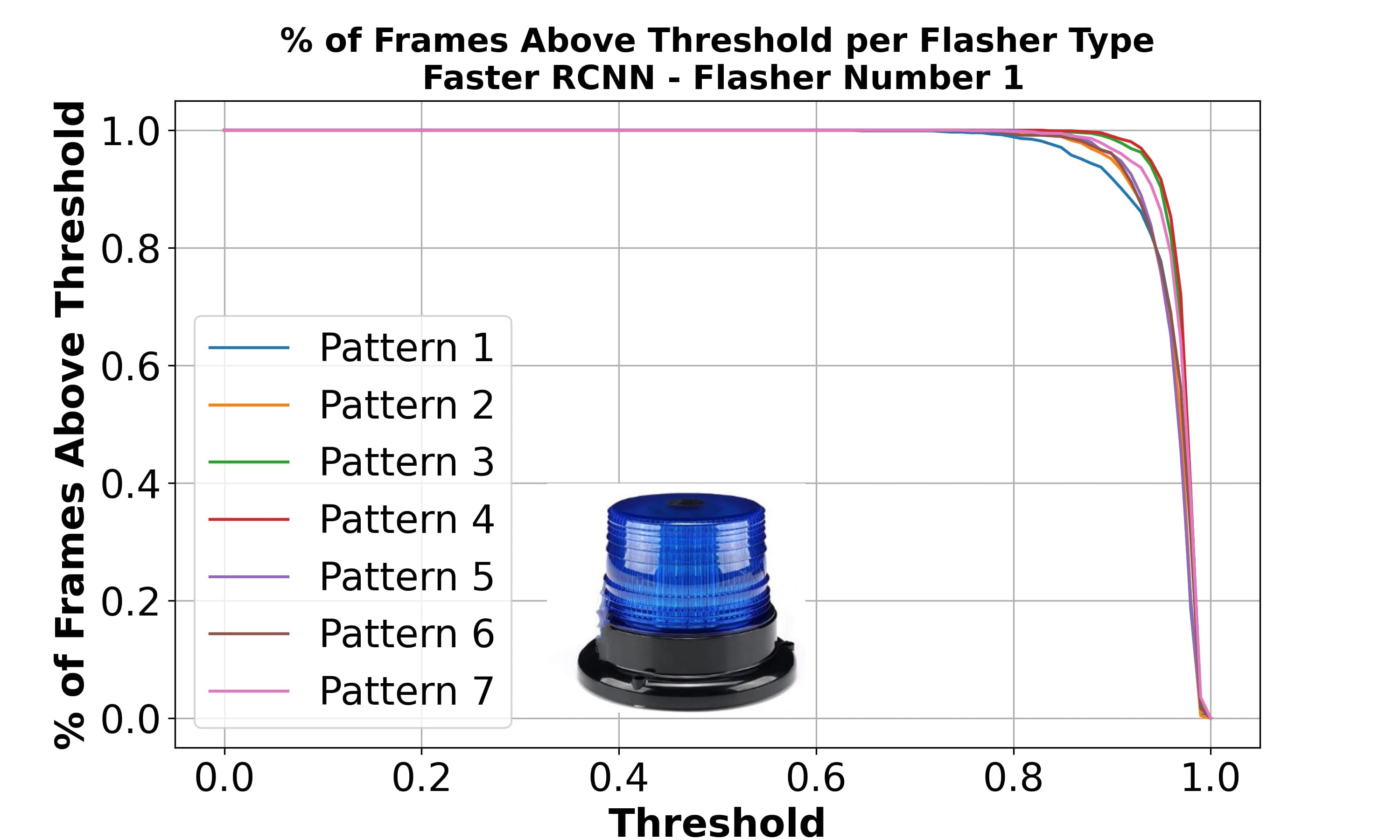}    
   \includegraphics[width=\textwidth]{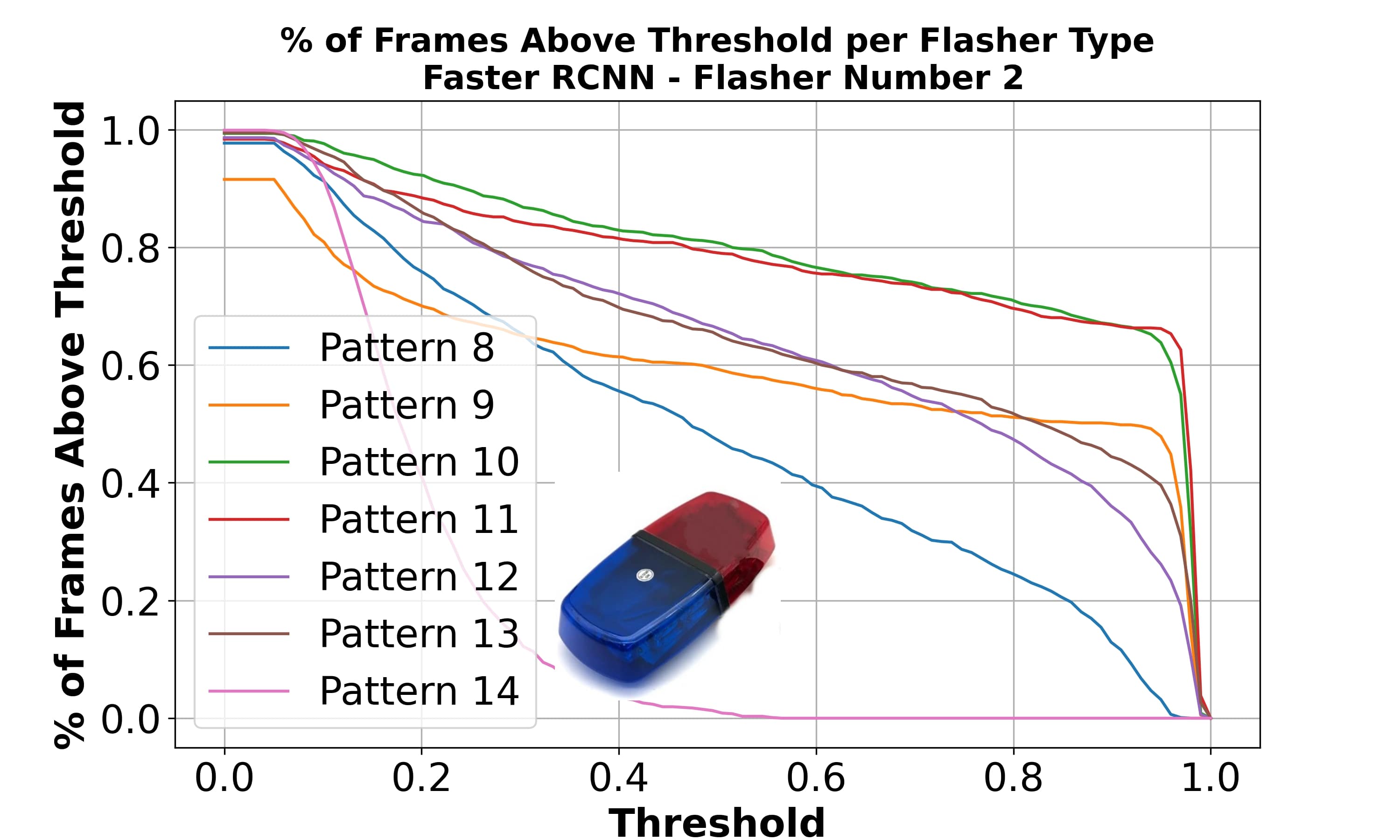} 
    \caption{The percentage of frames in which a car was detected by Faster R-CNN with confidence higher than the threshold (in the presence of emergency vehicle lighting).}
\label{fig:threshold-vs-percentage}
  \end{minipage}
  \hspace{0.1cm}
  \begin{minipage}[b]{0.31\textwidth}
    \includegraphics[width=\textwidth]{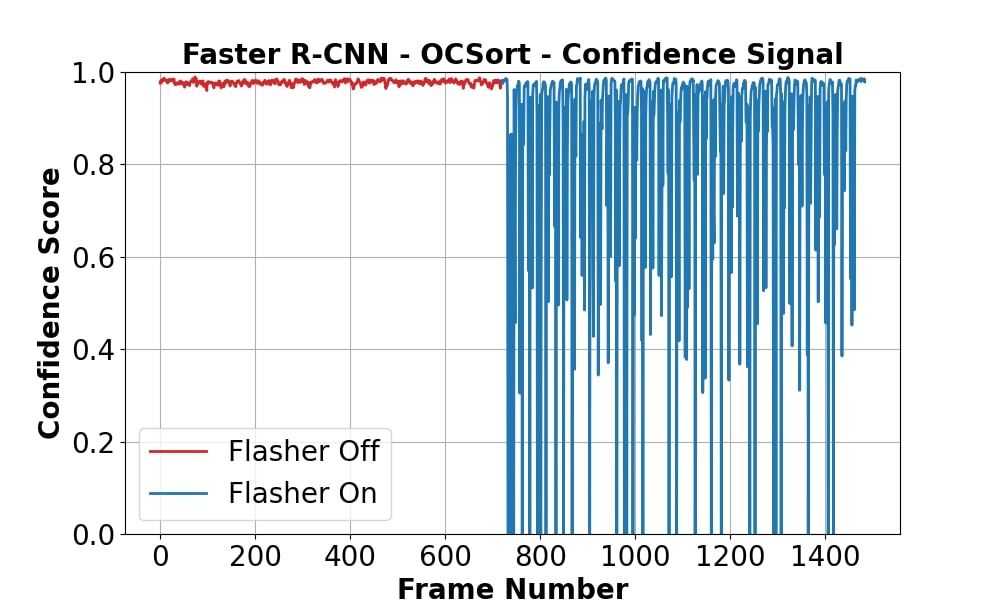}   
    \includegraphics[width=\textwidth]{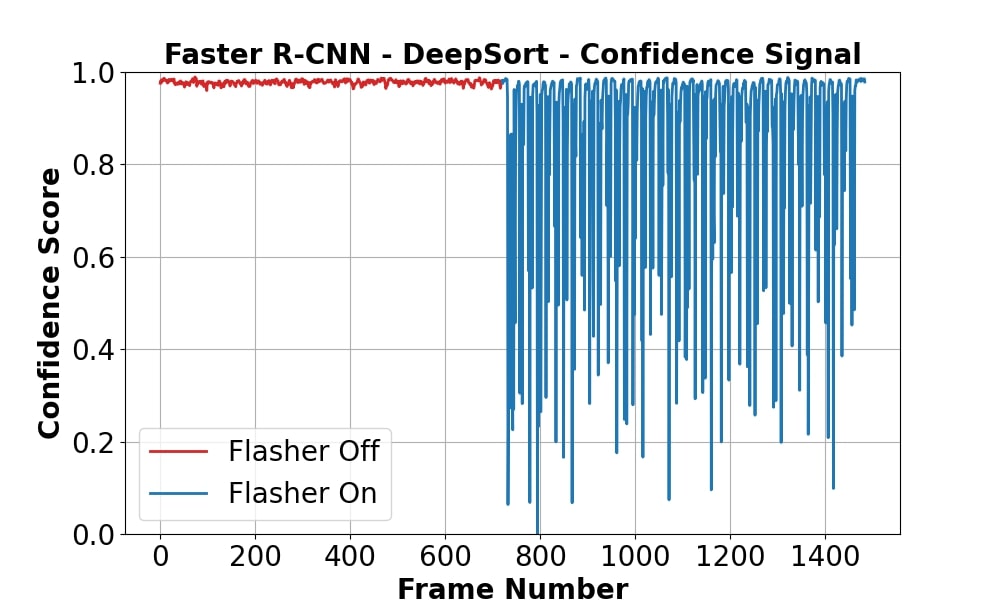} 
    \caption{The confidence score signals of two object tracking models (OCSort and DeepSort) applied on the Faster R-CNN object detector}
 \label{fig:object-tracking}
  \end{minipage}
  \hspace{0.1cm}
  \begin{minipage}[b]{0.33\textwidth}
    \includegraphics[width=\textwidth]{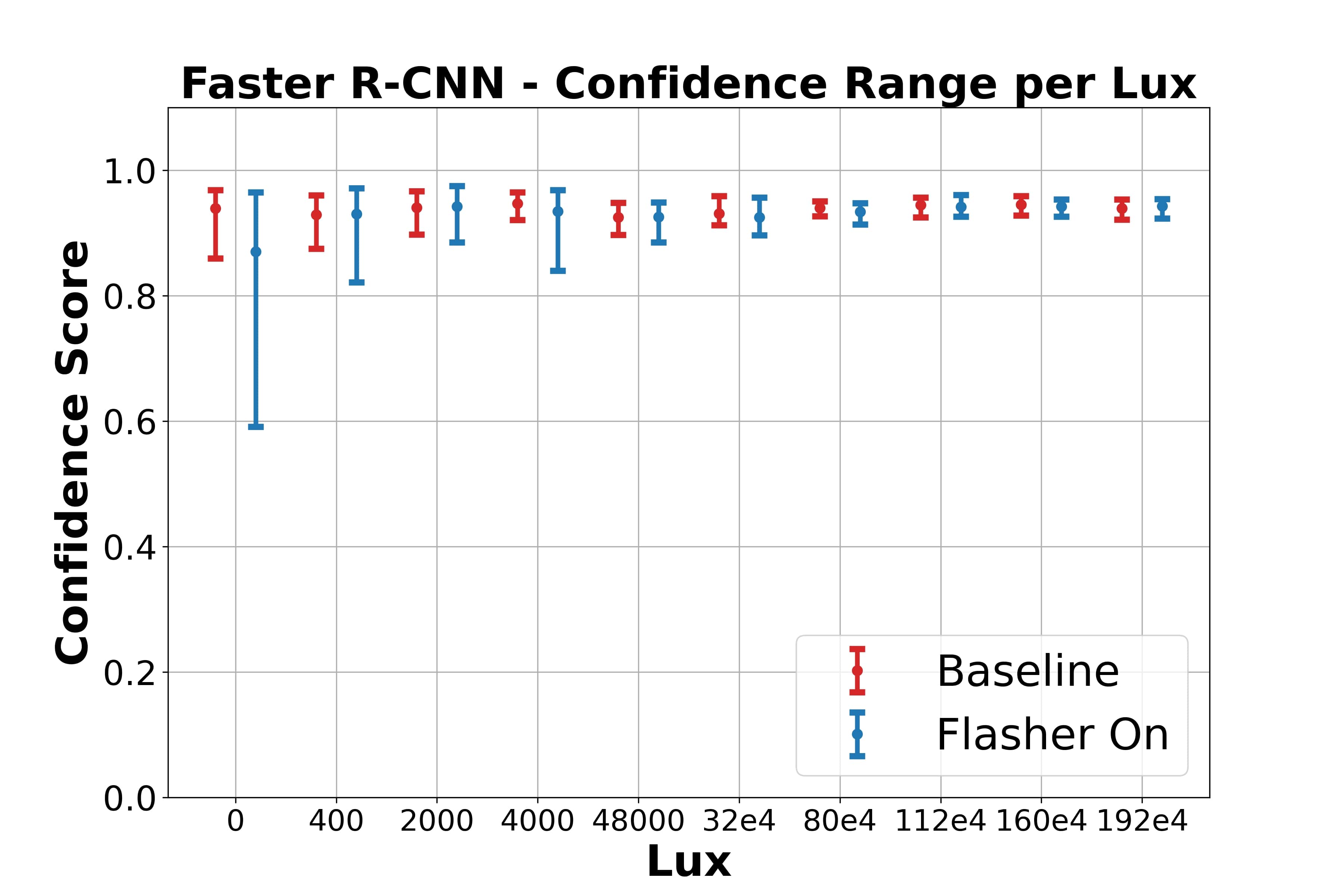}
    \includegraphics[width=0.48\textwidth]{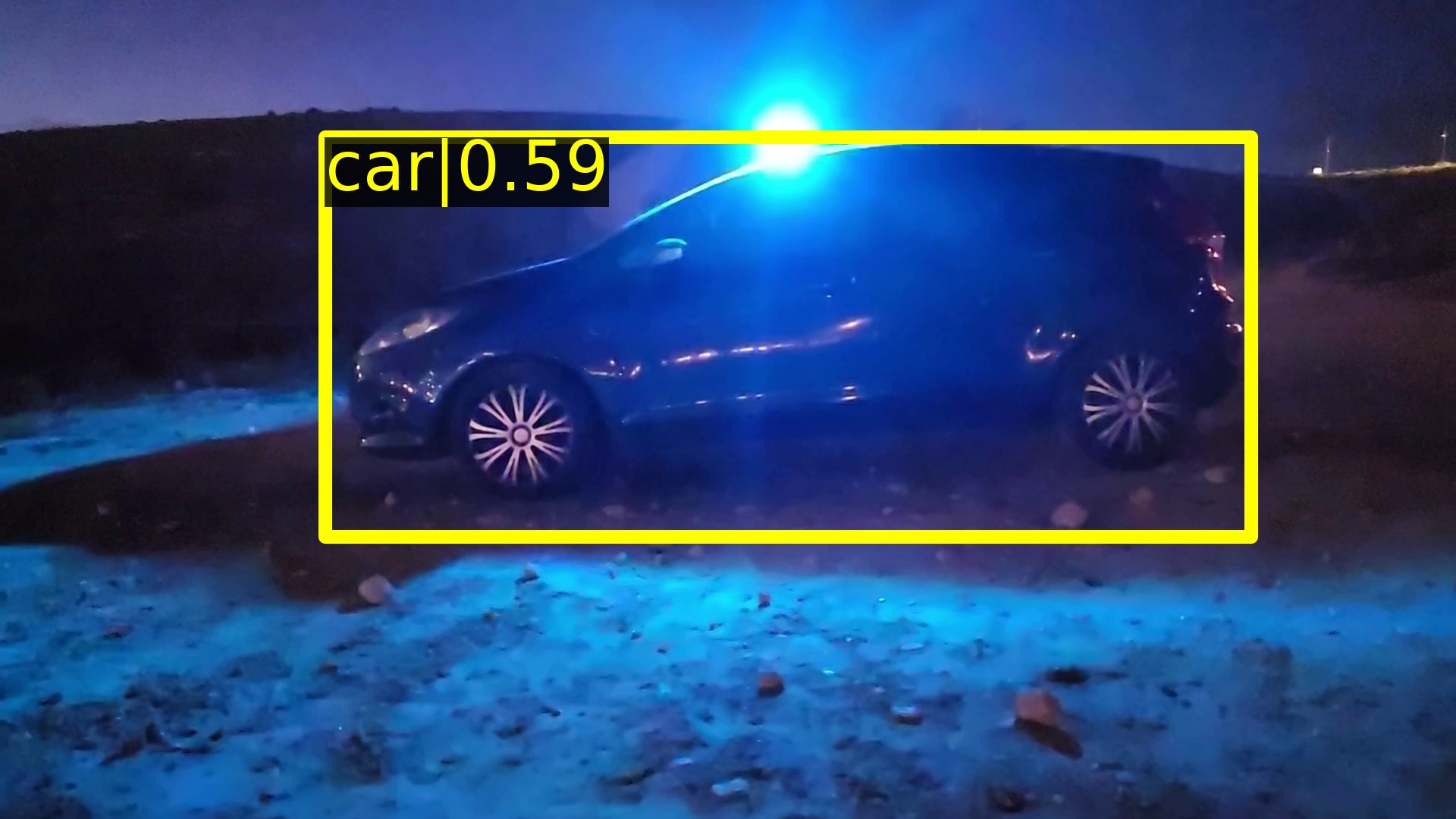}
    \includegraphics[width=0.48\textwidth]{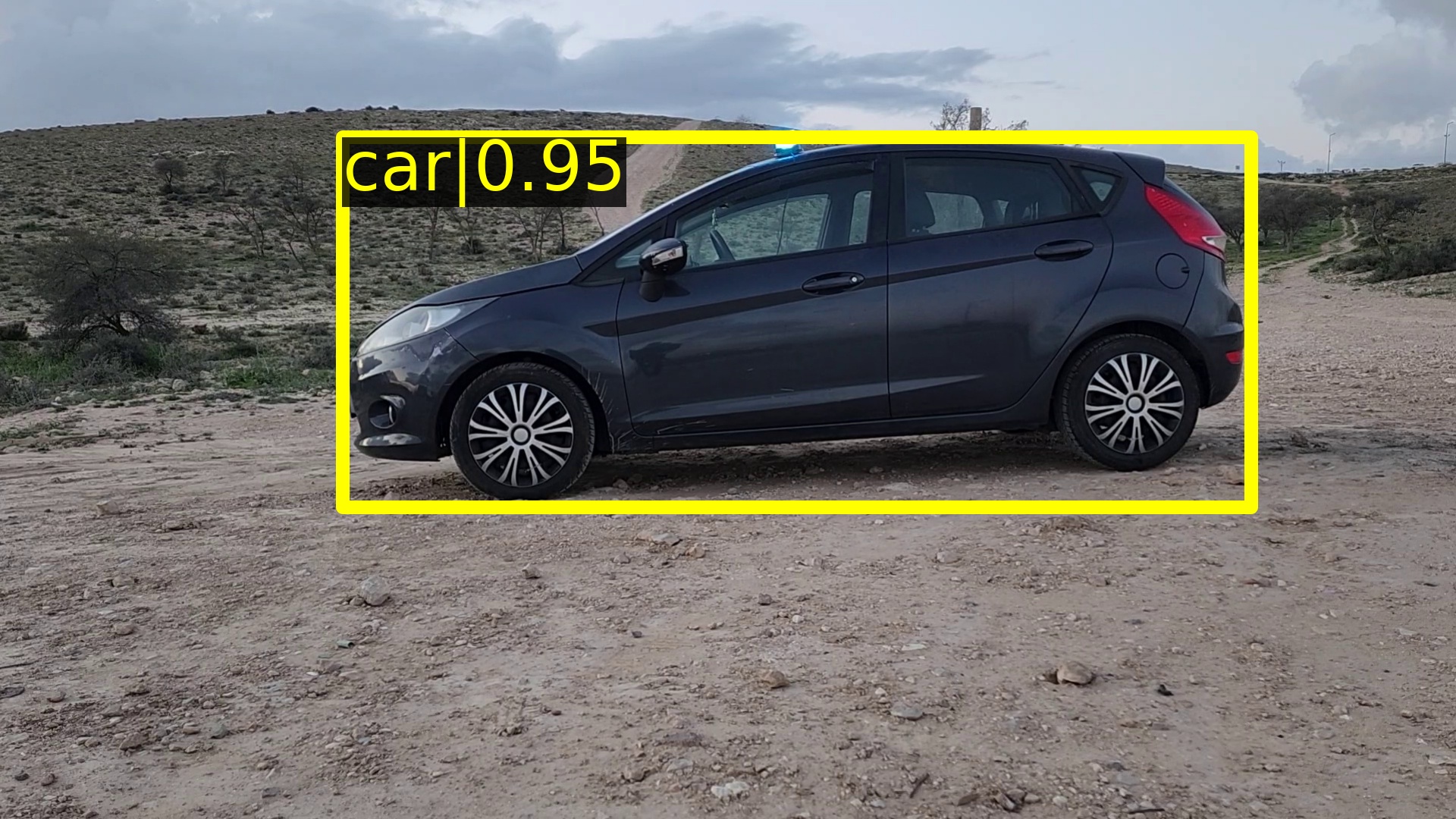}
    \caption{Top: Faster R-CNN confidence score signal range for each examined lux value. Bottom: There is a difference in the confidence score of a detected car of 0.36: in 0 lux a car is detected with a score of 0.59 (left) and in 160,000 lux a car is detected with a score of 0.95 (right).}
    \label{fig:analysis-ambiemt-light}
  \end{minipage}
  \end{figure*}

\textbf{Results:} 
The results of the FFT graphs extracted from the detections of Faster R-CNN are presented in Fig. \ref{fig:photodiode} (left), and the results for YOLO, SSD, and RetinaNet are presented in Fig. \ref{fig:ffts-all-ods} in the appendix. 
All FFT graphs show a peak around 1.3 Hz when the emergency vehicle lighting is activated which does not appear when the emergency vehicle lighting is off.

In an attempt to explain the behavior that appears in the four FFT signals, we conducted an additional experiment.

\textbf{Experimental Setup:} We directed a Thorlabs PDA100A2 photodiode (an optical sensor used to convert the intensity of captured light to electrical voltage) in a lab setup and captured optical measurements for 60 seconds: in the first 30 seconds of the experiment, the emergency vehicle lighting was off and in the last 30 seconds the emergency vehicle lighting was on. 
The voltage (optical measurements) was obtained from the photodiode using a 24-bit ADC NI-9234 card that was connected to a laptop and was processed in a LabVIEW script that we wrote. 
The experimental setup is presented in Fig. \ref{fig:photodiode} (middle).

\textbf{Results.} Fig. \ref{fig:photodiode} (right) presents the application of FFT on the optical measurements.
We can see a peak around 1.3 Hz in the FFT graph when the emergency vehicle lighting is activated (which does not appear when the emergency vehicle lighting is off), presenting the same behavior we observed when analyzing the confidence signal of the four object detectors. 

\vspace{2mm} 
\begin{insight} \label{insight:2}
The frequency of the pattern of the emergency vehicle lighting determines/dictates the behavior of the object detector's confidence in detecting the car over time.
\end{insight}



\subsubsection{Analysis of Detection Loss}

Next, we analyzed the detection loss (the rate at which the object detector failed to detect a car in the frames) due to activated emergency vehicle lighting.

\textbf{Experimental Setup.} We used two emergency vehicle lighting setups: (1) blue emergency vehicle lighting\footref{fn:emrgency-flasher-amazon-blue-2}, and (2) blue-red emergency vehicle lighting\footref{fn:emrgency-flasher-ali-red} (see Fig. \ref{fig:commercial-adas}).
Each of the emergency vehicle lighting setups has settings that enable seven different light patterns.
We mounted each of the emergency vehicle lighting setups to the roof of a car and used AZDOME to record 30-second videos for each of the 14 light patterns (seven from each emergency vehicle lighting setup). 
We applied the four object detectors to the frames of each of the videos and calculated the percentage of frames (0-1.0) in which the car was detected beyond a given threshold (0-1.0).


  

\textbf{Results.} The results of Faster R-CNN for detecting cars in the two videos are presented in Fig. \ref{fig:threshold-vs-percentage}, while the results for YOLOv9, SSD, and RetinaNet are presented in Fig. \ref{Fig:retina-two-flashers} in the appendix. 
As can be seen from the results presented in Fig. \ref{fig:threshold-vs-percentage}, the two emergency vehicle lighting setups yield different results for Faster R-CNN. 
For blue emergency vehicle lighting, the detection loss of Faster R-CNN is marginal up to a threshold of 0.9, meaning that a car is detected in 90\% of the frames at a detection threshold of 0.9.
On the other hand, the detection loss of the red-blue emergency vehicle lighting is significant as 20\% of the scores of the cars in the frames are detected with a confidence lower than 0.4. 

\vspace{2mm} 
\begin{insight} \label{insight:1.1}
The \EC\ phenomenon presents a tradeoff between a high true positive rate of an object detector (which could be satisfied with low detection thresholds) and the object detector's accuracy/low false positive rate (which could be satisfied with high detection thresholds).
\end{insight}

\vspace{2mm} 
\begin{insight} \label{insight:2.1}
The flare added from the emergency vehicle lighting changes the tonal distribution of the car in the frame captured by the video camera. 
As a result, the behavior and performance of an object detector are affected by the pattern and frequency of the emergency vehicle lighting. This causes detection losses in object detectors that are unique and dictated by the emergency vehicle lighting's pattern/frequency.
\end{insight}

\subsection{Analysis of Object Tracker Behavior}

We note that object trackers are employed one layer above the object detectors in some ADASs \cite{gruyer2017perception} and are used to identify the paths of detected objects. These paths are provided as input to the decision-making algorithms (which are employed one layer on top of the trackers).
One might argue the analysis conducted so far lacks the full picture, because it does not consider the use of object trackers, which might compensate for an object detector's detection loss in the presence of activated emergency vehicle lighting.
In the next experiment, we analyze the behavior of object trackers in response to the \EC\ phenomenon.

\textbf{Experimental Setup.} We applied three object trackers (OCSORT \cite{OCSort}, DeepSORT \cite{DeepSort}, and ByteTrack \cite{ByteTrack}) to the previously recorded 60-second video captured by AZDOME.
Object trackers rely on a preliminary step of object detection, so we applied the three object trackers with the Faster R-CNN.


\textbf{Results.} 
The results of Faster R-CNN, with object trackers DeepSort and OCSort presented in Fig. \ref{fig:object-tracking} while the results of ByteTrack object tracker can be found in the appendix in Fig. \ref{Fig:bytetrack_no_mot}. 
As can be seen, the object tracking algorithms do not compensate for Faster R-CNN's detection loss.
The same behavior is seen in response to the activated emergency vehicle lighting regardless of whether an object tracker was used or not. 
Thus, the intended behavior of object trackers (associating new detections of objects with previous detections) does not mitigate the object detectors' detection loss in the presence of activated emergency vehicle lighting. 
\vspace{1mm}
\begin{insight} \label{insight:2.2}
Object tracking fails to compensate for object detectors' detection loss in the presence of activated emergency vehicle lighting.
\end{insight}

\subsection{Effect of Ambient Light} 
Here we examine how the level of ambient light affects object detectors' confidence in recognizing a car.

\textbf{Experimental Setup:} We positioned a Samsung Galaxy S22 Ultra five meters away from a grey Ford Fiesta with emergency vehicle lighting mounted on its roof. 
We recorded 10 different 30-second videos of the car, with a 15-minute interval between the videos, allowing us to examine the effect of progressively diminishing light levels (transitioning from daylight to darkness). 
We used a professional lux meter (Extech HD450 \cite{Extech}) to measure the lux level. 
We applied the four pretrained object detectors to the frames from each video and measured the car's detection confidence range as a function of the video's lux level. 
This experimental setup was repeated twice: first when the emergency vehicle lighting was off and a second time when it was on.

\textbf{Results:} The results of this analysis for the Faster R-CNN detector are presented in Fig. \ref{fig:analysis-ambiemt-light}. 
The results for the other object detectors (YOLOv3, SSD, RetinaNet) are presented in Fig. \ref{Fig:ambient_appendix} in the appendix. 
As can be seen, for all four object detectors, detection confidence remained high when either: (1) there was sufficient ambient light, or (2) the emergency vehicle lighting was off. 
The results also show that the confidence range changed depending on the environmental conditions, indicating less stability in the object detectors' confidence, particularly when: (1) ambient light levels were low, and (2) the emergency vehicle lighting was on.
The frames associated with the lowest confidence scores in the videos recorded when the ambient light was 0 lux and 160,000 lux are presented in Fig. \ref{fig:analysis-ambiemt-light} and show that the emergency vehicle lighting significantly affects the image captured in darkness with respect to the image captured in light.

\vspace{2mm}
\begin{insight} \label{insight:2.3}
The \EC\  phenomenon creates wider confidence score ranges in dark environments, leading to less stable object detector performance.
\end{insight}

In addition, we further investigate the effect of an ADAS' camera settings, the effect of various characteristics of the emergency vehicle (color, orientation, etc.), and the influence of distance between the camera and the emergency vehicle lighting, on the \EC\ phenomenon. 
These analyses can be found in the appendix.
\begin{figure*}[h]
       \begin{minipage}{0.49\textwidth}
          \includegraphics[width=0.48\textwidth]{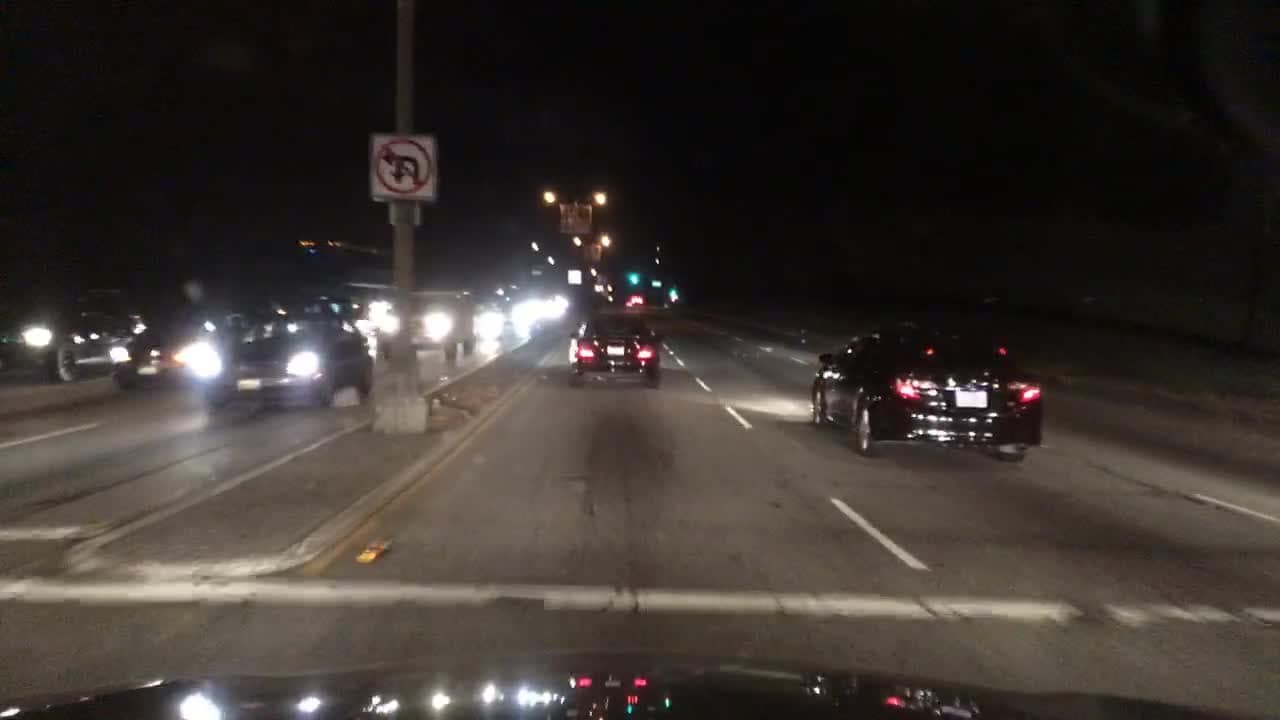}  
     \includegraphics[width=0.48\textwidth]{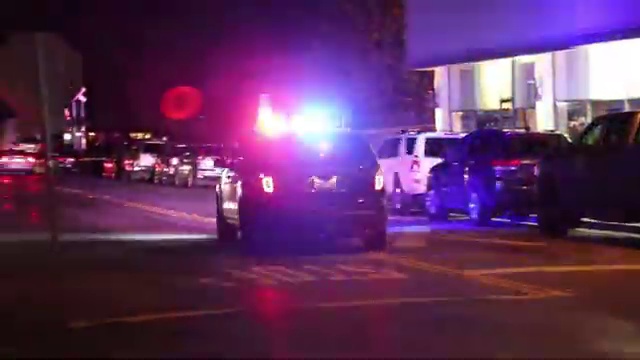}
      \caption{Samples taken from two authentic collected datasets; a sample from the Berkeley dataset (left) and YouTube dataset (right).} 
      \label{fig:real-world-ds}
    \end{minipage} 
    \hspace{0.1em}
    \begin{minipage}{0.49\textwidth}
     \includegraphics[width=0.48\textwidth]{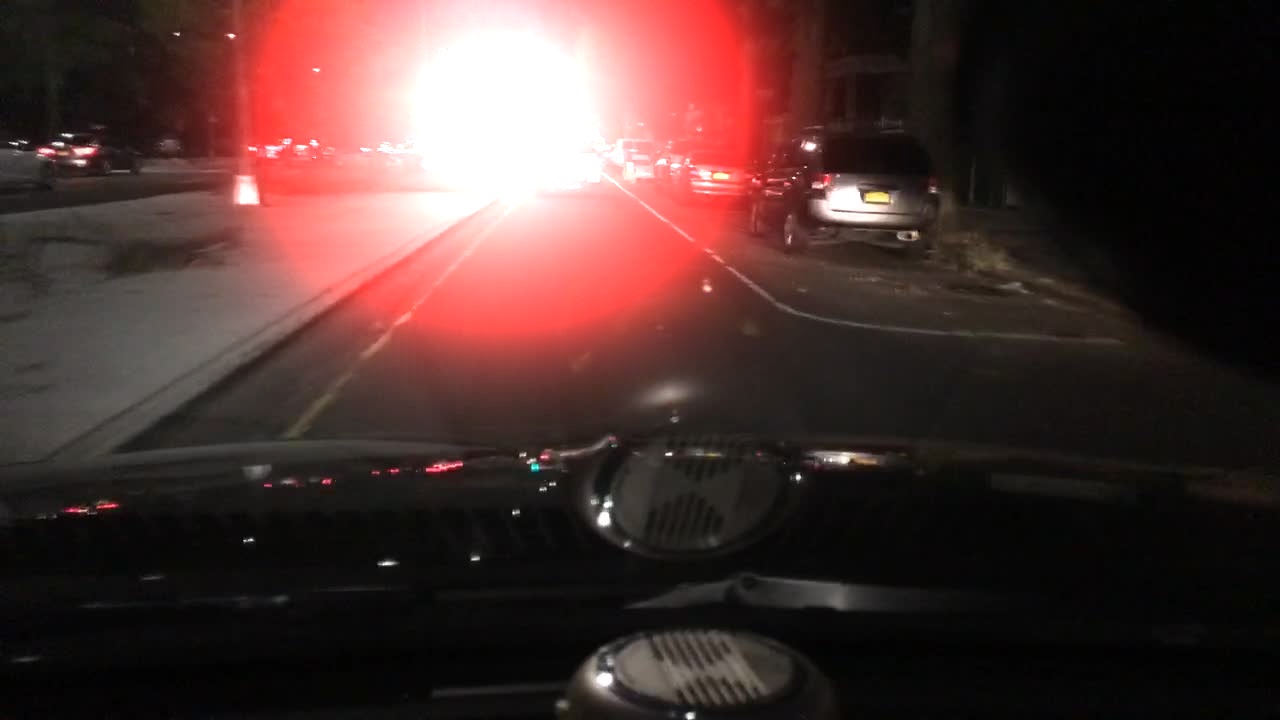}    
    \includegraphics[width=0.48\textwidth]{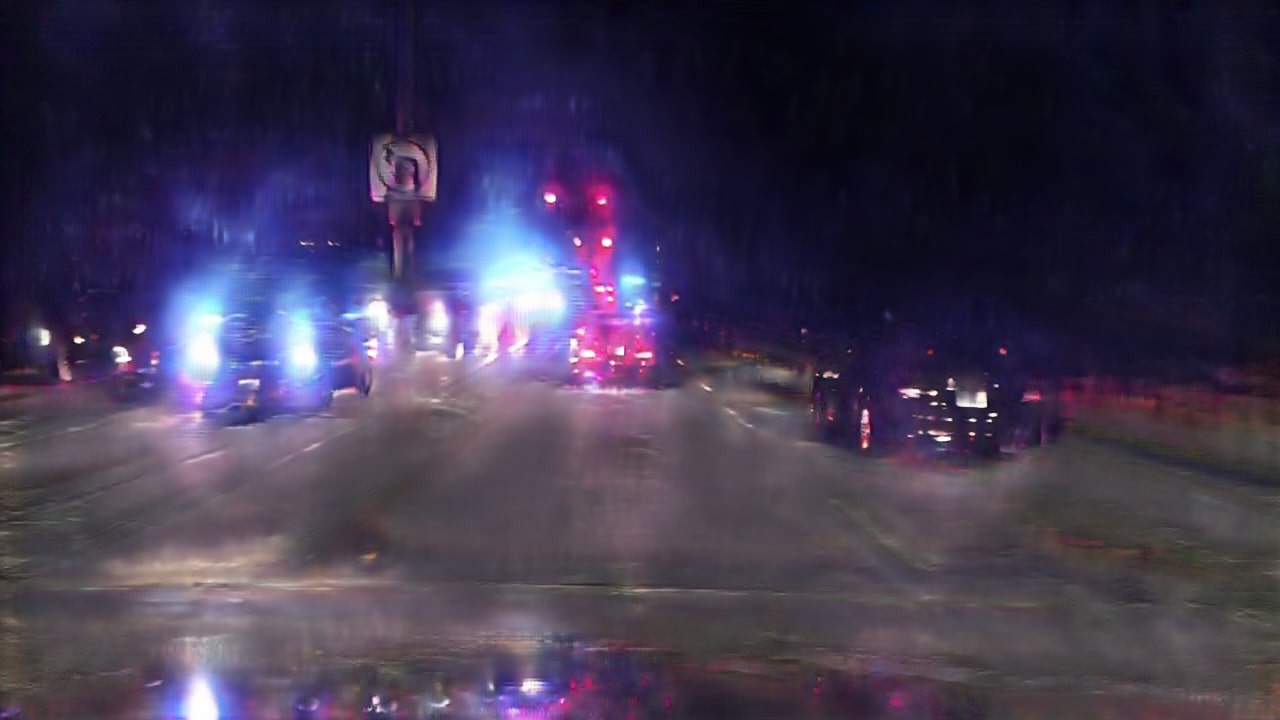}      
    \caption{Samples from the 'Berkeley - Manual Flasher Augmentation' dataset (left) and the 'Berkeley - CycleGAN-based synthesis' dataset (right). } 
    \label{fig:augmented-ds}
    \end{minipage}    
    \vspace{-3mm}
\end{figure*}


\section{The Caracetamol Framework}
\label{sec:countermeasures}
In this section, we introduce \textit{Caracetamol}, a framework aimed at enhancing the resilience and accuracy of object detection in the presence of emergency vehicle lighting that takes the constraints of real-time driving into account.
In the subsections that follow, we discuss the considerations for developing such a mitigation, describe the datasets employed for training and evaluation, and perform an evaluation of state-of-the-art (SOTA) flare removal methods and conclude that current methods do not meet the performance requirements for use in a real-time autonomous driving scenario.
At the end of the section, we propose a mitigation, describe its architecture, and evaluate its performance.

\subsection{Considerations}
Various factors must be considered when developing an effective and practical framework capable of mitigating the \textit{PaniCar} phenomenon in real-world conditions. 

\textbf{(1) Real-time Performance.} In high-speed autonomous driving scenarios, any mitigation must work in reasonable latency that will allow 30-60 frame per second rate detection and reaction to a detected object. 
Therefore, the required mitigation must be capable of making split-second decisions to ensure the safety of occupants and other road users. 
The mitigation runtime must be minimized to ensure that objects are identified quickly by object detectors, allowing sufficient time for ADAS to respond appropriately and avoid a collision. 
Methods that introduce significant delays may not be suitable for addressing the \textit{PaniCar} phenomenon under real-time autonomous driving constraints. 

\textbf{(2) Preserving Original Performance.} The object detectors in ADASs are responsible for detecting a wide range of objects in the driving environment under diverse driving conditions. Therefore, a mitigation aimed at mitigating the \textit{PaniCar} phenomenon must not compromise object detectors' ability to detect other objects.  

\textbf{(3) Limited Dataset Availability.} When undertaking this research we found that there was a lack of comprehensive datasets containing images of emergency vehicles with their emergency vehicle lighting on at night. To overcome this obstacle and enable the development of a suitable mitigation, we created augmented datasets which were used to train the models; the models' performance was evaluated with real images of first responder cars obtained from YouTube.

\textbf{(4) Maximal Deployment}. Due to the widespread occurrence of ADAS with a single video camera (e.g., Mobileye 630 PRO, "manufacturer C" and the five level 1 ADASs we analyzed in our study), the mitigation must be easy and cost-effective to deploy and capable of deployment on both new and existing single video camera ADASs. 
Therefore, we aimed to develop a software-based mitigation; doing so eliminates the need to install costly dedicated hardware or update existing hardware (e.g., adding LiDAR or RADAR). 

\subsection{Evaluation Metrics \& Datasets}
\label{subsection:metrics}
\subsubsection{Metrics} To evaluate a mitigation's robustness, we measured the object detector's performance using a number of metrics: (1) average object detector confidence (average performance), (2) minimum confidence (lower bound of performance), (3) maximum confidence (upper bound of performance), (4) confidence range (stability of performance), and (5) percentage of detections with confidence scores above 0.5/0.6/0.7/0.8 (distribution of high confidence detections)

\subsubsection{Datasets} We first introduce the datasets used to evaluate SOTA methods for flare removal and later for the proposed mitigation. We collected two authentic open-source image/video datasets and created two augmented datasets to address the lack of datasets containing images that include emergency vehicles with their emergency vehicle lighting on at night. The augmented datasets are available online\footref{fn:datasets} to enable the research community to advance research in this area. Samples from the datasets can be seen in Figs. \ref{fig:real-world-ds} and \ref{fig:augmented-ds}. 

\underline{Authentic Datasets.}
The datasets utilized in this work are as follows: 
\textbf{(1) YouTube Dataset.} This dataset consists of 243 videos that we downloaded from YouTube, comprising over 30,000 frames. The video frames demonstrate a diverse range of: 1) camera resolutions, 2) car orientations, and 3) emergency vehicle lighting patterns/configurations. This dataset was used to both evaluate \textit{Caracetamol} and facilitate the training of a CycleGAN model that was used to generate synthetic emergency vehicle lighting imagery (described below). 
\textbf{(2) BDD100K (Berkeley) Dataset.} The BDD100K dataset \cite{yu2020bdd100k}, also known as the Berkeley dataset, is a large and diverse labeled driving video dataset containing 100,000 car-related images. Despite its large and diverse range of images, the Berkeley dataset lacks images of activated emergency vehicle lighting at night. 
We use the Berkeley dataset as a base to augment the emergency vehicles lighting on top its picture.

\underline{Augmented Datasets.}
\label{sec:dataset-modifications}
Due to the limited number of images with emergency vehicle lighting in the Berkeley dataset \cite{yu2020bdd100k}, we created additional datasets containing images with emergency vehicles whose emergency vehicle lighting is on at night: 'Berkeley - Manual Flasher Augmentation' and 'Berkeley - CycleGAN-Based Synthesis'. 
These datasets were created using the Berkeley dataset by performing a process aimed at integrating emergency vehicle lighting elements into the nighttime images. 
We note that these datasets were only used for training. These datasets were not used for testing/evaluating the performance of \textit{Caracetamol} or any baseline object detector.
To achieve this, we identified nighttime images in the Berkeley dataset and added the effect of emergency vehicle lighting to the identified images. For each image, we used two methods to add the emergency vehicle lighting images: (1) a manual code-based process, and (2) a CycleGAN-based approach. Images modified by the first method were included in the 'Berkeley - Manual Flasher Augmentation' dataset and images modified by the second method were included in the 'Berkeley - CycleGAN-Based Synthesis' dataset.
Each augmented dataset consisted of: (1) unmodified daytime images from the Berkeley dataset (\textit{i.e.,} all the images not identified as nighttime images) and (2) nighttime images modified using the respective modification method (\textit{i.e.,} all the images identified as nighttime images, with emergency vehicle lighting augmentations).

\textbf{Preprocessing}. To identify nighttime images in the Berkeley dataset for targeted modification, we employed a method proposed in prior research \cite{sandnes2010towards} in which the average brightness of each pixel in the image is calculated to determine whether the image represents a daytime or nighttime scene. 
Images with an average brightness value above a predefined threshold of 60 are categorized as daytime, while those below the threshold are classified as nighttime. 
Using this method, we identified approximately 27,500 nighttime images in the Berkeley dataset, to which we added emergency vehicle lighting imagery.

\textbf{Augmentation}. Two methods were used to add emergency vehicle lighting elements to the nighttime images:

\textbf{(3) Berkeley - MFA (Manual Flasher Augmentation).} In this method, synthetic emergency vehicle lighting imagery was added to nighttime images utilizing random noise patterns resembling the lights from emergency vehicle lighting. This allowed us to simulate the presence of emergency vehicles in nighttime car-related images and generate more diverse training data (the Python script we used can be found in the Appendix in Listing \ref{listing-script}). 
We used the script to add emergency vehicle lighting imagery on 27,500 images from the Berkeley dataset (one for each nighttime image) enriching the dataset with images "containing" emergency vehicle lighting. 
The color (red or blue) of the emergency vehicle lighting was randomly chosen for each image, and it was applied in a random area of the image.
The application of the emergency vehicle lighting imagery in random areas of the image was performed in order to simulate an attacker placing a siren in random locations, not necessarily on the roof of an emergency vehicle (e.g., in a tree, on the ground, etc.).
By performing this modification, we created a new dataset called the 'Berkeley - MFA' dataset.

\textbf{(4) Berkeley - CycleGAN-Based Synthesis.} In this method, a CycleGAN-based \cite{CycleGAN2017} emergency vehicle lighting generator was used to add emergency vehicle lighting imagery to images in the Berkeley dataset. 
To train the model, two groups of images were used: Group A (without emergency vehicle lighting), which consisted of approximately 1,000 images from the Berkeley dataset, and Group B (with emergency vehicle lighting), which consisted of approximately 1,000 images from the YouTube dataset. 
The model was trained for 200 epochs on all the images.
This training process produced two models/generators: one generator that adds imagery of emergency vehicle lighting (\textit{i.e.,} streaks of light and color) to input images and a second generator, a denoiser, that filters the flare caused by emergency vehicle lighting from input images.
We used the first generator to add emergency vehicle lighting imagery on 27,500 images (one for each nighttime image) from the Berkeley dataset.
By performing this modification, we created a new dataset called the 'Berkeley - CycleGAN-Based Synthesis' dataset.
Later in this section we discuss the use of the second generator, that filters flare from a picture, which we refer to as the \textit{denoiser}.

\subsection{Performance of SOTA Flare Removal Methods }
\label{sec:comparison-sota-flasher-removal}

Here, we evaluate various state-of-the-art (SOTA) methods for emergency vehicle lighting flare removal from still video frames. The goal of this evaluation is to demonstrate performance and latency issues in current SOTA flash removal methods, which hinder their utility in real-time driving scenarios. We evaluate four SOTA methods according to two performance metrics: (1) robustness of the object detection model after performing the flare removal method and (2) computational efficiency (processing delay) of the flare removal method. The methods analyzed are as follows:
(1) The method presented in \cite{wu2021train}(HTNN), which uses a U-Net-based convolutional neural network to remove lens flare from images.
(2) The method presented in \cite{zhou2023improving} (ILFR), which uses a dual-part neural network architecture to perform flare segmentation across varied lighting conditions and refine the image by removing the identified flares and correcting color imbalances.
(3) The method presented in \cite{dai2022flare7k}  (Flare7K), which involves augmenting object detector training with Flare7K, a dataset intended for nighttime flare removal.
Flare7K includes 7,000 synthetic images of scattering and reflective flares, developed by analyzing real-world night flares.
(4) In addition, the authors of Flare7K \cite{dai2022flare7k} produced a second dataset, Flare7K++, which consists of Flare7K augmented with 962 recorded images of flare patterns from the Flare-R dataset. We separately evaluate both Flare7K and Flare7K++.

\begin{table}[h]
    \centering
    \caption{Performance comparison of  Flare7k++, Flare7k, ILFR, and HTNN on RTX-3090 and RTX-6000 GPUs.}\begin{tabular}{lccccc}
        \toprule
        Method & Resolution & Adaptation &  \makecell{RTX-3090\\ (ms)} &  \makecell{RTX-6000\\ (ms)} \\
        \midrule
        Flare7k++ & 1280$\times$720 & 512$\times$512 & 69.84 & 40.27 \\
        Flare7k & 1280$\times$720 & 512$\times$512 & 69.63 & 40.97 \\
        ILFR
        & 1280$\times$720 & 512$\times$512
        & 167.4 & 102.89 \\
        HTNN
        & 1280$\times$720 & 512$\times$512 & 18.62 & 12.28 \\
        \bottomrule
    \end{tabular}
\label{tab:related-methods-gpu_performance}
    \vspace{-5mm}
\end{table}

\begin{table*}[h]
    \centering
    \caption{Confidence metrics for YOLOv3 and Faster R-CNN on images processed by state-of-the-art flare removal methods (Flare7k, Flare7k++, ILFR, HTNN).}\begin{tabular}{lcccccccc}
        \toprule
        \textbf{Method} & \textbf{Avg Conf} & \textbf{Min Conf} & \textbf{Max Conf} & \textbf{Range} & \textbf{\% >0.5} & \textbf{\% >0.6} & \textbf{\% >0.7} & \textbf{\% >0.8} \\
        \midrule
        Flare7k (Faster R-CNN) & 0.57 & 0.22 & 0.91 & 0.68 & 0.60 & 0.52 & 0.41 & 0.30 \\
        Flare7k (YOLOv3) & 0.30 & 0.07 & 0.73 & 0.65 & 0.26 & 0.20 & 0.15 & 0.11 \\
        Flare7k++ (Faster R-CNN) & 0.57 & 0.22 & 0.91 & 0.69 & 0.61 & 0.52 & 0.42 & 0.30 \\
        Flare7k++ (YOLOv3) & 0.51 & 0.18 & 0.87 & 0.68 & 0.52 & 0.41 & 0.32 & 0.23 \\
        ILFR (Faster R-CNN) & 0.44 & 0.07 & 0.88 & 0.81 & 0.41 & 0.30 & 0.21 & 0.12 \\
        ILFR (YOLOv3) & 0.50 & 0.18 & 0.86 & 0.68 & 0.50 & 0.41 & 0.31 & 0.22 \\
        HTNN (Faster R-CNN) & 0.59 & 0.23 & 0.92 & 0.69 & 0.63 & 0.54 & 0.44 & 0.31 \\
        HTNN (YOLOv3) & 0.49 & 0.17 & 0.86 & 0.68 & 0.49 & 0.40 & 0.31 & 0.22 \\
        \bottomrule
    \end{tabular}
\label{tab:related-methods-confidence_metrics}
    \vspace{-5mm}
\end{table*}

\subsubsection{Evaluation of Model Processing Delay}

\textbf{Experimental Setup.}
We fed 10,000 images from the Berkeley dataset to each flare removal method (HTNN, ILFR, Flare7k, Flare7k++). 
We loaded each method on a GPU and calculated the average runtime of each component per image. We repeated this evaluation twice, each time with a different GPU (RTX-3090, RTX-6000).

\textbf{Results.}
The results are presented in Table \ref{tab:related-methods-gpu_performance}. We found that on RTX-3090 and RTX-6000, HTNN has the lowest average runtime of 18.62ms and 12.28ms, respectively, while the other methods introduce additional overheads beyond 40 ms (excluding detection).


\subsubsection{Evaluation of Model Robustness}

\textbf{Experimental Setup.}
We fed the YouTube data test set to each flare removal method (HTNN, ILFR, Flare7k, Flare7k++). We then took the outputs of each method and fed them into an object detector, measuring the object detector's robustness according to the metrics described above. We repeated this evaluation with two object detector architectures to which we applied the various flare removal methods: YOLOv3 and Faster R-CNN, both pretrained on the MS-COCO dataset.

\textbf{Results.}
The results are presented in Table \ref{tab:related-methods-confidence_metrics}. We can see that for Faster R-CNN and YOLOv3, HTNN demonstrates the highest average confidence scores of 0.59, and 0.49, respectively. In addition, for Faster R-CNN, HTNN demonstrated the highest proportion of detections with confidence greater than 0.5/0.6/0.7/0.8. 


\begin{insight} \label{insight:SOTA}
All SOTA flare removal methods achieve an average confidence score below 0.6 and often incur significant overhead, introducing more than 40 ms of additional latency (excluding detection), which makes them unsuitable for real-time applications at 30–60 FPS.
\end{insight}

To address these shortcomings in latency and performance, we present in the following sections a method that improves on these two limitations for real-time driving scenarios.



\subsection{The Framework}




\textit{Caracetamol} is a software-based framework for existing object detection algorithms aimed at increasing their robustness in the presence of emergency vehicle lighting in nighttime settings.
The system works as follows on each frame processed by the ADAS:
\textbf{(1)} The frame is passed through a denoiser, in order to remove artifacts caused by flashes from emergency vehicle lighting. The denoiser is derived from a CycleGAN model trained to receive images with flashes and output the same image without flashes, consisting of the CycleGAN generator described in Section \ref{sec:dataset-modifications}. 
\textbf{(2)} The processed image, output by the denoiser, is then fed to a fine-tuned model, a copy of the original object detector that was fine tuned on one of the datasets we augmented.
\textbf{(3)} The original (not processed) frame is fed to the original ADAS object detector, in order to ensure that all detections made by the original ADAS are also made by the \CC\ framework. This prevents \CC\ from inadvertently compromising the performance of the original ADAS.
\textbf{(4)} The outputs from (2) and (3) are combined in a combiner layer, and the aggregated results are output as the detections of the \CC\ framework.

\begin{table*}[h]
\centering
\caption{ Confidence metrics for YOLO, Faster R-CNN (FRCNN) and SSD fine-tuned on different datasets, with and without utilizing the \CC\  denoiser preprocessor.}
\begin{tabular}{|l|c|c|c|c|c|c|}
\hline
\textbf{Metric} &
\makecell{$\text{YOLO}_{\text{COCO}}$} &
\makecell{$\text{YOLO}_{\text{MFA}}$} &
\makecell{$\text{YOLO}_{\text{CycleGAN}}$} &
\makecell{Denoiser +\\$\text{YOLO}_{\text{COCO}}$} &
\makecell{Denoiser +\\$\text{YOLO}_{\text{MFA}}$} &
\makecell{Denoiser +\\$\text{YOLO}_{\text{CycleGAN}}$} \\
\hline
Average Confidence & 0.50 & 0.54 & 0.54 & 0.48 & 0.71 & 0.48 \\
Absolute Range     & 0.69 & 0.52 & 0.70 & 0.75 & 0.23 & 0.76 \\
Above 0.5          & 0.51 & 0.89 & 0.52 & 0.46 & 1.00 & 0.42 \\
Above 0.6          & 0.42 & 0.38 & 0.46 & 0.35 & 0.94 & 0.35 \\
Above 0.7          & 0.33 & 0.01 & 0.40 & 0.25 & 0.63 & 0.28 \\
Above 0.8          & 0.24 & 0.00 & 0.33 & 0.17 & 0.07 & 0.22 \\
Minimum Value      & 0.18 & 0.15 & 0.20 & 0.13 & 0.59 & 0.15 \\
Maximum Value      & 0.86 & 0.67 & 0.90 & 0.88 & 0.82 & 0.90 \\
\hline
\end{tabular}
\vspace{2mm}

\begin{tabular}{|l|c|c|c|c|c|c|}
\hline
\textbf{Metric} &
\makecell{$\text{FRCNN}_{\text{COCO}}$} &
\makecell{$\text{FRCNN}_{\text{MFA}}$} &
\makecell{$\text{FRCNN}_{\text{CycleGAN}}$} &
\makecell{Denoiser +\\$\text{FRCNN}_{\text{COCO}}$} &
\makecell{Denoiser +\\$\text{FRCNN}_{\text{MFA}}$} &
\makecell{Denoiser +\\$\text{FRCNN}_{\text{CycleGAN}}$} \\
\hline
Average Confidence & 0.63 & 0.80 & 0.75 & 0.63 & 0.81 & 0.77 \\
Absolute Range     & 0.73 & 0.49 & 0.55 & 0.73 & 0.53 & 0.57 \\
Above 0.5          & 0.71 & 0.91 & 0.85 & 0.71 & 0.95 & 0.91 \\
Above 0.6          & 0.59 & 0.85 & 0.76 & 0.59 & 0.90 & 0.82 \\
Above 0.7          & 0.44 & 0.76 & 0.64 & 0.44 & 0.80 & 0.69 \\
Above 0.8          & 0.28 & 0.62 & 0.49 & 0.28 & 0.61 & 0.50 \\
Minimum Value      & 0.21 & 0.48 & 0.41 & 0.21 & 0.45 & 0.40 \\
Maximum Value      & 0.94 & 0.97 & 0.96 & 0.94 & 0.98 & 0.98 \\
\hline
\end{tabular}
\vspace{2mm}

\begin{tabular}{|l|c|c|c|c|c|c|}
\hline
\textbf{Metric} &
\makecell{$\text{SSD}_{\text{COCO}}$} &
\makecell{$\text{SSD}_{\text{MFA}}$} &
\makecell{$\text{SSD}_{\text{CycleGAN}}$} &
\makecell{Denoiser +\\$\text{SSD}_{\text{COCO}}$} &
\makecell{Denoiser +\\$\text{SSD}_{\text{MFA}}$} &
\makecell{Denoiser +\\$\text{SSD}_{\text{CycleGAN}}$} \\
\hline
Average Confidence & 0.49 & 0.40 & 0.31 & 0.46 & 0.56 & 0.41 \\
Absolute Range     & 0.65 & 0.68 & 0.52 & 0.69 & 0.70 & 0.66 \\
Above 0.5          & 0.47 & 0.27 & 0.13 & 0.39 & 0.56 & 0.28 \\
Above 0.6          & 0.39 & 0.20 & 0.09 & 0.31 & 0.42 & 0.18 \\
Above 0.7          & 0.31 & 0.15 & 0.07 & 0.23 & 0.31 & 0.11 \\
Above 0.8          & 0.23 & 0.11 & 0.05 & 0.16 & 0.21 & 0.07 \\
Minimum Value      & 0.20 & 0.16 & 0.18 & 0.18 & 0.23 & 0.21 \\
Maximum Value      & 0.85 & 0.84 & 0.70 & 0.87 & 0.93 & 0.87 \\
\hline
\end{tabular}

\label{tab:yolo-confidence-comparison}
\vspace{-5mm}
\end{table*}

\definecolor{highlightgreen}{RGB}{0,255,0}
\definecolor{highlightred}{RGB}{255,200,200}

\begin{table}[!t]
\centering
\renewcommand{\arraystretch}{1.0}
\caption{Inference time and average confidence across RTX-3090 and RTX-6000 GPUs. The color marks if the method can run (in green) in 30–60 FPS or not (in red). The added percentage of overhead (in latency) and improvement (in average score) with respect to the original model that was trained on COCO appear in the bracket.}
\label{tab:timing-confidence}
\resizebox{\columnwidth}{!}{%
\begin{tabular}{|l|c|c|c|}
\hline
\rowcolor{white}
\textbf{Model} &
\makecell{\textbf{RTX-3090}\\(Latency / FPS)} &
\makecell{\textbf{RTX-6000}\\(Latency / FPS)} &
\makecell{\textbf{Score}\\($\triangle$)} \\
\hline
Denoiser & 5 ms & 4 ms & — \\
\hline
{$\text{YOLO}_{\text{MFA}}$} &
\cellcolor{highlightgreen} \makecell{21 ms (+0.0\%) /\\ 47 FPS} &
\cellcolor{highlightgreen} \makecell{16 ms (+0.0\%) /\\ 62 FPS} &
\makecell{0.54\\ (+8.0\%)} \\
{$\text{YOLO}_{\text{CycleGAN}}$} &
\cellcolor{highlightgreen}\makecell{21 ms (+0.0\%) /\\ 47 FPS} &
\cellcolor{highlightgreen}\makecell{16 ms (+0.0\%) /\\ 62 FPS} &
\makecell{0.54\\ (+8.0\%)} \\
\makecell{Denoiser +\\{$\text{YOLO}_{\text{MFA}}$}} &
\cellcolor{highlightgreen} \makecell{26 ms (+23.8\%) /\\ 38 FPS} &
\cellcolor{highlightgreen} \makecell{20 ms (+25.0\%) /\\ 50 FPS} &
\makecell{0.71\\ (+42.0\%)} \\
\hline
{$\text{FRCNN}_{\text{MFA}}$} &
\cellcolor{highlightgreen} \makecell{32 ms (+0.0\%) /\\ 31 FPS} &
\cellcolor{highlightgreen} \makecell{26 ms (+0.0\%) /\\ 38 FPS} &
\makecell{0.80\\ (+26.98\%)} \\
\makecell{Denoiser +\\{$\text{FRCNN}_{\text{MFA}}$}} &
\cellcolor{highlightred} \makecell{37 ms (+15.6\%) /\\ 27 FPS} &
\cellcolor{highlightgreen} \makecell{30 ms (+15.4\%) /\\ 33 FPS} &
\makecell{0.81\\ (+28.57\%)} \\
\hline
\end{tabular}
}
\vspace{-2mm}
\end{table}

\subsection{Evaluation}

In this section we evaluate \textit{Caracetamol}'s performance in detecting vehicles with emergency vehicle lighting and compare its performance when three different base object detectors (YOLO, SSD, Faster R-CNN) are used with \textit{Caracetamol}. We evaluate the performance (\textit{i.e.,} average detection confidence) of \textit{Caracetamol}, as well as the runtime overhead of \textit{Caracetamol}.

\subsubsection{Performance Evaluation}
\textbf{Training Setup.}
We utilized three object detector architectures: YOLOv3, SSD, and Faster R-CNN.
For each model architecture, we evaluated the performance of: (1) the original unaltered model, (2) the model after fine-tuning on the Berkeley-MFA dataset, and (3) the model after fine-tuning on the Berkeley-CycleGAN dataset.
We fed the YouTube data test to each model and analyzed the model performance according to the metrics described in Section \ref{subsection:metrics}.

We then repeated these evaluations after initially running the YouTube data through the \textit{Caracetamol} denoiser.
The fine-tuning data consisted of 100,000 images from the Berkeley-MFA, or Berkeley-CycleGAN dataset, with a 70/10/20 training/validation/test split.
The YOLO model fine-tuning was performed for 50 epochs with early stopping, using the SGD optimizer with a learning rate of 0.1, momentum of 0.9, and weight decay of 0.0005.
The SSD model fine-tuning was performed for 50 epochs with early stopping, using the SGD optimizer with a learning rate of 0.0001, momentum of 0.9, and weight decay of 0.0001.
The Faster R-CNN model fine-tuning was performed for 50 epochs with early stopping, using the SGD optimizer with a learning rate of 0.01, momentum of 0.9, and weight decay of 0.0001.
Training for each model was stopped if 10 consecutive epochs failed to yield an improvement of at least 0.01 in the average confidence score. 



\textbf{Results}. 
The results for YOLOv3, SSD, and Faster R-CNN are presented in Table \ref{tab:yolo-confidence-comparison}
We can see that for YOLOv3, fine-tuning the original model on the augmented datasets (Berkeley-MFA and Berkeley-CycleGAN) improves the average confidence from 0.50 to 0.54. When applying the denoiser before the model trained on Berkeley-MFA, the average confidence rises to 0.71, a 42\% increase in confidence over the original YOLOv3 model, and the confidence range is reduced from 0.69 to 0.23. We also can see an increase in the proportion of detections with confidence over 0.5, from 0.51 in the original YOLOv3 model to 1.0 when applying the denoiser before the model trained on Berkeley-MFA, a 96\% increase.
We can also see similar performance improvements for Faster R-CNN, where utilizing the denoiser and fine-tuning the original model on Berkeley-MFA results in an average confidence improvement from 0.63 to 0.81. In addition, the proportion of detection confidences above 0.5 improved for Faster R-CNN from 0.71 to 0.95.
Despite the promising results for YOLOv3 and Faster R-CNN, we see less significant improvements for SSD, such as an average confidence improvement from 0.49 to 0.56, and the proportion of detection confidences above 0.5 improving from 0.47 to 0.56. We leave improving on SSD for future work.

\subsubsection{Runtime Analysis}
\label{sec:runtime-analysis}
In order to be used in real-time ADASs, \textit{Caracetamol} must exhibit real-time performance. 
We evaluate the runtime of (1) the \textit{Caracetamol} denoiser, and (2) an object detector (YOLOv3 or Faster R-CNN) with a \textit{Caracetamol} denoiser preprocessor, to ensure that the solution is efficient and capable of operating within the constraints of real-world driving situations.

\textbf{Experimental Setup:}
We fed 10,000 images from the Berkeley dataset to (1) the \textit{Caracetamol} denoiser, and (2) to each of the YOLOv3 and Faster R-CNN object detectors, with and without a \textit{Caracetamol} denoiser preprocessor. We performed the evaluations on models fine-tuned on "Berkeley-MFA", as well as a YOLOv3 model (without the denoiser) fine-tuned on "Berkeley-CycleGAN".
We loaded each component on a GPU and calculated the average runtime of each component on each image. 
We repeated this evaluation using the RTX-3090 and RTX-6000 GPUs.

\textbf{Results:}
As can be seen from the results in Table \ref{tab:timing-confidence}, on the RTX-3090 and RTX-6000 GPUs, the models' runtime is less than 37 milliseconds and 30 milliseconds, respectively. This means that in 9 out of 10 evaluated cases, \textit{Caracetamol} can process at least 31 frames per second (FPS) during a flash, while maintaining high performance (average detection of 0.71 - 0.81). 

We note that Tesla and other manufacturers of semi-autonomous cars use a dedicated GPU intended for real-time object detection. 
Since we do not have any access to the GPUs used by the car industry, we decided to measure the percentage of overhead and the percentage of improvement that \textit{Caracetamol} adds. 
This may give a better indication regarding the results and hopefully can transfer from one GPU to another.
As can be seen in Table \ref{tab:timing-confidence}, \textit{Caracetamol} can improve the average detection in 26.98\% with no additional overhead in the case of Faster $R-CNN_{MFA}$. Alternatively, \textit{Caracetamol} can improve the average detection in 42\% with an additional latency overhead of 23.5\%-25\% in case the of $Denoiser + YOLO_{MFA}$.

\vspace{2mm}
\begin{insight} \label{insight:suitable}
With the use of a YOLOv3 or Faster R-CNN object detector, as well as a denoiser, we can support a 31-50 FPS rate, with an average detection confidence of 0.71 - 0.81, depending on the combination being used.
\end{insight}

\subsubsection{Robustness against Adversaries}

In this section we discuss the robustness of \textit{Caracetamol} against potential adversaries aiming to attack its functionality.
One potential attack vector would be to utilize adversarial perturbations, however due to the physical-world setting of \textit{Caracetamol}, perturbations cannot be applied to emergency vehicle lighting. 
This prevents an attacker from introducing adversarial perturbations in \textit{Caracetamol}'s input.

\section{Limitations}
\label{sec:limitaitons}

We performed this research independently, with no collaboration with the automotive industry. 
While this allowed us to avoid a conflict of interest preventing us from publishing our findings, some limitations with our work could have been overcome with the automotive industry's involvement. 
\textbf{Lack of Access to Reports on the Relevant Accidents.} The reports on the 16 incidents that include the footage obtained by Tesla's video cameras before the accidents occurred and the behavior of the relevant object detectors are not available online. 
We performed the most comprehensive research possible given the lack of access to confidential footage, reports, and analysis, relying purely on the limited data and information that is publicly available via the Internet.
\textbf{Difference in the Tested Components.} We note that in our research, we used commercial ADASs to obtain video footage and four commonly used object detectors to conduct the analysis. 
The transferability of our findings to Tesla's object detectors must be examined.
We hope that this research will motivate the automotive industry to validate our findings on real advanced driver-assistance systems (ADASs).
\textbf{Inability to Explore and Understand the Full Semi-Autonomous Cars Context.} While our research revealed the limitation of object detectors in the presence of activated emergency vehicle lighting, other technical factors must also be analyzed to understand the full picture and explore the potential interplay of various factors in the context of ADASs. 
For example, the reason why the integrated RADAR did not detect the parked emergency vehicles should also be analyzed. 
While some explanations regarding this can be found in Volvo's manual \cite{wired}, it is our hope that a detailed report covering all of the technological aspects will be published and shed light on the perfect storm that caused Teslas to crash into emergency vehicles. 
\textbf{Limited Availability of Datasets Containing Emergency Vehicles.} One challenge we faced in designing our framework was the limited number of nighttime images of emergency vehicles with activated emergency vehicle lighting. 
While we overcame this limitation by augmenting the training data, we believe that improved detection confidence results could be obtained by fine-tuning the object detector using actual images of emergency vehicles with activated emergency vehicle lighting at night.

\section{Related Work}
\label{sec:related-works}

In recent years, researchers have investigated the robustness and limitations of advanced driver-assistance systems (ADASs). 
One line of research investigated the robustness and limitations of the various sensors incorporated in ADASs.
This includes studies that tested the limitations of LiDAR \cite{petit2015remote, cao2019adversarial, LIDAR-USENIX20, cao2021invisible, cao2023you, sato2023lidar} and video cameras \cite{yan2022rolling, lovisotto2021slap, ji2021poltergeist}.
A second avenue of research investigated the robustness and limitations of object detectors \cite{song2018physical, chen2018shapeshifter, Zhao-CCS2019, zolfi2021translucent, sato2024invisible}.
Other studies used simulators to investigate the implications of attacks \cite{shen2020drift, cao2021invisible}.
These studies above shed light on the limitations of object detection algorithms and sensors. 
However, a recent SoK \cite{shen2022sok} argued that the findings of the abovementioned studies might not apply to real ADASs, because \textit{"AI component-level errors do not necessarily lead to system-level effects (e.g., vehicle collisions)"}.

A third area of research investigated the robustness and limitations of real ADASs (e.g., Tesla, Mobileye, HARMAN’s ADAS, and OpenPilot). 
This research includes studies that examined the limitations of real ADASs, using attacks that triggered an undesired reaction from the controlled cars.
A few studies emphasized the limitations of an ADAS' video camera to time domain attacks \cite{nassi2020phantom, nassi2023protecting}, projections \cite{nassi2019mobilbye, nassi2020phantom}, infrared light \cite{wang2021can, sato2024invisible},
laser beams \cite{petit2015remote}, and physical patches \cite{sato2021dirty, KeenLabs, morgulis2019fooling, mcafee2020, jing2021too, nassi2022badvertisement}. 
Other studies investigated the limitations of an ADAS' RADAR \cite{hunt2023madradar, yan2016can}, ultrasonic sensors \cite{yan2016can}, and GPS sensor \cite{GPS-Spoofing-Regulus}.

\section{Discussion}
\label{sec:discussion}

The objective of this research was to shed light on the technical factors underlying the documented incidents in which semi-autonomous cars crashed into emergency vehicles, and it was performed in response to the lack of empirical research on this topic and the limited information available on the Internet.
By conducting experiments independently and in a controlled environment using video footage obtained by commercial ADAS, we found that activated emergency vehicle lighting degraded object detectors' performance and proposed a framework aimed at securing object detectors in the presence of activated emergency vehicle lighting. On YOLOv3 and Faster R-CNN, \CC\ improves the models’ average confidence of car detection by 0.20, the lower confidence bound by 0.33, and reduced the fluctuation range by 0.33. In addition, \CC\ is capable of processing frames at a rate of between 30 and 50 FPS, enabling real-time ADAS object detection.
We hope our findings will motivate the automotive industry, NHTSA, and other stakeholders to validate our findings on additional ADASs and cars.

\bibliographystyle{IEEEtran}
\bibliography{IEEEabrv, main}
\begin{figure*}
  \centering
    \begin{minipage}[b]{0.33\textwidth}
    \includegraphics[width=\textwidth]{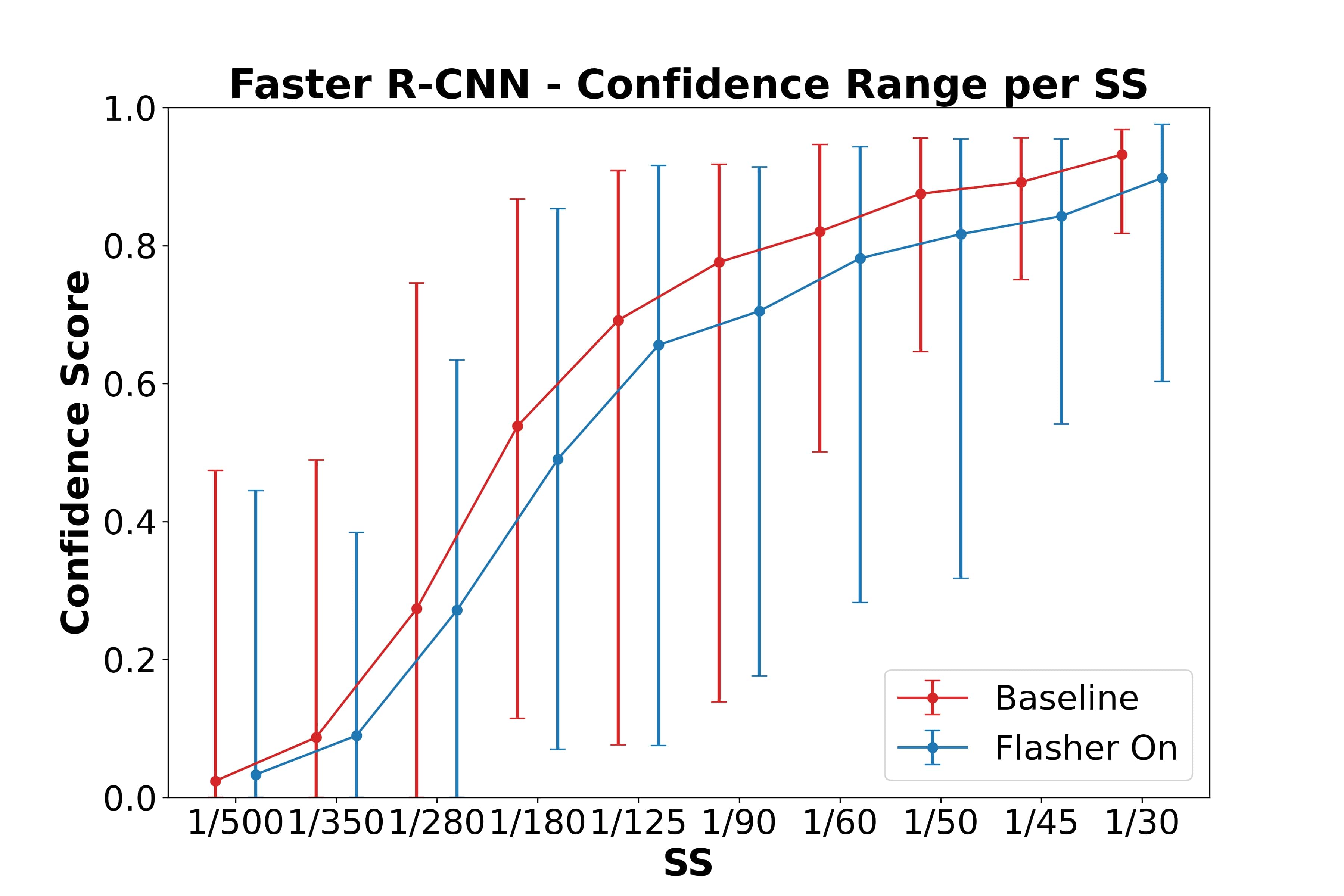} 
    \caption{Faster R-CNN confidence score signal range for each examined shutter speed (SS). The results obtained when the emergency vehicle lighting was on are presented in blue, while the results obtained when the emergency vehicle lighting was off are in red; the points in each range indicate the average value for each signal.}
    \label{fig:SS}
  \end{minipage}
  \hspace{0.02em}
  \begin{minipage}[b]{0.32\textwidth}
\includegraphics[width=\textwidth]{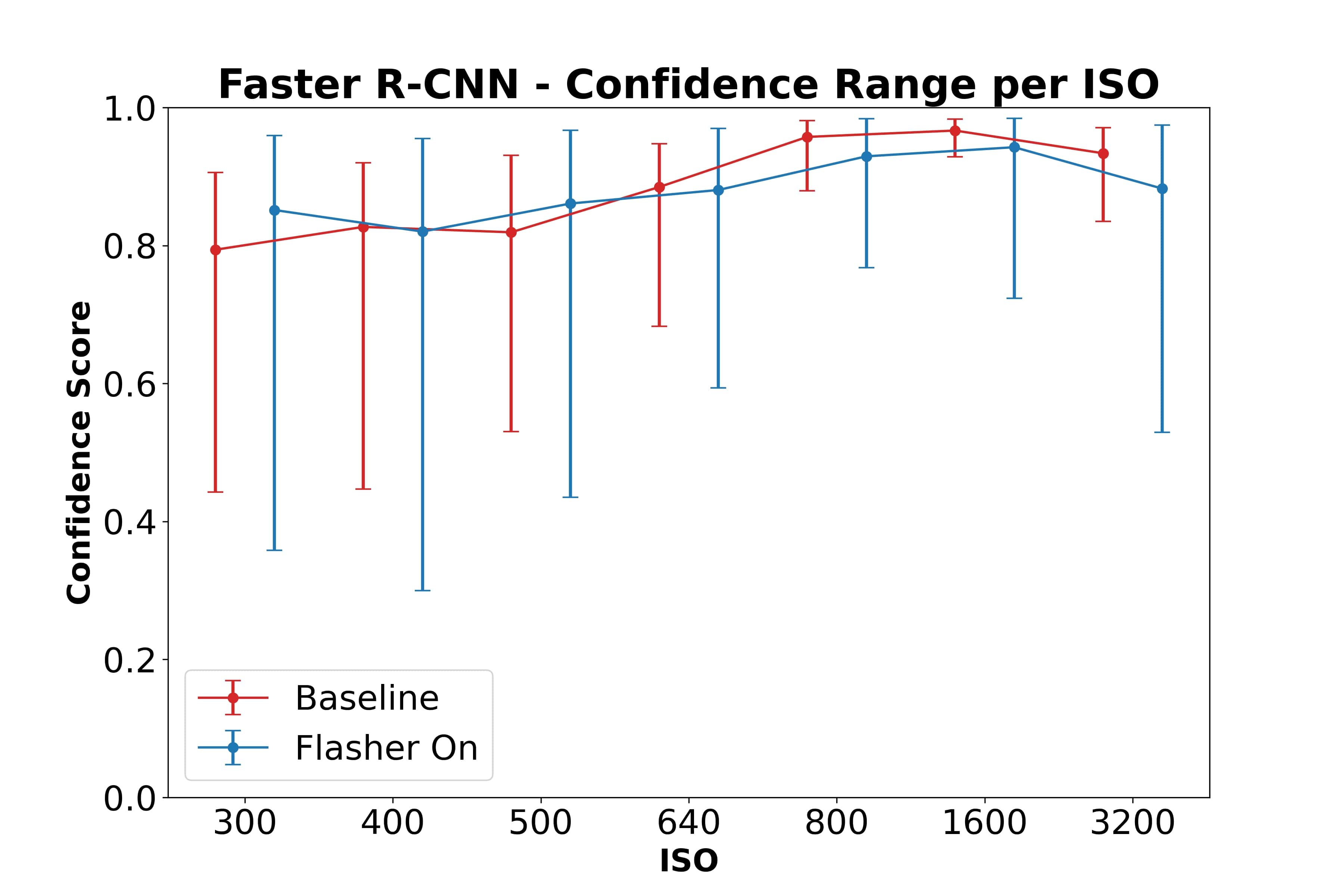}
\caption{Faster R-CNN confidence score signal range for each examined ISO value. The results obtained when the emergency vehicle lighting was on are presented in blue, while the results obtained when the emergency vehicle lighting was off are in red; the points in each range indicate the average value for each signal.}
\label{fig:ISO}
  \end{minipage}
\hspace{0.02em}
  \begin{minipage}[b]{0.32\textwidth}
  \includegraphics[width=\textwidth]{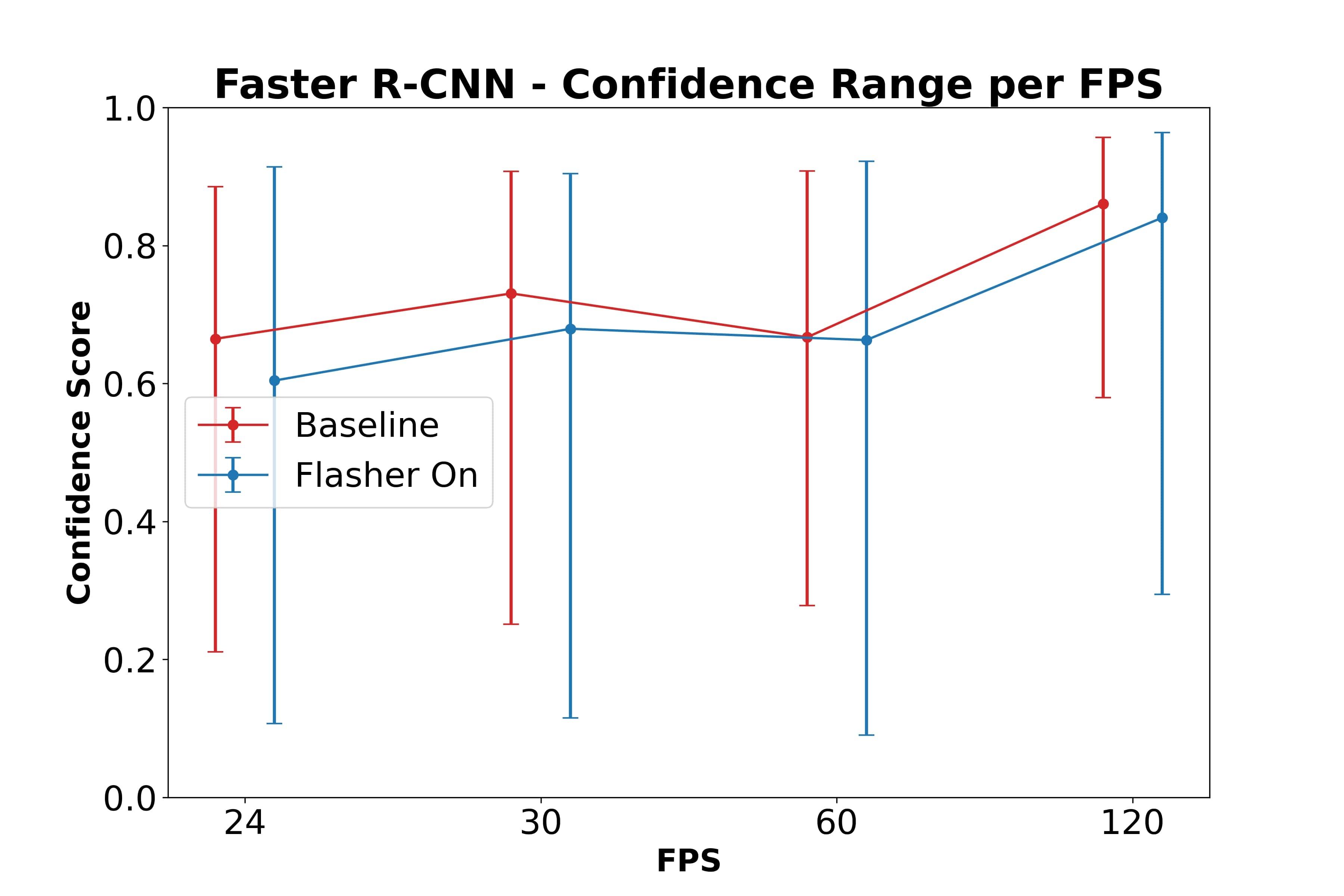} 
\caption{Faster R-CNN confidence score signal range for each examined FPS value. The results obtained when the emergency vehicle lighting was on are presented in blue, while the results obtained when the emergency vehicle lighting was off are in red; the points in each range indicate the average value for each signal.}
    \label{fig:FPS}
  \end{minipage}
  \vspace{-1.5em}
\end{figure*}

\section{Appendix A: Effect of the Camera Settings}

Here we examine whether advanced driver-assistance system (ADAS) camera settings influence the \textit{\EC} phenomenon, focusing on: (1) the shutter speed, (2) the ISO sensitivity, and (3) the frame rate of the vehicle's camera.
Due to space limitations, we only present the analysis performed using Faster R-CNN. 

\subsection{Effect of the Camera's Shutter Speed} In this experiment, we investigate whether the shutter speed of an ADAS' camera affects the confidence score range of object detectors when detecting a vehicle whose emergency vehicle lighting is on and off.

\textbf{Experimental Setup:} A Samsung Galaxy S22 Ultra was positioned five meters away from a grey Ford Fiesta with an emergency vehicle lighting mounted on its roof in a dark setting. Ten 30-second videos of the car were recorded with the emergency vehicle lighting on and with the emergency vehicle lighting off, varying the camera's shutter speed (1/500, 1/350, 1/250, 1/180, 1/125, 1/90, 1/60, 1/50, 1/45, 1/30 seconds). Faster R-CNN was applied to the video frames, and for each video, we assessed the confidence score range of the object detectors when detecting the car. This experiment was conducted twice: first with the emergency vehicle lighting on and again when it was off.

\textbf{Results:} The results of this analysis for the Faster R-CNN detector are presented in Fig. \ref{fig:SS}. 
As can be seen, a decrease in the camera shutter speed leads to reduced stability in the detection confidence and an increase in the confidence score range.
\newline

\begin{insight} \label{insight:8}
As the camera's shutter speed decreases, the effect of the \textit{\EC} phenomenon increases.
\end{insight}

\subsection{Effect of the Camera's ISO} Here we investigate whether the ISO setting of an ADAS' camera influences the confidence score range of object detectors when detecting emergency vehicles whose emergency vehicle lighting is on and off.

\textbf{Experimental Setup:} A Samsung Galaxy S22 Ultra was positioned five meters away from a grey Ford Fiesta with emergency vehicle lighting mounted on its roof in a dark setting. Seven 30-second videos of the car were recorded with the emergency vehicle lighting on and with the emergency vehicle lighting off, varying the camera's ISO setting in each video (300, 400, 500, 640, 800, 1600, 3200). Faster R-CNN was applied, and for each video we assessed the confidence score range of the object detectors when detecting the car. This experiment was conducted twice: first with the emergency vehicle lighting on and again when it was off.

\textbf{Results:} The results of the Faster R-CNN object detector are presented in Fig. \ref{fig:ISO}. 
As can be seen, higher camera ISO values correspond to increased stability in object detection confidence and a smaller confidence range.
\newline

\begin{insight} \label{insight:9}
The \textit{\EC} phenomenon's effect decreases as the camera's ISO value increases.
\end{insight}

\subsection{Effect of the Camera's Frame Rate} In this experiment, we investigate whether the camera's frame rate has an impact on the confidence score range of object detectors when detecting a vehicle whose emergency vehicle lighting is on and off.

\textbf{Experimental Setup:} A Samsung Galaxy S22 Ultra was positioned five meters away from a Ford Fiesta with emergency vehicle lighting mounted on its roof in a dark setting. Four 30-second videos of the car were recorded with the emergency vehicle lighting on and with the emergency vehicle lighting off, varying the camera's frame rate in each video (24, 30, 60, 120 FPS). Faster R-CNN was applied to the video frames, and for each video we assessed the confidence range of the object detectors when detecting the car. 
This experiment was performed twice: first with the emergency vehicle lighting on and again when it was off.

\textbf{Results:} The results of this analysis for the Faster R-CNN detector are presented in Fig. \ref{fig:FPS}. 
As can be seen, the confidence range of the detectors remains unchanged despite variations in the FPS value.
\newline

\begin{insight} \label{insight:10}
The \textit{\EC} phenomenon is not affected by the camera's frame rate.
\end{insight}

\begin{figure*}[]
  \centering
  \begin{minipage}[b]{0.32\textwidth}
    \centering
  \includegraphics[width=1\textwidth]{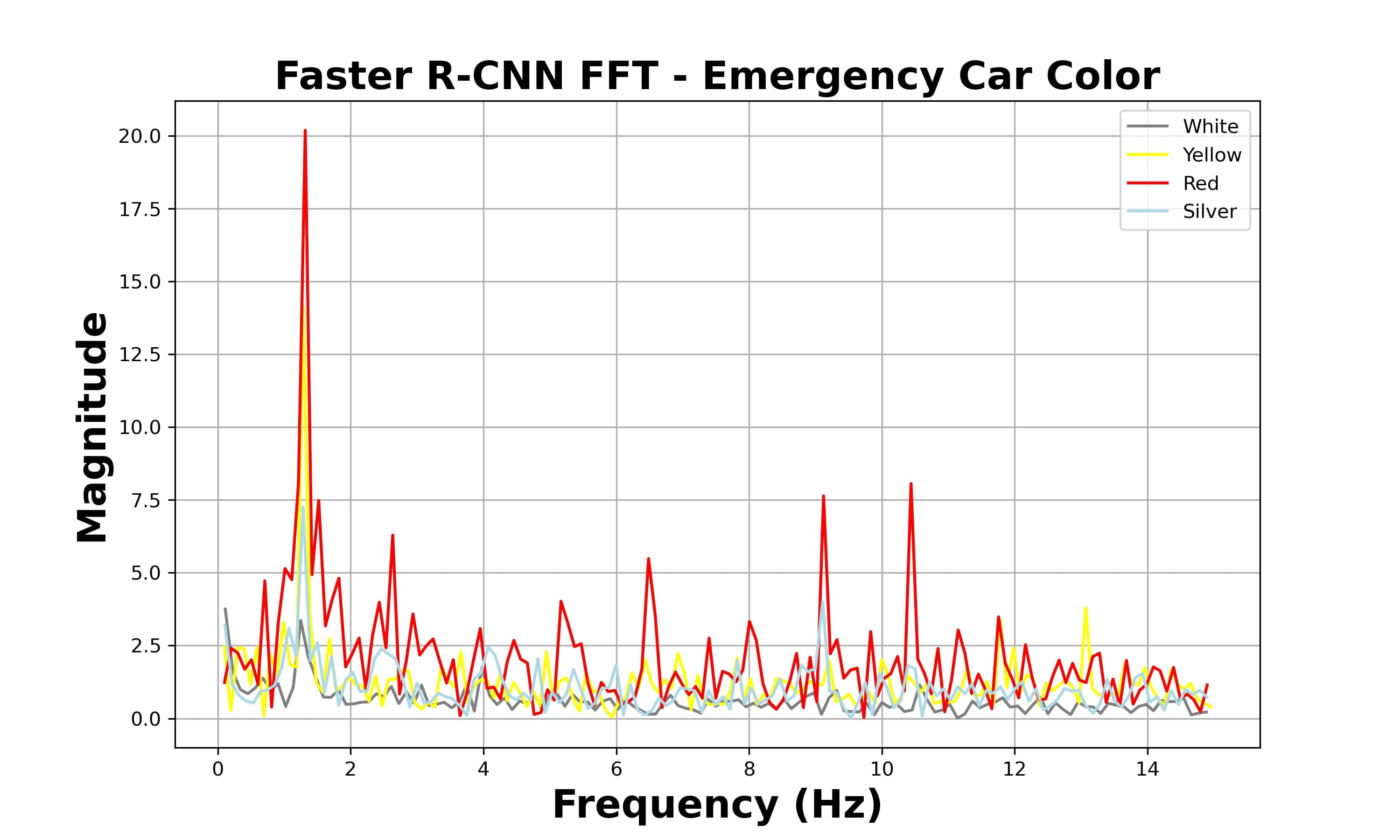} 
\caption{Fast Fourier transform (FFT) of the confidence score signal produced by Faster R-CNN for each examined emergency vehicle color. All vehicles were equipped with active emergency lighting. A dominant frequency peak at ~1.3 Hz is observed across all color conditions, indicating a consistent temporal pattern in detection response.}
    \label{fig:car_color}
  \end{minipage}
  \hspace{0.02em}
  \begin{minipage}[b]{0.32\textwidth}
    \centering
  \includegraphics[width=1\textwidth]{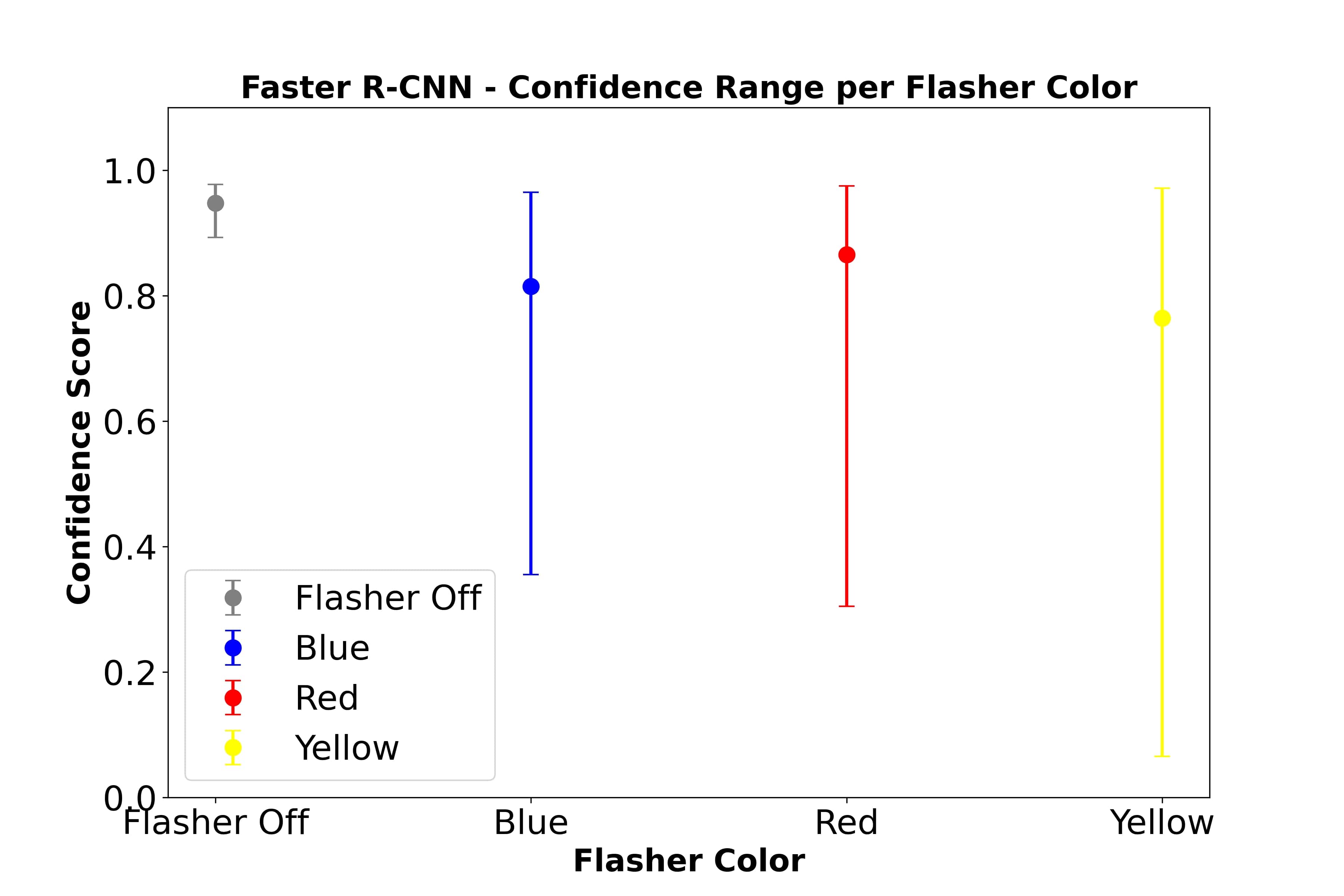} 
\caption{Faster R-CNN confidence score signal range for each examined emergency vehicle lighting color, as well as when the emergency vehicle lighting was off. The points in each range indicate the average value for each signal.}
    \label{fig:flasher_color}
  \end{minipage}
  \hspace{0.02em}
  \begin{minipage}[b]{0.32\textwidth}
    \centering
\includegraphics[width=1\textwidth]{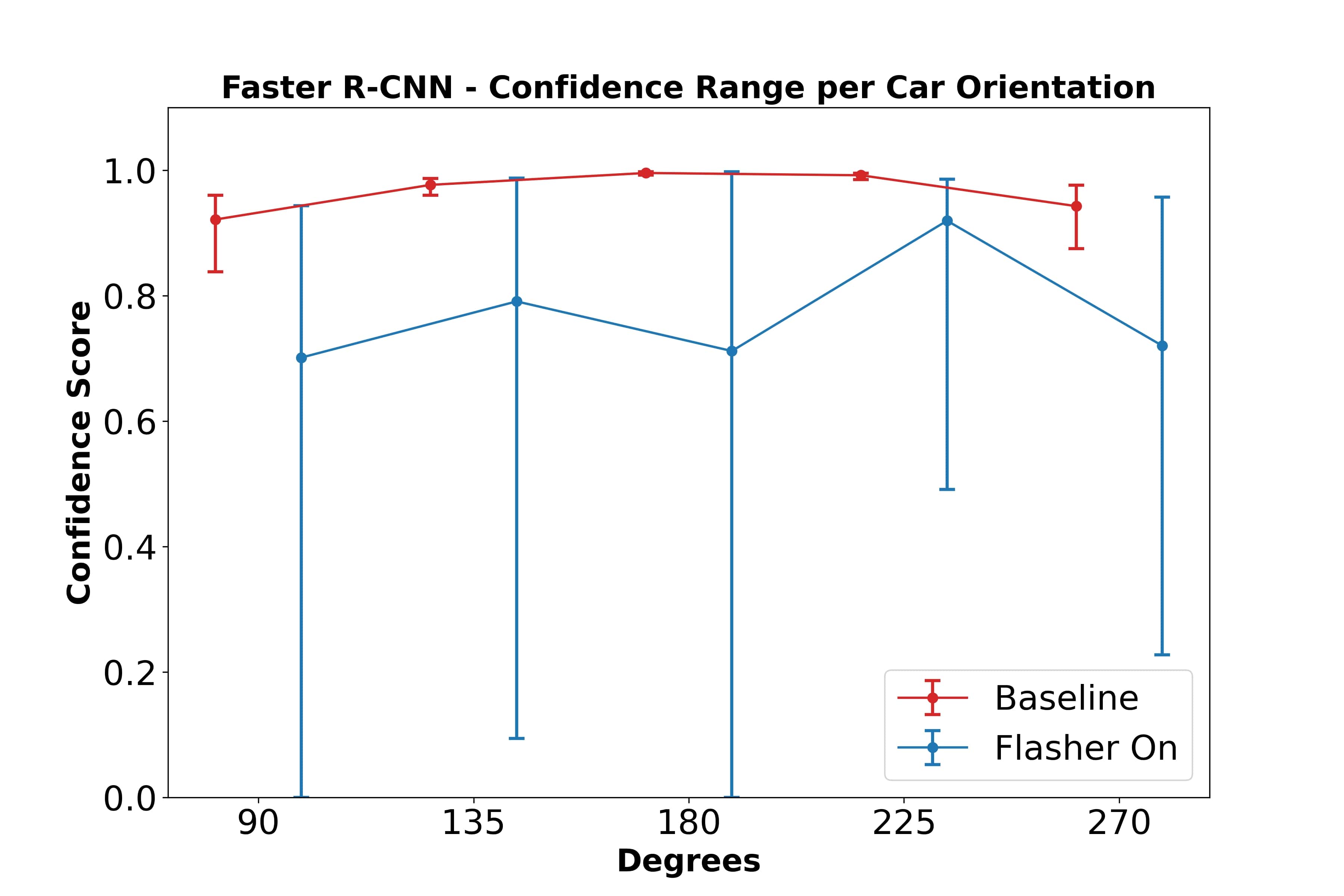}
\caption{Faster R-CNN confidence score signal range for each examined orientation between an ADAS' camera and an observed vehicle. The results obtained when the emergency vehicle lighting was on and off are presented in blue and red, respectively; the points in each range indicate the average value for each signal.}
\label{fig:orientation}
  \end{minipage}
\hspace{0.02em}
  \vspace{-1.5em}
\end{figure*}

\section{Appendix B: Effect of Emergency Vehicle's Characteristics}

In this section we examine the influence of an emergency vehicle's characteristics on the \textit{\EC} phenomenon, specifically investigating the effects of (1) the color of the emergency vehicle, (2) the color of the emergency vehicle lighting, and (3) the orientation of the vehicle requiring detection relative to the object detector/camera.
Due to space limitations, we only present the analysis performed using Faster R-CNN. 

\subsection{Effect of Emergency Vehicle Color} Here we investigate whether the color of the emergency vehicle impacts the detection capabilities of object detectors, analyzing their confidence range when detecting a vehicle whose emergency vehicle lighting is on.

\textbf{Experimental Setup:} A Samsung Galaxy S22 Ultra was positioned five meters away from four vehicles (a white Toyota RAV4, a silver Toyota RAV4, a yellow Fiat Panda, and a red Peugeot 207), each with emergency vehicle lighting mounted and activated on the roof. 
A 30-second video (24 FPS) of each car was recorded in a dark environment. 
Faster R-CNN was applied to the video frames, and an FFT (fast Fourier transform) graph was extracted from each detector's confidence score signal. 

\textbf{Results:} The results for the Faster R-CNN detector are presented in Fig. \ref{fig:car_color}. 
As can be seen, a consistent frequency peak of approximately 1.3 Hz was maintained for the confidence score signals of all car colors examined.

\vspace{0.2cm}
\begin{insight} \label{insight:4}
The \textit{\EC} phenomenon is present in cars with various colors.
\end{insight}

\subsection{Effect of Emergency Vehicle Lighting Color} Here we investigate whether the color of activated emergency vehicle lighting impacts the detection capabilities of object detectors, analyzing their confidence range when detecting a vehicle whose emergency vehicle lighting is on.

\textbf{Experimental Setup:} A Samsung Galaxy S22 Ultra was positioned five meters away from a grey Ford Fiesta with emergency vehicle lighting mounted on its roof. We recorded a 30-second video of the car at a rate of 24 FPS. Faster R-CNN was applied to the frames of each video, and the confidence range of each detector when recognizing the vehicle was measured for each video. This experiment was conducted four times: first with the emergency vehicle lighting turned off, and then three additional times with the emergency vehicle lighting activated, using different colored emergency vehicle lighting (blue, yellow, red).

\textbf{Results:} The results of this analysis for Faster R-CNN are presented in Fig. \ref{fig:flasher_color}. As can be seen, (1) detection confidence was higher when the emergency vehicle lighting was off, and (2) the confidence ranges were much larger when the emergency vehicle lighting were activated, regardless of color.

\begin{insight} \label{insight:7}
The \textit{\EC} phenomenon is not affected by the color of the emergency vehicle lighting.
\end{insight}





\subsection{Effect of Emergency Vehicle Orientation} Here we investigate how an emergency vehicle's orientation relative to the object detector/camera impacts the performance of object detectors in the presence of activated emergency vehicle lighting.

\textbf{Experimental Setup:} A Samsung Galaxy S22 Ultra was positioned five meters away from a grey Ford Fiesta with emergency vehicle lighting mounted on its roof. Five 30-second videos were recorded of the car; in each video, the angle between the camera and the front of the car was different (90, 135, 180, 225, 270 degrees). Four pretrained object detectors (YOLOv3, SSD, RetinaNet, Faster R-CNN) were applied to the frames from each video, and the confidence range of each detector when recognizing the vehicle was measured for each video. This experiment was conducted twice: first with the emergency vehicle lighting on and again when it was off.

\textbf{Results:} The results of this analysis for the Faster R-CNN detector are presented in Fig. \ref{fig:orientation}. 
As can be seen, the detection confidence was at or near its peak when there was a 225 degree angle between the camera and the car. The analysis showed that the confidence range tended to increase, indicating decreased stability in the object detector's confidence, when: (1) the camera and the car were positioned at angles other than 225 degrees, and (2) the emergency vehicle lighting was on.
\newline

\begin{insight} \label{insight:6}
The \textit{\EC} phenomenon is present in all analyzed orientations between the ADAS camera and the car requiring detection.
\end{insight}

\begin{figure*}[t]
  \centering
  \includegraphics[width=0.325\textwidth]{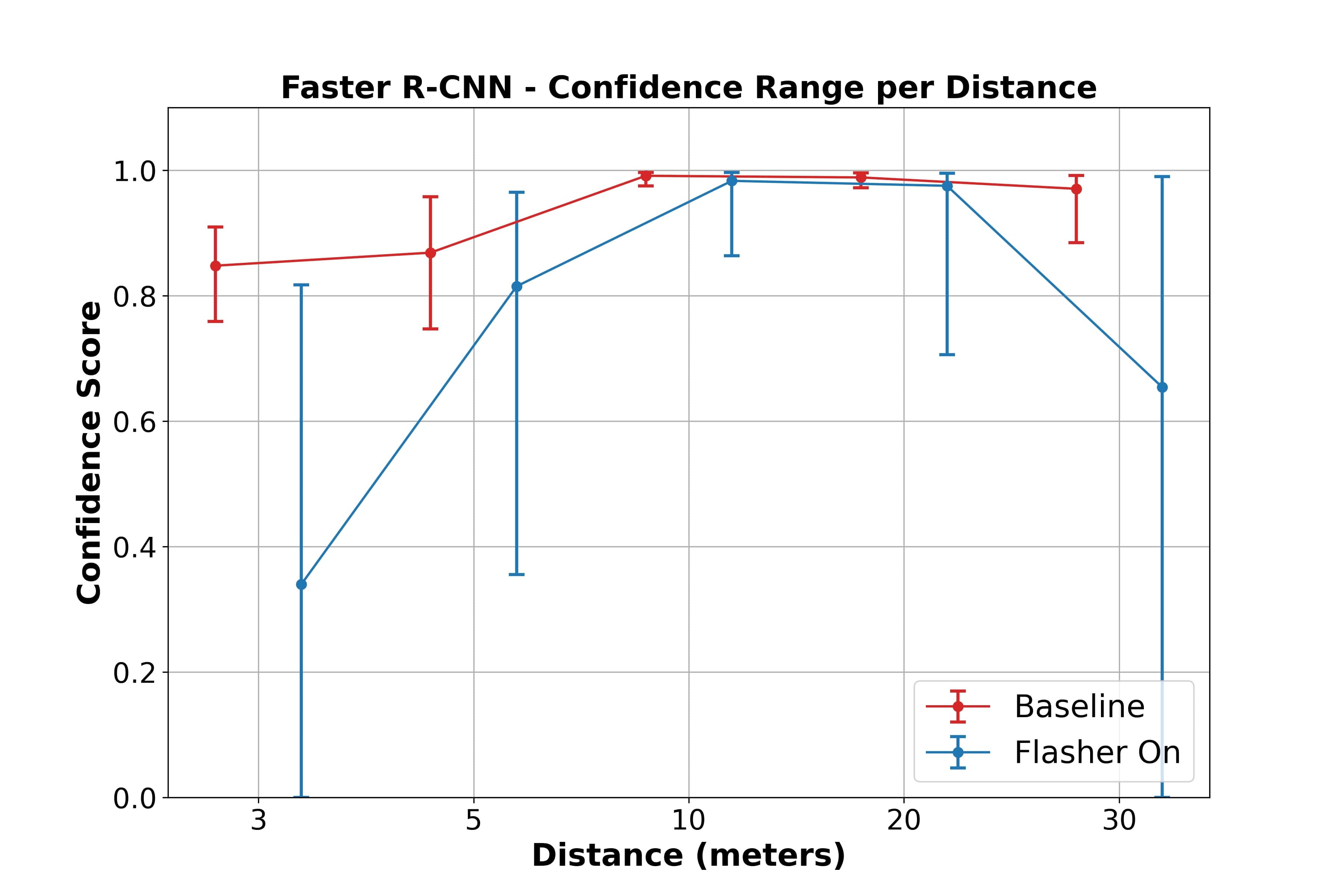}
\includegraphics[width=0.3\textwidth]{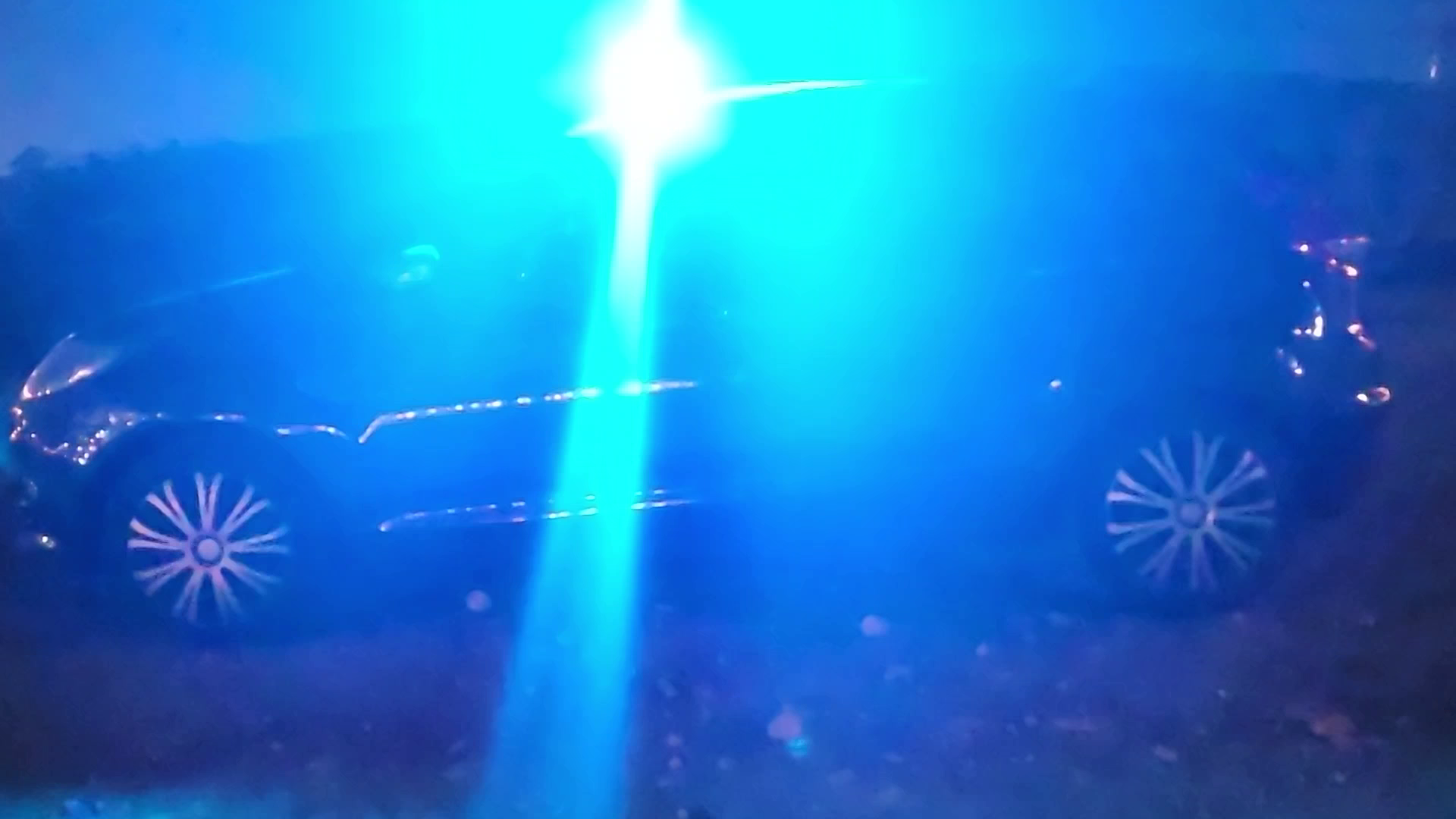}
\includegraphics[width=0.3\textwidth]{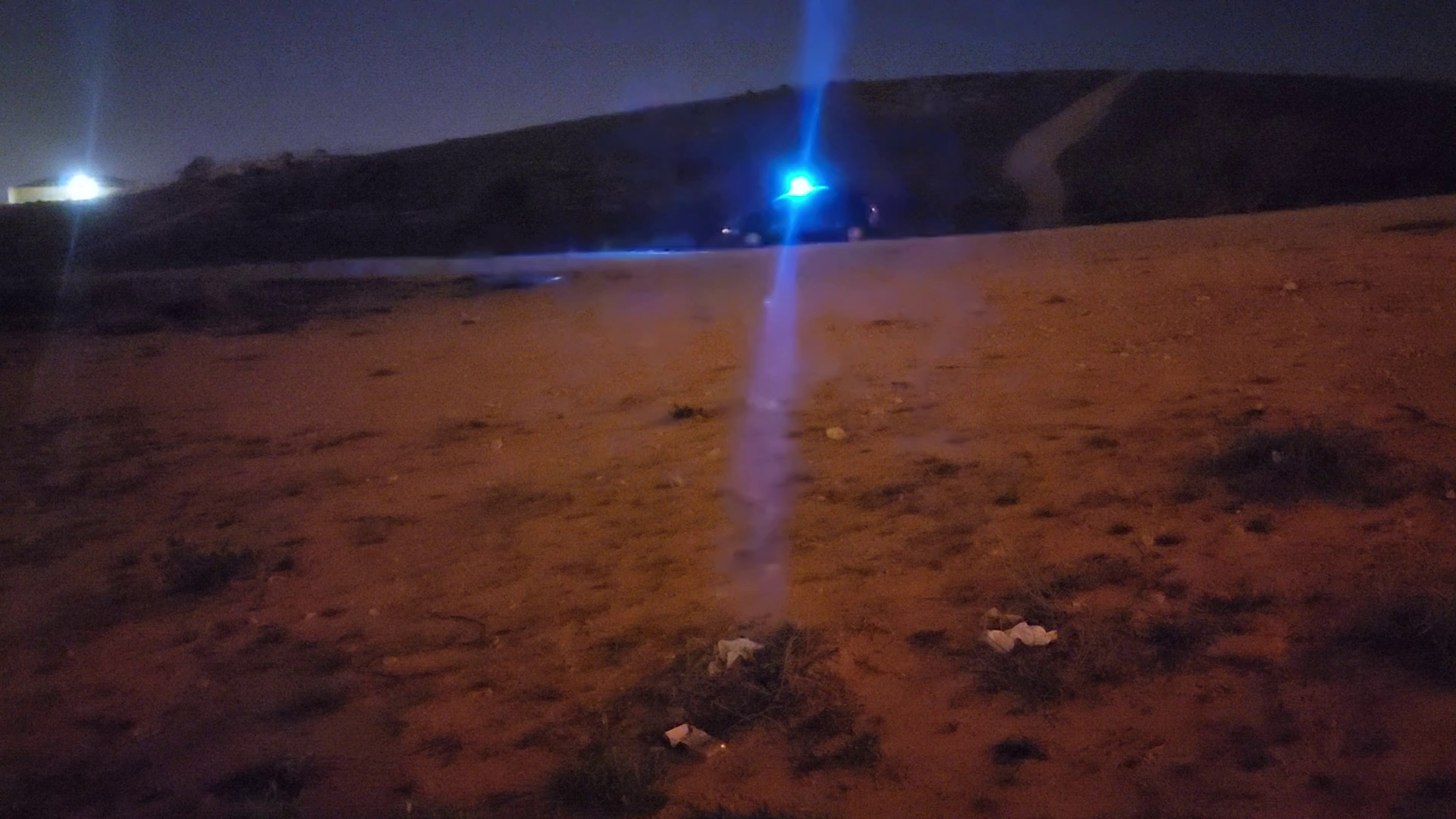}
\caption{Left: Faster R-CNN confidence score signal range for each examined distance between an autonomous vehicle's camera and an observed vehicle. The results obtained when the emergency vehicle lighting was on are presented in blue, while results obtained when the emergency vehicle lighting was off are in red; the points in each range indicate the average value for each signal. A car is not detected in the frames associated with the lowest confidence scores of Faster R-CNN in the videos recorded from 3 meters (center) and 30 meters (right).} 
    \label{fig:analysis-distances}
\end{figure*}

\section{Appendix C: Effect of Camera Distance} Here we analyze the effect of the distance between the semi-autonomous car's camera and the emergency vehicle.

\textbf{Experimental Setup:} A Samsung Galaxy S22 Ultra was positioned in front of a grey Ford Fiesta with emergency vehicle lighting mounted on its roof. 
Five experiments were conducted, each with the camera at a different distance from the car: 3 meters away (where the car occupies 74\% of the frame), 5 meters away (where the car occupies 42\% of the frame), 10 meters away (where the car occupies 7\% of the frame), 20 meters away (where the car occupies 1.9\% of the frame), and 30 meters away (where the car occupies 0.9\% of the frame). 
In each experiment, we recorded a 60-second video in which the emergency vehicle lighting was off for the first 30 seconds and on for the last 30 seconds.
Four object detectors were applied to the frames. The confidence score range for car detection was measured as a function of the distance. 

\textbf{Results:} The results for the Faster R-CNN object detector are presented in Fig. \ref{fig:analysis-distances}. The results for YOLOv3, SSD, and RetinaNet are presented in Fig. \ref{Fig:distance_appendix} in the appendix. 
As can be seen from the results, the average of the confidence range behaves in a parabolic manner with a peak score at a distance of 10 meters and decreased performance between 3-10 meters (because the emergency vehicle lighting significantly saturates the image at closer distances) and 10-30 meters (because the car cannot be seen in the picture due to the emergency vehicle lighting's effect at greater distances). 
The frames associated with the lowest confidence scores in the ranges associated with 3 and 30 meters are presented in Fig. \ref{fig:analysis-distances}.

\vspace{0.2cm}
\begin{insight} \label{insight:3.1}
The \textit{\EC} phenomenon is dependent on the distance between the autonomous car's camera and the vehicle requiring detection.
\end{insight}

\section{Appendix D: Additional Material}

\begin{figure*}
  \centering
\includegraphics[width=0.3\textwidth]{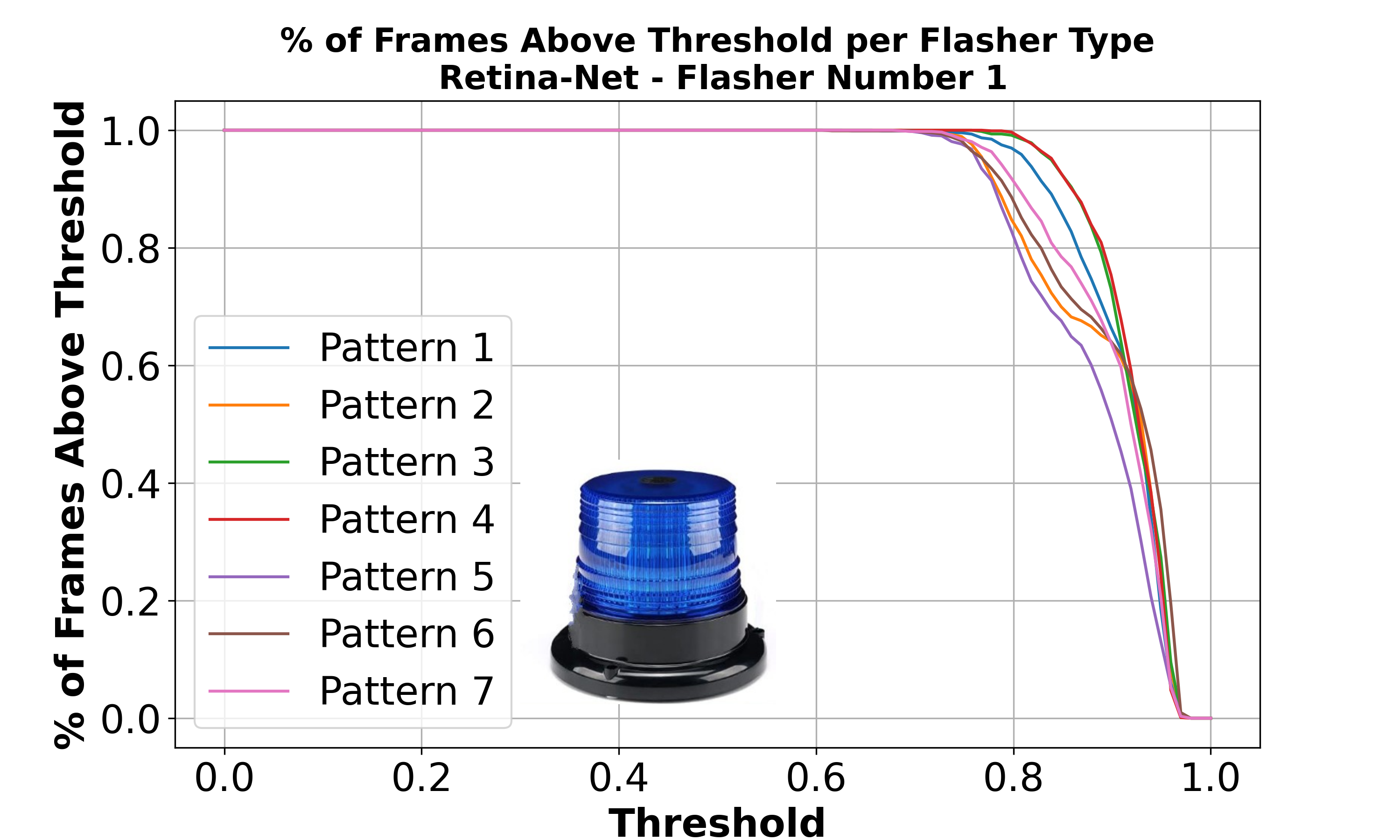}
\includegraphics[width=0.325\linewidth]{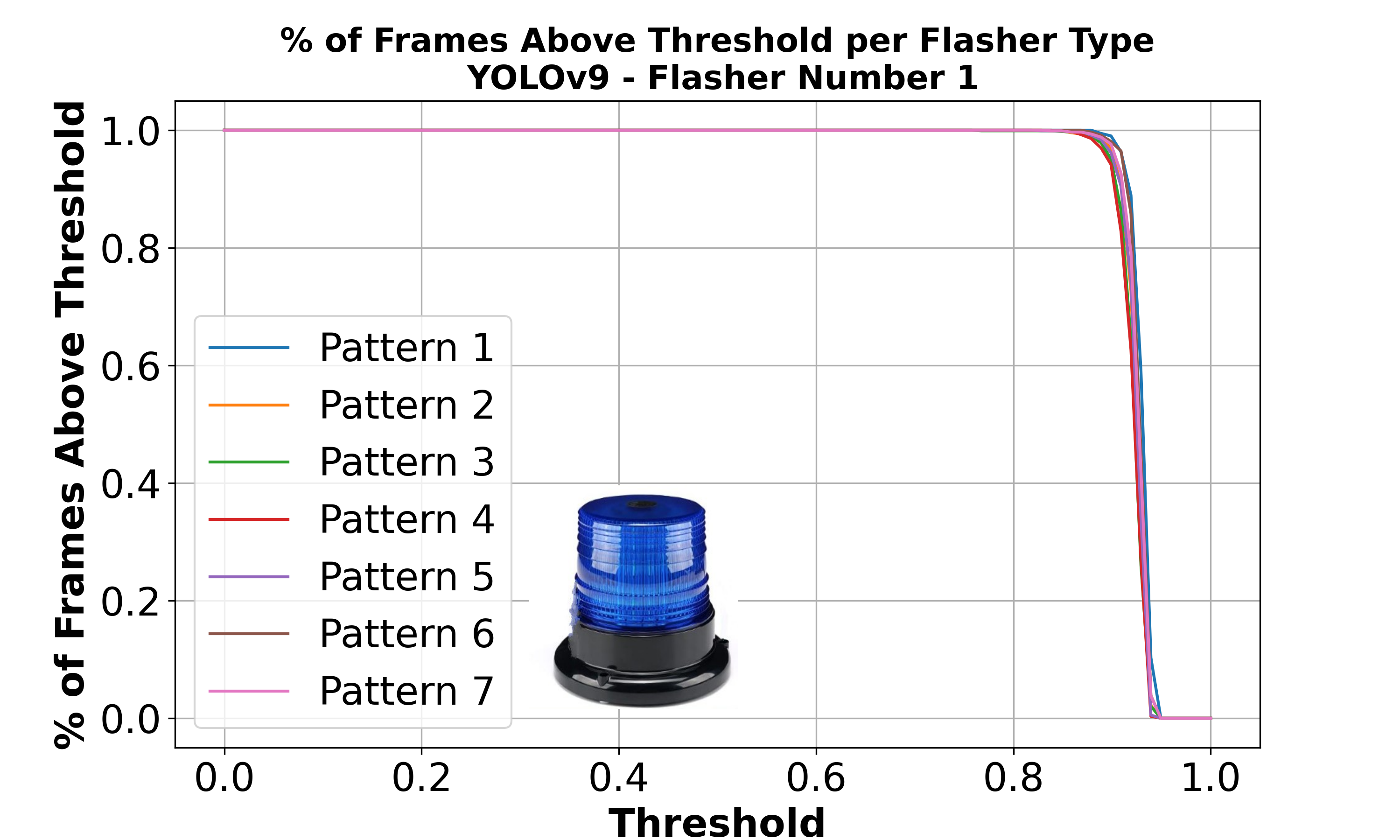}
   \includegraphics[width=0.325\linewidth]{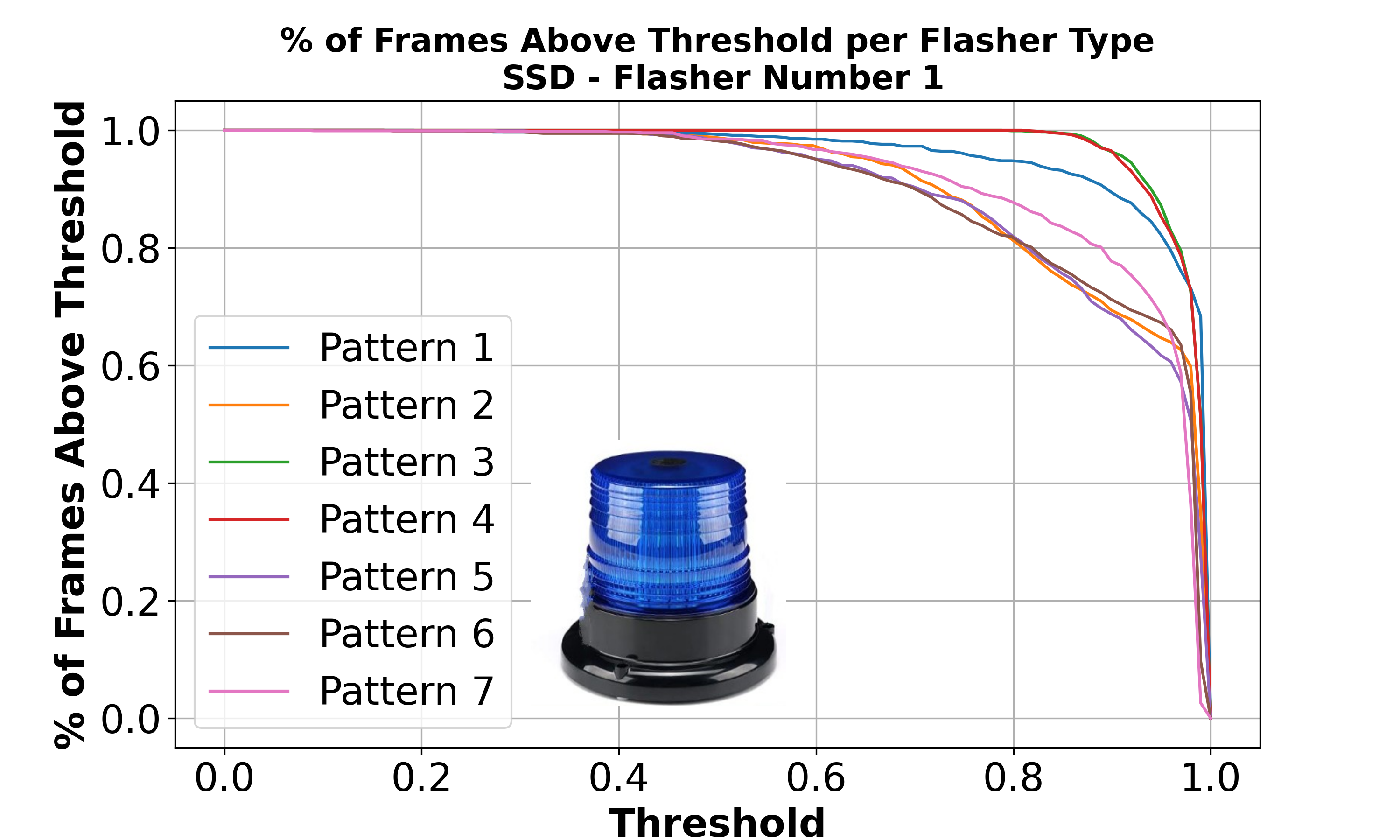}\\
    \includegraphics[width=0.3\textwidth]{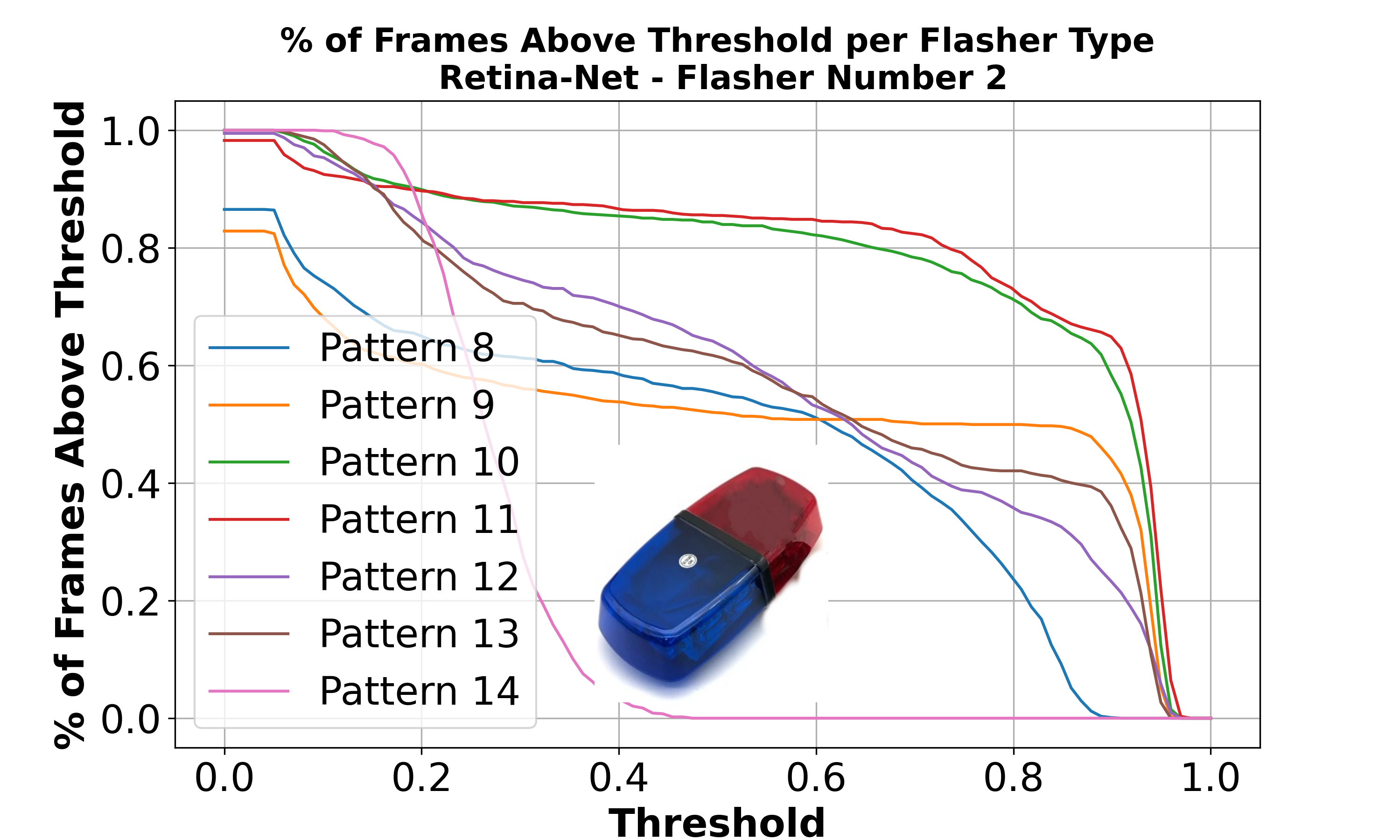}
    \includegraphics[width=0.325\linewidth]{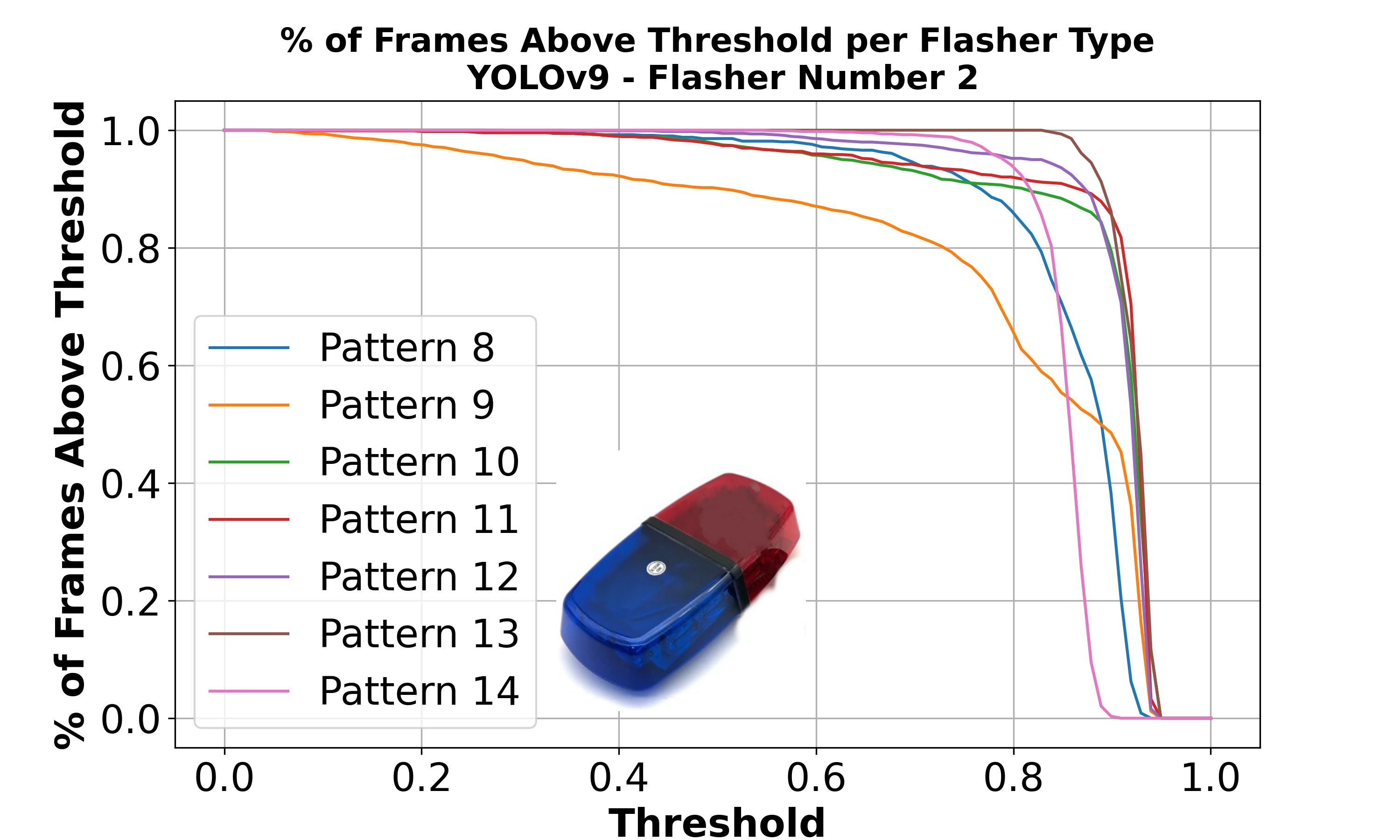}
    \includegraphics[width=0.325\linewidth]{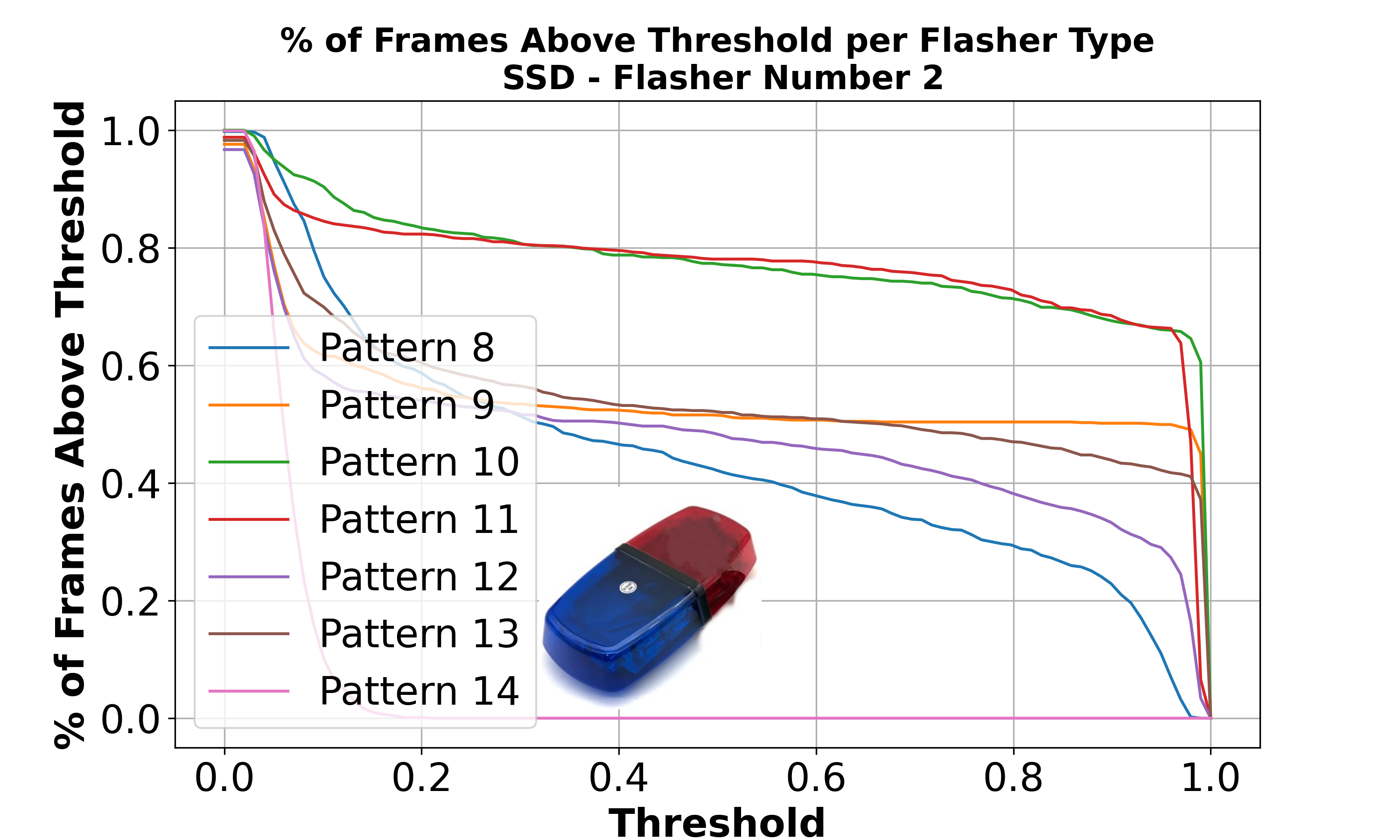}
\caption{Comparison of percentage of frames with confidence higher than threshold for 14 emergency vehicle lighting patterns observed by RetinaNet, YOLOv3, and SSD.}
  \label{Fig:retina-two-flashers}
\end{figure*}


\begin{figure*}[t]
  \centering
\includegraphics[width=0.32\textwidth]{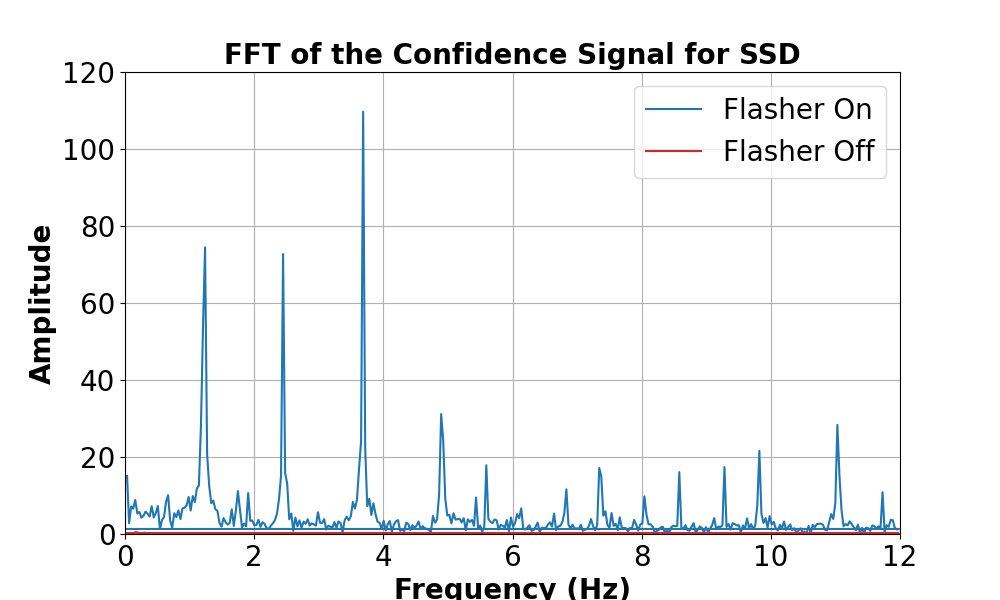}
\includegraphics[width=0.32\textwidth]{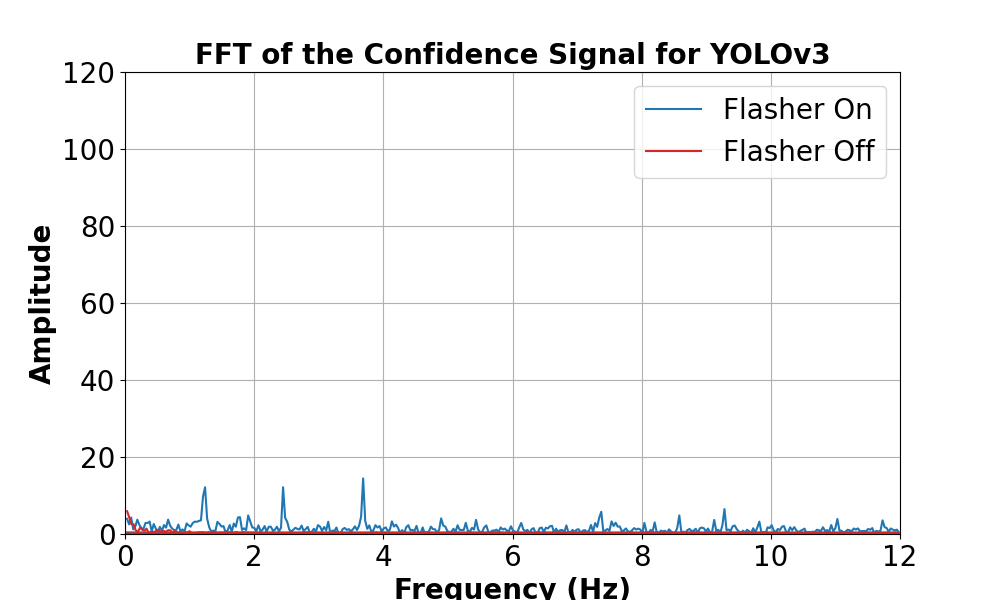}
\includegraphics[width=0.32\textwidth]{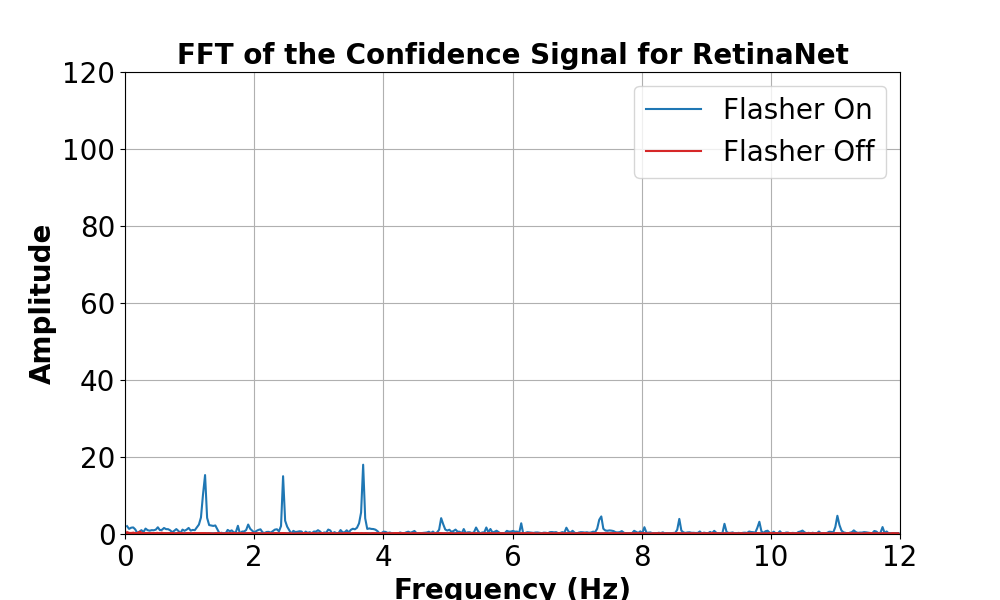}
\caption{Extracted FFT graphs from the confidence signals of three object detectors when the emergency vehicle lighting is on. A peak around 1.3 Hz can be observed in both FFT graphs. }
    \label{fig:ffts-all-ods}
\end{figure*}



\begin{figure*}
  \centering

\includegraphics[width=0.3\textwidth]{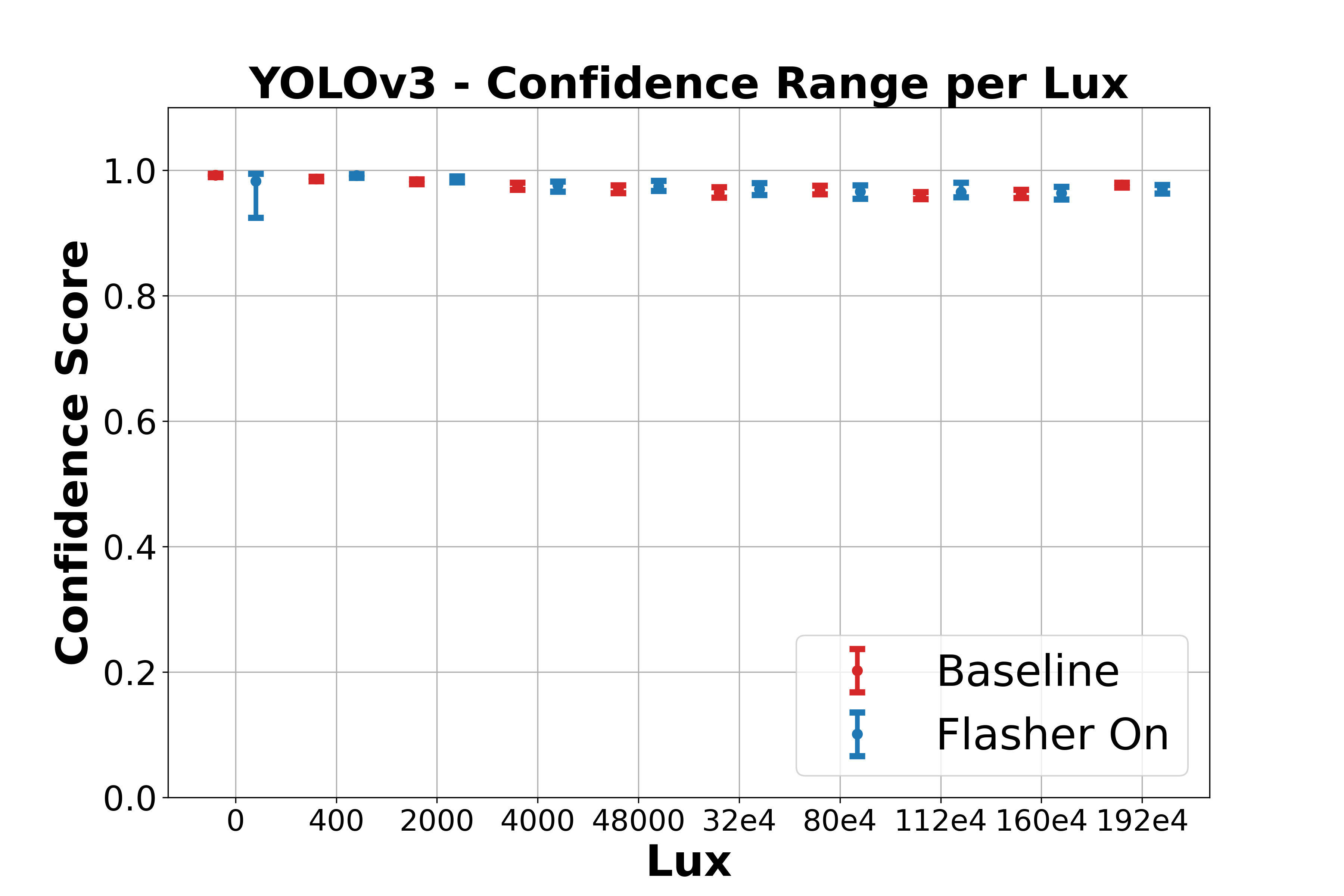}
\includegraphics[width=0.3\textwidth]{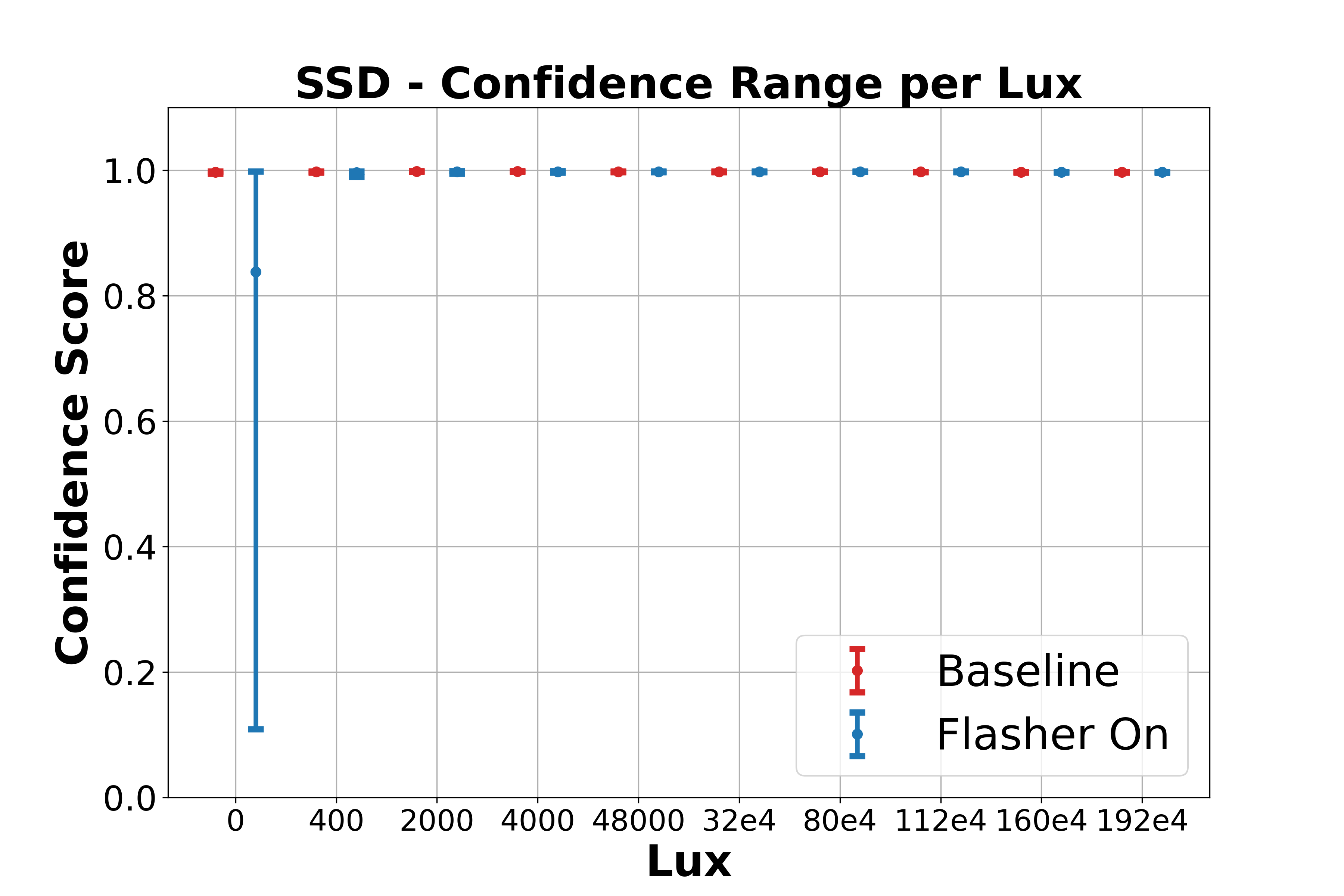}
\includegraphics[width=0.3\textwidth]{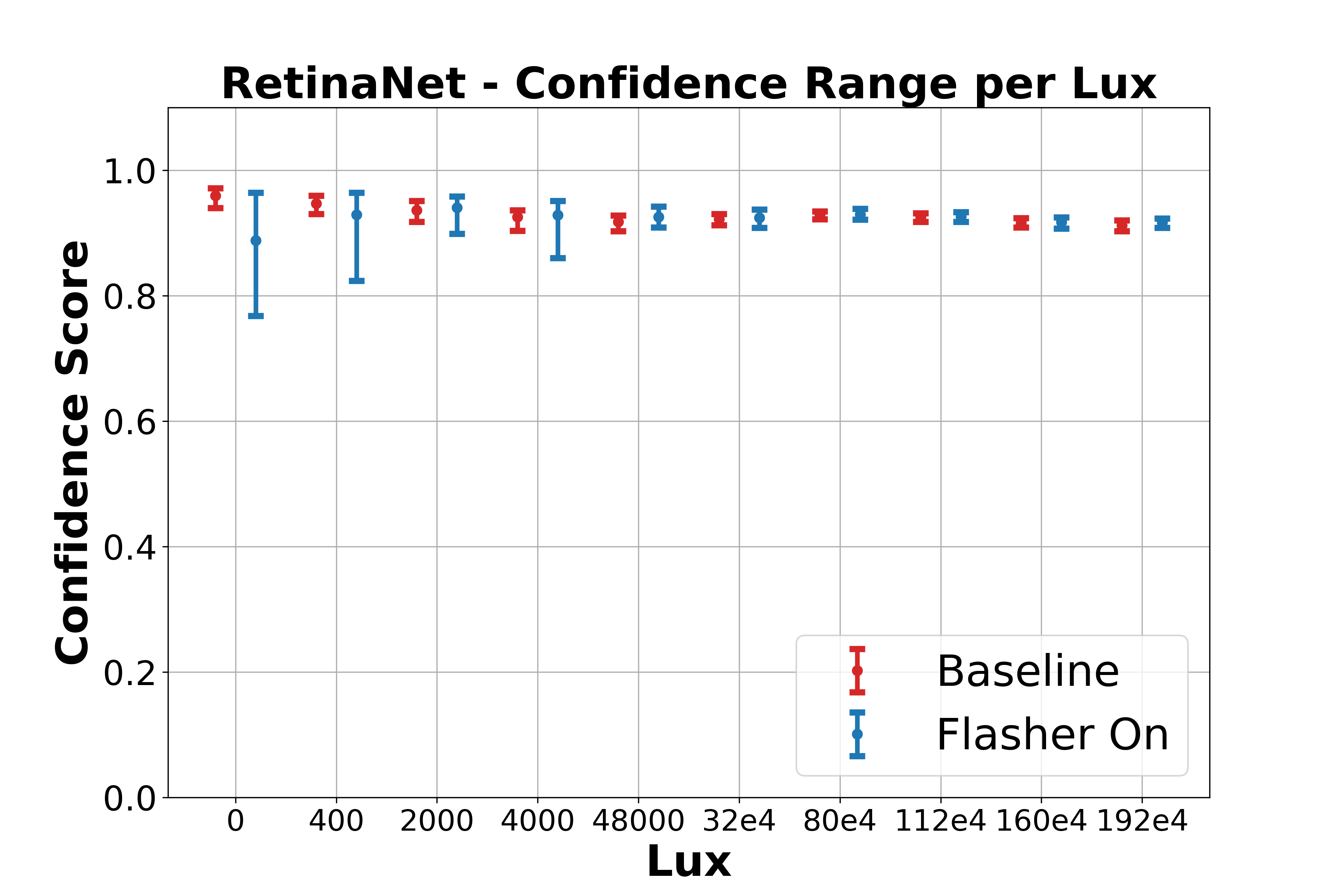}
\caption{Confidence score signal ranges of various object detectors (YOLO, SSD, and RetinaNet) for each examined lux value. The results obtained when the emergency vehicle lighting was on are presented in blue, while the results obtained when the emergency vehicle lighting was off are in red; the points in each range indicate the average value for each signal.
}
  \label{Fig:ambient_appendix}
\end{figure*}

\begin{figure*}
  \centering

\includegraphics[width=0.3\textwidth]{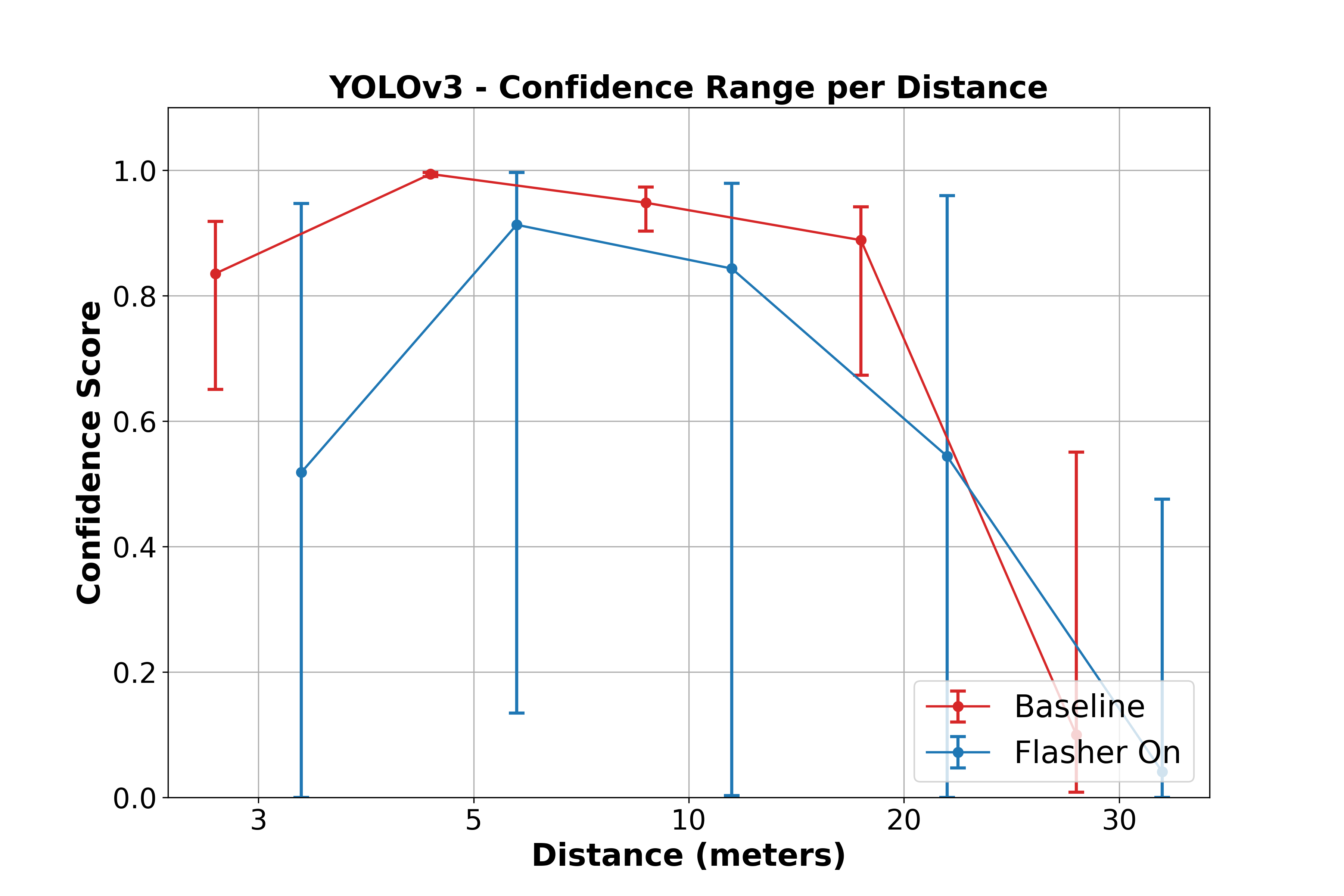}
\includegraphics[width=0.3\textwidth]{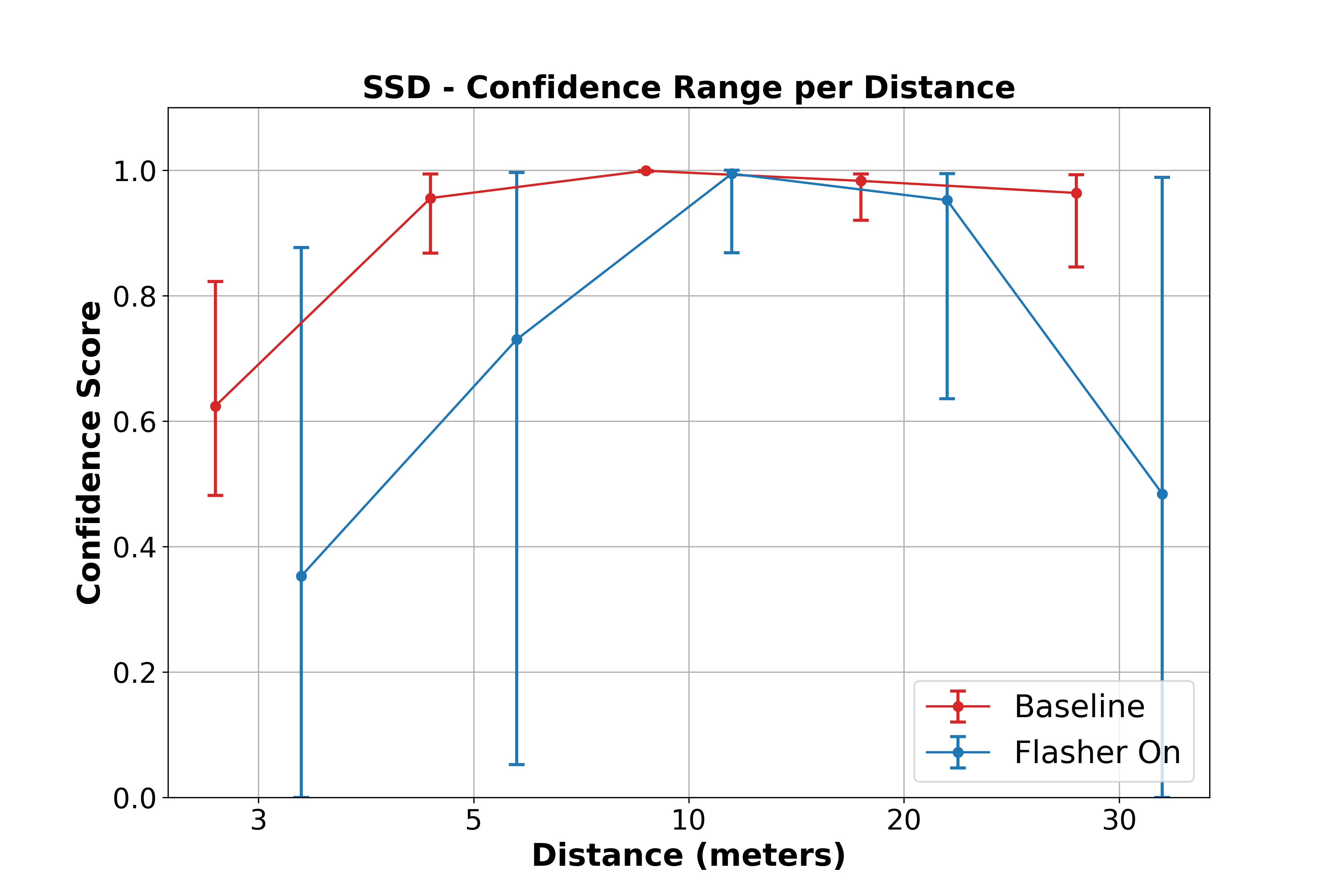}
\includegraphics[width=0.3\textwidth]{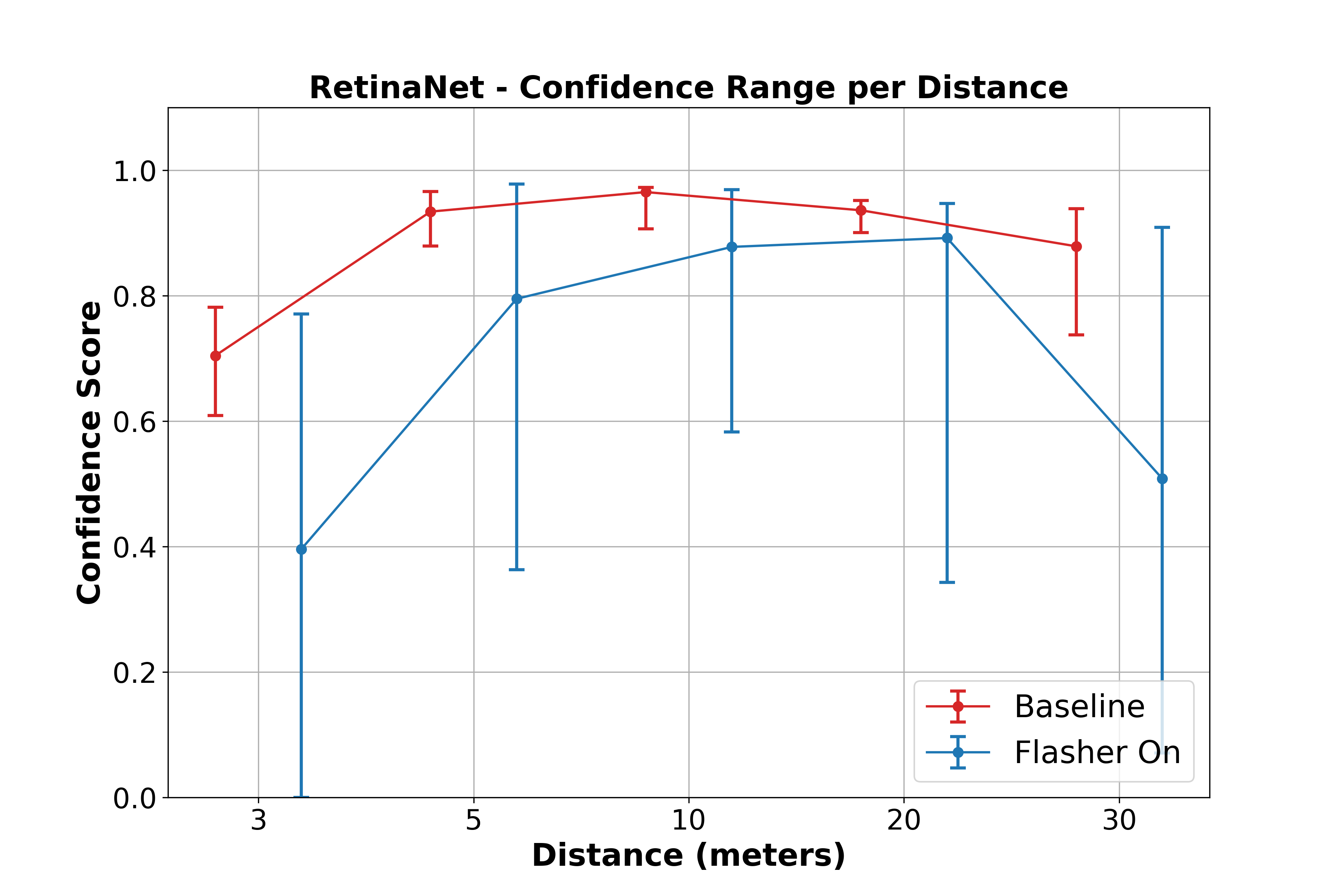}
\caption{Confidence score signal ranges of various object detectors (YOLO, SSD, and RetinaNet) for each examined distance between an ADAS' camera and an observed vehicle. The results obtained when the emergency vehicle lighting was on are presented in blue, while the results obtained when the emergency vehicle lighting was off are in red; the points in each range indicate the average value for each signal.
}
  \label{Fig:distance_appendix}
\end{figure*}

\begin{figure*}
  \centering

\includegraphics[width=0.3\textwidth]{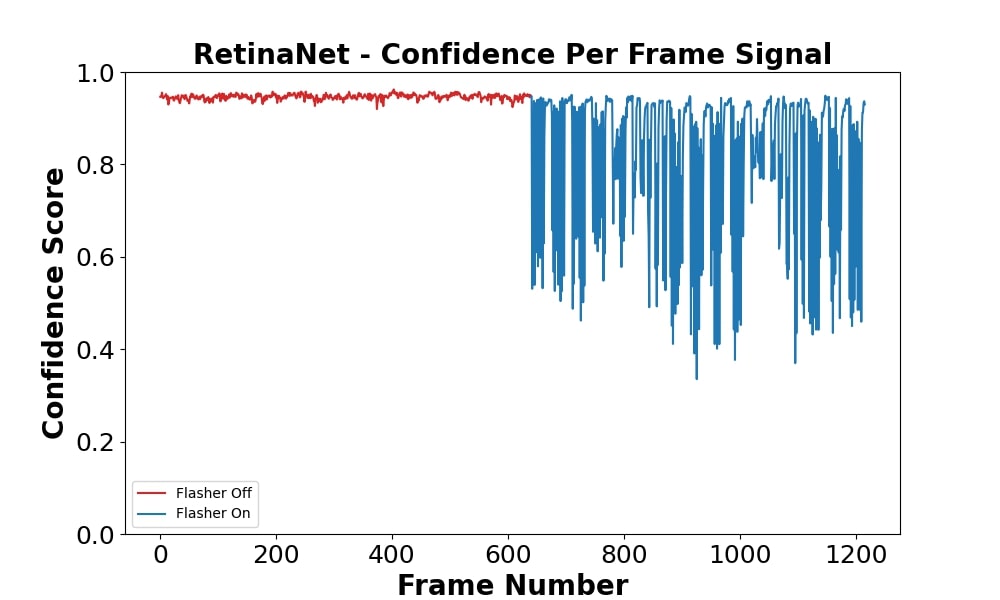}
\caption{confidence score per frame with Tesla 2023 Tesla model 3 frontal camera obtained with RetinaNet  
}
  \label{Fig:retina_showcase}
\end{figure*}

\begin{figure*}
  \centering

\includegraphics[width=0.3\textwidth]{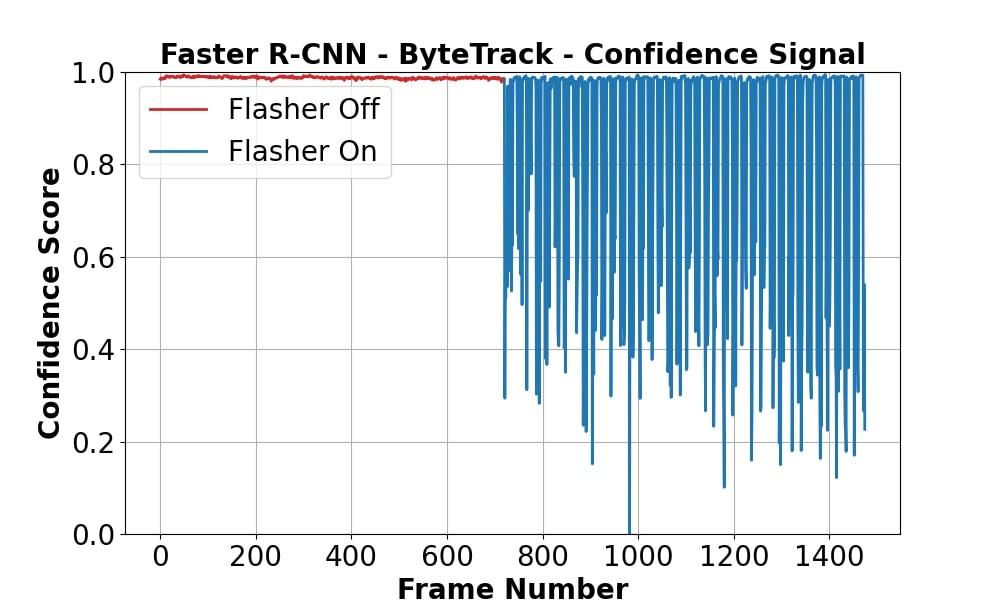}
\includegraphics[width=0.3\textwidth]{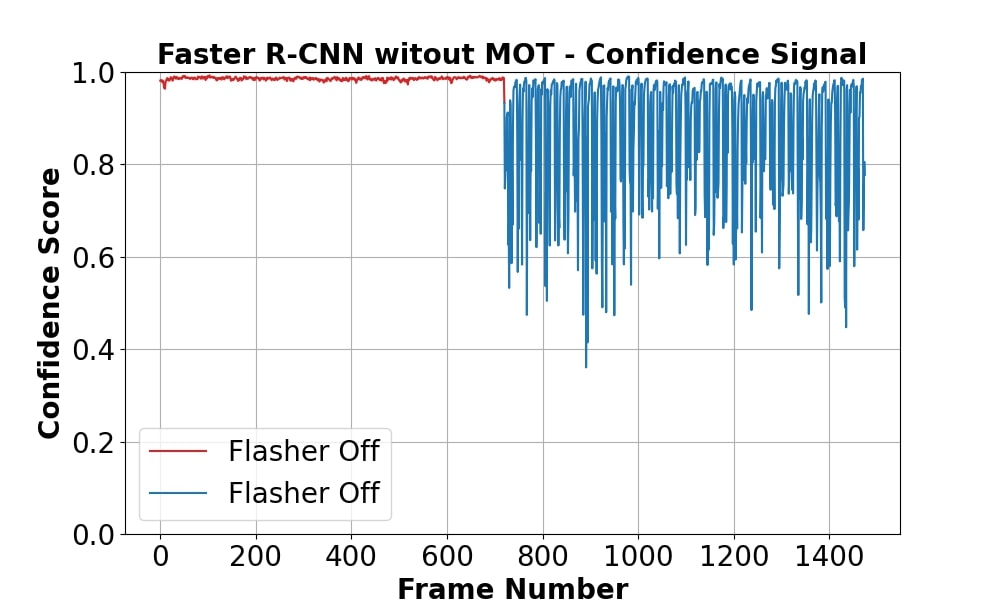}

\caption{The confidence score signals of bytetrack object tracker model applied on the Faster R-CNN object detector and Faster R-CNN object detector without any multi object tracking model.
}
  \label{Fig:bytetrack_no_mot}
\end{figure*}

\begin{lstlisting}[language=python, breaklines= true, numbersep=0pt,showstringspaces=false,label = listing-script, frame=single, escapechar={|}, captionpos=b,caption = The script used to simulate the manual noise.]
def make_light(x, y, img_shape, option):
    r_w, r_h = option_dict[option]
    channel = 0 if np.random.rand() > 0.5 else 2
    layer_c = np.zeros(img_shape)
    layer_c[(y - r_h):(y + r_h), (x - r_w):(x + r_w), channel] = 255 * 65
    layer_c[:, :, channel] = gaussian_filter(layer_c[:, :, channel], sigma=100, mode='constant')
    layer_w = np.zeros(img_shape)
    layer_w[(y - r_h):(y + r_h), (x - r_w):(x + r_w), :] = 255 * 900
    layer_w = gaussian_filter(layer_w, sigma=100, mode='constant')    
    return np.clip(layer_c + layer_w, 0, 255)
\end{lstlisting}

\footnotesize 
\Urlmuskip=0mu plus 1mu\relax

\end{document}